\documentclass[10pt,journal,commsoc]{IEEEtran}

\IEEEoverridecommandlockouts
\usepackage{cite}
\usepackage{amsmath,amssymb,amsfonts}
\usepackage{algorithmic}
\usepackage{graphicx}

\usepackage{hyperref}
\hypersetup{hidelinks}       % hyperlinks

\usepackage{xcolor}
\usepackage{array}
\usepackage{subfigure}
\usepackage{graphicx}
\usepackage{grffile}
\usepackage{amssymb}
\usepackage{amsmath}
\usepackage{amsthm}
\usepackage{multirow}
\usepackage{color}
\usepackage{url}
\usepackage{cite}
\usepackage{textcomp}
\usepackage{xcolor}
\usepackage[lined,linesnumbered,ruled,norelsize]{algorithm2e}
\usepackage{todonotes}
\setuptodonotes{inline}
\newtheorem{lemma}{Lemma}

\newtheorem{theorem}{Theorem}
\newtheorem{proposition}{Proposition}
\newtheorem{definition}{Definition}
\newtheorem{assumption}{Assumption}

\def\BibTeX{{\rm B\kern-.05em{\sc i\kern-.025em b}\kern-.08em
    T\kern-.1667em\lower.7ex\hbox{E}\kern-.125emX}}

\begin{document}

\title{Incentivizing Massive Unknown Workers for Budget-Limited Crowdsensing: From Off-Line and On-Line Perspectives}

\author{Feng~Li, %~\IEEEmembership{Member,~IEEE,}
        Yuqi~Chai, 
        Huan~Yang,
        Pengfei~Hu,
        Lingjie~Duan 
\IEEEcompsocitemizethanks{\IEEEcompsocthanksitem F. Li, Y. Chai and Pengfei Hu are with School of Computer Science and Technology, Shandong University, Qingdao 266237, China.\protect\\
% note need leading \protect in front of \\ to get a newline within \thanks as
% \\ is fragile and will error, could use \hfil\break instead.
E-mail: fli@sdu.edu.cn, 202015066@mail.sdu.edu.cn, phu@sdu.edu.cn
\IEEEcompsocthanksitem H. Yang is with College of Computer Science and Technology, Qingdao University, Qingdao, China. \protect\\
E-mail: cathy\_huanyang@hotmail.com
\IEEEcompsocthanksitem L. Duan is with The Engineering Systems and Design Pillar, Singapore
University of Technology and Design, Singapore. \protect\\
E-mail: lingjie\_duan@sutd.edu.sg
}% <-this % stops an unwanted space
%\thanks{Manuscript received April 19, 2005; revised August 26, 2015.}
}

\IEEEtitleabstractindextext{
\begin{abstract}
  How to incentivize strategic workers using limited budget is a very fundamental problem for crowdsensing systems; nevertheless, since the sensing abilities of the workers may not always be known as prior knowledge due to the diversities of their sensor devices and behaviors, it is difficult to properly select and pay the unknown workers. Although the uncertainties of the workers can be addressed by the standard \textit{Combinatorial Multi-Armed Bandit} (CMAB) framework in existing proposals through a trade-off between exploration and exploitation, we may not have sufficient budget to enable the trade-off among the individual workers, especially when the number of the workers is huge while the budget is limited. Moreover, the standard CMAB usually assumes the workers always stay in the system, whereas the workers may join in or depart from the system over time, such that what we have learnt for an individual worker cannot be applied after the worker leaves. To address the above challenging issues, in this paper, we first propose an off-line \textit{Context-Aware CMAB-based Incentive} (CACI) mechanism. We innovate in leveraging the exploration-exploitation trade-off in an elaborately partitioned context space instead of the individual workers, to effectively incentivize the massive unknown workers with a very limited budget. We also extend the above basic idea to the on-line setting where unknown workers may join in or depart from the systems dynamically, and propose an on-line version of the CACI mechanism. Specifically, by the exploitation-exploration trade-off in the context space, we learn to estimate the sensing ability of any unknown worker (even it never appeared in the system before) according to its context information. We perform rigorous theoretical analysis to reveal the upper bounds on the regrets of our CACI mechanisms and to prove their truthfulness and individual rationality, respectively. Extensive experiments on both synthetic and real datasets are also conducted to verify the efficacy of our mechanisms.
\end{abstract}

% Note that keywords are not normally used for peerreview papers.
}

\maketitle

\IEEEdisplaynontitleabstractindextext
\IEEEpeerreviewmaketitle

\section{Introduction} \label{sec:intro}
  Recent years have witnessed the proliferation of smart devices (e.g., smart phones). These devices are usually equipped with various sensors and can be utilized to conduct sensing tasks ubiquitously. This development has given birth to a paradigm of crowdsensing~\cite{KhanXAA-SURV13,CapponiFKFKB-COMST19}. In particular, thousands or even millions of human crowds (a.k.a., workers) can be recruited to sense valuable data with their sensor devices in a broad-scale area for various applications (e.g., traffic jam alerts, air quality monitoring, urban business survey, spectrum sensing etc.) \cite{WangTLP-INFOCOM17,LiuLZMZ-IMWUT18,JiangZZLSMFWL-TMC21}.

  The effectiveness of crowdsensing systems considerably depends on the quality of workers; nevertheless, individuals may be hesitant to contribute their sensing capabilities, due to concerns of costs and risks. For example, for a smartphone user, participating in a task of collecting sensory data inevitably results in resource consumption. Moreover, the collected data may include location information, which may cause privacy-sensitive users to feel uneasy. Therefore, incentivizing strategic workers under a given budget is a very fundamental problem for crowdsensing systems~\cite{ZhangYSLTXM-COMST16}. In this paper, we consider a reverse auction process where the workers first communicate their private bids to a task auctioneer (or assigner), and the task assigner decides which workers to select and how to pay the selected workers according to the workers' sensing abilities and bids. There have been many state-of-the-art proposals investigating the incentive problem (e.g.,\cite{FengZZNV-INFOCOM14,ZhaoLM-INFOCOM14,GaoXWHH-TMC18,PengWC-TMC18}), and most of them assume the sensing abilities of workers are known as prior knowledge. Unfortunately, the assumption may not consistently apply in real-world scenarios, as it is often challenging to pre-profile workers with various sensor devices and behaviors. This difficulty becomes more pronounced when considering the potentially vast number of unknown workers.

  To address the uncertainties of workers, one popular choice is to apply the \textit{Combinatorial Multi-Armed Bandit} (CMAB) framework, where the unknown workers are sequentially selected through a trade-off between exploration and exploitation~\cite{HanZL-TON16,RangiF-AAMAS18,GaoWYXC-ICPADS19,ZhaoXWXHZ-TMC21,GaoWXC-INFOCOM20,SongJ-INFOCOM21}. The CMAB framework also has been applied in designing incentive mechanisms to select and pay unknown workers, e.g., in \cite{BiswasJMN-AAMAS15,XiaoWZG-ICDCS20,GaoHXWSZ-INFOCOM21}. Nonetheless, the standard CMAB framework addresses the exploration-exploitation trade-off directly among the individual unknown workers. Consequently, when the budget is limited whereas the number of the unknown workers is huge, employing the standard CMAB framework to learn the sensing abilities of each individual worker can lead to substantial recruitment costs. For instance, if the given budget is insufficient to select and compensate each worker at least once, the incentive mechanisms based on the standard CMAB framework may be impractical to initiate.

  Another challenging issue is the dynamic arrival and departure of workers. In particular, workers may not be consistently available; they may intermittently join in or leave from the crowdsensing platform over time~\cite{ZhaoLM-TON16,LiLYZZW-TNSE21,ZhengYXWGC-TMC22}. In this case, when the workers' sensing abilities are unknown, the standard CMAB framework is not applicable, since it learns the sensing ability of each individual. In particular, for any unknown worker, even its sensing ability was well learnt, the knowledge about the worker's sensing ability could not be used after the worker leaves. Although there have been many variants of bandits which can be applied to volatile arms (i.e., the on-line workers in our case)~\cite{KleinbergNS-ML10,BnayaPSF-AAAI13}, they are based on the standard \textit{Multi-Armed Bandit} (MAB), and applying their ideas to combinatorial problems is highly non-trivial such that little effort has been made to extend them to CMAB setting.

  In this paper, we first propose an off-line \textit{Context-Aware CMAB-based Incentive} (CACI) mechanism to incentivize a fixed large group of unknown workers with limited budget \footnote{Different from many existing proposals where an ``off-line'' solution means the workers' sensing abilities are already known (e.g., \cite{BiswasJMN-AAMAS15,XiaoWZG-ICDCS20,GaoHXWSZ-INFOCOM21}), we hereby claim our CACI mechanism is \textit{off-line} since we assume the unknown workers whom we select and pay are given in an \textit{off-line} manner such that they are fixed and always stay in the system. In fact, we will introduce our \textit{baseline} incentive mechanisms where we suppose the workers' sensing abilities are already known in Sec.~\ref{sec:baseline}, to inspire the design and analysis of our mechanisms for unknown workers.}. A common wisdom suggests that workers with similar contexts usually have similar sensing abilities~\cite{MullerTSK-TON18,WuLMXLW-IOTJ19}, and we thus innovate in leveraging the CMAB framework to learn the correlation between the context and the sensing ability instead of the sensing abilities of the individual workers. In particular, our off-line CACI mechanism mainly consists of two phases, i.e., \textit{exploration} phase and \textit{exploitation} phase. In the exploration phase, by partitioning the context space into a finite set of sub-spaces (i.e., so-called ``hypercubes'') with a fine tuned granularity, we iteratively explore the hypercubes to estimate the \textit{Upper Confidence Bounds} (UCBs) on the sensing abilities of the workers with similar contexts in each hypercube. In the following exploitation phase, we exploit the workers according to the estimates in a greedy manner. We also design a policy to pay the selected workers, such that the total budget constraint as well as the truthfulness and individual rationality of our incentive mechanism are respected. Through our off-line CACI mechanism to explore and exploit the finite context space partitions, we can effectively incentivize a (fixed) huge group of unknown workers to maximize expected cumulative sensing revenue, even when the total budget is rather limited.

  We then apply the above basic idea to design an on-line version of our CACI mechanism, when the massive unknown workers are given dynamically over time. Similar to its off-line counterpart, our on-line CACI mechanism leverages the exploration-exploitation trade-off in the elaborately partitioned context space but in a quite different way due to the dynamics of the workers. In particular, in each exploration-and-exploitation iteration, we update the UCBs of the available workers to make the selection and payment decisions in an on-line manner. Since we innovate in learning the correlation between context space and sensing ability through our CACI mechanisms, even given a worker who never appeared in the system before, we can accurately estimate its sensing ability according to its context and make proper selection and payment decision accordingly. To the best of our knowledge, we are the first one to design incentive mechanism for on-line unknown workers in crowdsensing systems.

  In summary, our main contributions are as follows:
  \begin{itemize}
    \item To the best of our knowledge, this is the first work studying how to incentivize massive unknown workers for budget-limited crowdsensing by harnessing the workers' context information.
    \item We innovate in leveraging the CMAB framework by conducting the exploration-exploitation trade-off in finite context space partitions instead of individual workers, and propose an off-line mechanism to incentivize a fixed group of unknown workers, where the number of the workers is quite huge whereas the given budget is limited.
    \item We extend the above basic idea and design an on-line mechanism to incentivize dynamic workers who may join in or depart from the system over time.
    \item Through rigorous theoretical analysis, we finally quantify the performance gaps (a.k.a. \textit{regret}) between our mechanisms and the corresponding baseline mechanisms where the workers' sensing abilities are known as prior knowledge. We also perform extensive experiments on both synthetic and real datasets to verify their efficacy.
  \end{itemize}

  The remaining of our paper is organized as follows. Related literature is surveyed in Sec.~\ref{sec:survey}. We introduce our system model and describe our problems in Sec.~\ref{sec:sys}. In Sec.~\ref{sec:baseline}, we present two baseline mechanisms where qualities of workers are assumed to be known. Inspired by the baseline mechanisms, we then design and analyze our off-line and on-line CACI mechanisms in Sec.~\ref{sec:offline} and Sec.~\ref{sec:online-caci}, respectively. We evaluate our mechanisms with both synthetic and real datasets in Sec.~\ref{sec:exp}. We finally conclude this paper in  and Sec.~\ref{sec:con}.

\section{Related Work}  \label{sec:survey}
  Incentive mechanism design plays an important role in crowdsensing~\cite{ZhangYSLTXM-COMST16}. There have been many existing proposals investigating the incentive problem from various perspectives~\cite{FengZZNV-INFOCOM14,ZhaoLM-INFOCOM14,GaoXWHH-TMC18,PengWC-TMC18}. Nevertheless, most of these proposals assume the sensing abilities of workers are already known.

  To address the uncertainties of workers in crowdsensing systems, one popular choice is to apply the CMAB framework. For example, in \cite{HanZL-TON16,RangiF-AAMAS18,ZhaoXWXHZ-TMC21}, workers with unknown sensing abilities are selected sequentially under a limited total budget to perform a given sensing task. \cite{SongJ-INFOCOM21} adopts the empirical entropy of the data reported by workers to measure the sensing revenue. In \cite{GaoWYXC-ICPADS19,GaoWXC-INFOCOM20}, multi-task assignment problems are investigated. In the above proposals, the arms in the CMAB framework (e.g., the workers in \cite{HanZL-TON16,RangiF-AAMAS18,ZhaoXWXHZ-TMC21,SongJ-INFOCOM21} or the worker-task combinations in \cite{GaoWYXC-ICPADS19,GaoWXC-INFOCOM20}) are exploited and explored individually. Therefore, these conventional CMAB-based algorithms are of low efficiency especially when the number of arms is huge while the total budget is limited. Moreover, although the uncertainty of the workers is considered, they do not take into account the incentive issue.

  The CMAB framework also can be applied to design incentive mechanisms for crowdsensing systems where the workers' sensing abilities are unknown. For example, \cite{BiswasJMN-AAMAS15,XiaoWZG-ICDCS20,GaoHXWSZ-INFOCOM21} propose incentive mechanisms for unknown worker recruitment with limited budget. Unfortunately, all these existing proposals adopt the CMAB framework to enable an exploration-exploitation trade-off among the individual workers and their performance is degraded when the number of the unknown workers is huge whereas the budget is limited, as will be shown in Sec.~\ref{ssec:offregret} and Sec.~\ref{sec:exp}.

  Context information is useful for crowdsensing systems and has been extensively utilized in designing worker selection algorithms \cite{YururLSLML-COMST16,NejadAMM-FGCS19}. For example, in \cite{HassaniHJ-ICPADS15}, sensing tasks are assigned to workers with similar context to them, considering a worker may be more eligible if its context is more similar to the one of the sensing tasks. The matching between sensing tasks and workers is also studied in \cite{YucelYB-TMC2021} where the requirements of tasks and the preferences of workers are considered. \cite{HanYYWYB-TMC23} proposes novel data structures to improve the performance of the task-worker matching. However, the uncertainties of the workers are not considered in these proposals. In \cite{LiuZWTC-INFOCOM17}, a context-based data quality classifier trained in an off-line manner is used to identify eligible workers. \cite{MullerTSK-TON18} employs the contextual MAB approach to understand how workers' sensing abilities depend on both the workers' and the tasks' context information. Although these proposals adopt machine learning techniques to select unknown workers according to context information, they do not consider the strategics of workers. In \cite{WuLMXLW-IOTJ19}, a context-aware MAB-based incentive mechanism is designed. However, no budget constraint is taken into account.

%\vspace{-4ex}
\section{System Model and Problem Description} \label{sec:sys}
  In Sec.~\ref{ssec:offmodel}, we first introduce our system model for the off-line case where a fixed group of workers are given. We then formulate our incentive problem in Sec.~\ref{ssec:offprobform}, and finally extend the system model and problem formulation to the on-line case with dynamic workers in Sec.~\ref{ssec:onlinemodel}. 
  \vspace{-2ex}
  \subsection{Off-Line System Model} \label{ssec:offmodel}
    Given a sensing task with budget $B$ and a fixed group of $N$ workers $\mathcal N = \{1,2,\cdots,N\}$, each worker $i \in \mathcal N$ first privately communicates its bid $b_i \in [b_{min}, b_{max}]$ to a task auctioneer (or assigner). Then, the task assigner decides which workers to select and how to pay the selected workers. We consider a time span $\mathcal T = \{1, 2, \cdots, T\}$. In each time slot $t \in \mathcal{T}$, a selected worker collects and reports one data sample. We suppose $r^{[t]}_i \in \{0, 1\}$ denotes an i.i.d. Bernoulli random variable indicating if the data collected by worker $i$ in time slot $t$ is qualified. It also can be explained as the reward we obtain by selecting worker $i$ to conduct the sensing task in time slot $t$. For each worker $ i \in \mathcal N$, we assume $\mu_i = \mathbb E [ r^{[t]}_i ]$ for $t=1,2,\cdots$, which actually represents worker $i$'s sensing ability (or quality) and thus the expected reward we can obtain by recruiting worker $i$. We assume $\mu_i$ is unknown for the task assigner and let $\mathbf{m} = (\mu_1, \mu_2, \cdots, \mu_N)$ be the sensing ability vector. We also let $\mathbf{b} = (b_1, b_2, \cdots, b_N)$ denotes the bid vector.

    Although the workers' sensing abilities are unknowns, we assume that, for each worker $i \in \mathcal N$, its context $s_i \in \mathcal S$ can be observed, where $\mathcal S \triangleq [0,1]^{M}$ is so-called ``\textit{context space}'' and $M$ is the number of the dimensions of the context space. The context dimensions include the proficiency of the workers in some required skills, the personal backgrounds of the workers and the performance parameters of their sensor devices etc., which are closely related to the workers' sensing abilities \cite{MullerTSK-TON18,HanYYWYB-TMC23}. Therefore, we can derive the workers' sensing abilities according to their contexts. In particular, we denote by $r: \mathcal S \rightarrow \{0,1\}$ a stochastic reward function, such that for any selected worker $i$ with specific context $s_i \in \mathcal S$, we have $r(s_i) = 1$ if the data sample it reports is qualified; otherwise, $r(s_i) = 0$. The sensing ability of worker $i$ with context $s_i$ is then defined by $\mu(s_i) = \mathbb E[r(s_i)]$. In another word, for any worker $i\in \mathcal N$ selected in slot $t$, we have $\mu_i = \mathbb E [ r^{[t]}_i ] = \mathbb E [ r(s_i) ]$. We let $\mu_{max} = \max_{s\in \mathcal S} \mu(s)$ and $\mu_{min} = \min_{s\in \mathcal S} \mu(s)$ and denote by $\mathbf{s} = (s_1, s_2, \cdots, s_N)$ the context vector of the workers $\mathcal{N}$.

    We denote by $x^{[t]}_{i} \in \{0,1\}$ an indicator variable specifying if worker $i$ is selected in time slot $t$. To address potential noise in the data collected from a single worker, especially when the sensing abilities of the workers are unknown, it is necessitated to enlist a minimum number of workers to enhance sensing robustness~\cite{HeSZC-INFOCOM14,HanZL-TON16,GaoWXC-INFOCOM20}. Hence, in each time slot $t$ (with sufficient residual budget), we suppose the task assigner selects at least $K$ workers for the purpose of robustness. Specifically, in any time slot $t$, we let $\mathsf{x}^{[t]}_{\mathsf{off}} = ( x^{[t]}_{1}, \cdots, x^{[t]}_{N} )$ denote the selection strategy such that $\sum_{i \in \mathcal{N}} x^{[t]}_{i} \geq K$. We denote by $\mathsf{X}_{\mathsf{off}} = ( \mathsf{x}^{[1]}_{\mathsf{off}}, \mathsf{x}^{[2]}_{\mathsf{off}}, \cdots )$ the selection policy across time span $\mathcal{T}$.

    Let $p^{[t]}_{i} \in [b_{min}, b_{max}]$ represent the payment to worker $i$ in time slot $t$, $\mathsf{P}^{[t]}_{\mathsf{off}} = ( {p}^{[t]}_1, \cdots, {p}^{[t]}_N )$ denote the payment strategy in time slot $t$, and $\mathsf{P}_{\mathsf{off}} = ( \mathsf{P}^{[1]}_{\mathsf{off}}, \mathsf{P}^{[2]}_{\mathsf{off}}, \cdots )$ be the payment policy across the $T$ time slots. We also let $c_i \in [c_{min}, c_{max}]$ be the true cost of worker $i$ to conduct the sensing task and let $\mathbf{c} = (c_1, \cdots, c_N)$ denote the vector of the true costs. Note that bid $b_i$ is the cost reported by worker $i$, and it may be different from worker $i$'s true cost $c_i$. For example, a strategic worker may communicate a bid different from its true cost to earn more. We define the \textit{utility} of worker $i$ in time slot $t$ as
    \begin{equation} \label{eq:utility}
      \phi^{[t]}_{i} \left( x^{[t]}_i, p^{[t]}_{i} \right)  = x^{[t]}_i \left( p^{[t]}_{i} - c_i \right)
    \end{equation}
    Our payment policy $\mathsf{P}_{\mathsf{off}}$ should be carefully designed such that each worker has to communicate its bid truthfully. Let $\mathbf{b}_{-i} = (b_1, \cdots, b_{i-1}, b_{i+1}, \cdots, b_{N})$ be the bid vector excluding $b_i$. Given any $\mathbf{b}_{-i} \in [b_{min}, b_{max}]^{N-1}$, we denote by ${\phi}^{[t]}_{i} \left(x^{[t]}_i,  p^{[t]}_i; b_i, \mathbf{b}_{-i} \right)$ the utility obtained by worker $i$ when worker $i$ submits bid $b_i$ and the others submit $\mathbf{b}_{-i}$. The truthfulness of a mechanism can be defined as \textbf{Definition}~\ref{def:truth}.
    \begin{definition}(Truthfulness) \label{def:truth}
      A mechanism is said to be truthful, if for any worker $i$, bidding true cost always maximizes its utility irrespective of the bids of the other workers. Specifically, a mechanism is truthful, if
      \begin{equation}  \label{eq:truth}
        {\phi}^{[t]}_i \left( x^{[t]}_i, p^{[t]}_i; c_i, \mathbf{b}_{-i} \right) \geq {\phi}^{[t]}_i \left( x^{[t]}_i, p^{[t]}_i; b_i, \mathbf{b}_{-i} \right)
      \end{equation}
      holds for $\forall i\in \mathcal N$, $\forall t \in \mathcal T$, $\forall b_i \in [b_{min}, b_{max}]$ and $\forall \mathbf{b}_{-i} \in [b_{min}, b_{max}]^{N-1}$.
    \end{definition}
    \noindent If a worker could not increase its utility by communicating an untruthful bid regardless of the bids of the others, bidding truthfully (i.e., $b_i = c_i$) would be its best choice. If the truthfulness of our mechanism is ensured, we have $b_i = c_i$ for $\forall i$ and thus $b_{min} = c_{min}$ and $b_{max} = c_{max}$.

    Our another concern is individual rationality.
    \begin{definition}(Individual Rationality) \label{def:ration}
      A mechanism is said to be individually rational, if each worker has a non-negative utility by participating in the auction
      \begin{equation} \label{eq:ration}
        {\phi}^{[t]}_i \left( {x}^{[t]}_i, {p}^{[t]}_i \right) \geq 0, ~\forall i\in\mathcal N, \forall t \in \mathcal T
      \end{equation}
    \end{definition}
    %
    %Our payment policy $\mathsf{P}_{\mathsf{off}}$ is supposed to guarantee that worker gets non-negative utility by participating in the sensing task.

  \subsection{Off-Line Problem Formulation} \label{ssec:offprobform} 
    Our off-line \textit{budget-limited $K$-Worker INcentive} ($K$-WIN) problem can be formulated as shown in $\mathfrak{P}_{\mathsf{off}}$ (see (\ref{eq:offline-obj})$\sim$(\ref{eq:offline-con-var})). Specifically, we design mechanism $\mathsf{F_{off}} = (\mathsf{X_{off}}, \mathsf{P_{off}})$ such that the objective function (\ref{eq:offline-obj}) (i.e., the expected cumulative sensing reward $R(\mathsf{F_{off}})$ obtained through our mechanism) is maximized with the  constraints (\ref{eq:offline-con-k})$\sim$(\ref{eq:offline-con-var}) satisfied.
    \begin{align}
      \mathfrak{P}_{\mathsf{off}}: ~\max& ~R \left( \mathsf{F_{off}} \right) = \mathbb E \left[ \sum^T_{t=1}\sum_{i\in \mathcal N}  x^{[t]}_{i} r^{[t]}_{i} \right]  \label{eq:offline-obj}\\
      \mathrm{s.t.}& ~\sum_{i\in \mathcal N} x^{[t]}_i \geq K \cdot z^{[t]}, ~\forall t \in \mathcal{T}   \label{eq:offline-con-k}\\
      & ~ \sum^T_{t=1} \sum_{i\in \mathcal N} p^{[t]}_i \leq B  \label{eq:offline-con-budget}\\
      & ~ x^{[t]}_i \leq  z^{[t]}, ~\forall i \in \mathcal{N}, t \in \mathcal{T}  \label{eq:offline-con-feas-1}\\
      & ~ z^{[t]} \geq z^{[t+1]}, ~\forall t \in \mathcal{T}  \label{eq:offline-con-feas-2}\\
      & ~ \phi^{[t]}_i \left( x^{[t]}_i, p^{[t]}_i; c_i, \mathbf{b}_{-i} \right) \geq {\phi}^{[t]}_i \left( x^{[t]}_i, p^{[t]}_i; b_i, \mathbf{b}_{-i} \right) \nonumber\\
      & ~ \forall i \in \mathcal{N}, \forall t \in \mathcal{T}, \forall b_i \in [b_{min}, b_{max}], \nonumber\\
      & ~ \forall \mathbf{b}_{-i} \in [b_{min}, b_{max}]^{N-1}  \label{eq:offline-con-true}\\
      & ~ \phi^{[t]}_i \left( x^{[t]}_i, p^{[t]}_i \right) \geq 0, ~\forall i \in \mathcal{N}, t \in \mathcal{T}  \label{eq:offline-con-rat}\\
      & ~ x^{[t]}_i, z^{[t]} \in \{0,1\}, ~\forall i \in \mathcal{N}, t \in \mathcal{T} \label{eq:offline-con-var}
    \end{align}
    We provide explanations to the constraints as follows.
    \begin{itemize}
      \item \textbf{Robustness constraint} (\ref{eq:offline-con-k}): In each time slot $t$, we have at least $K$ workers selected for the purpose of robustness. Since we may not have sufficient budget to recruit workers in each time slot $t$, we introduce $z^{[t]} \in \{0,1\}$ to indicate whether we recruit $K$ workers in time slot $t$.
      \item \textbf{Budget constraint} (\ref{eq:offline-con-budget}). The total payment across the $T$ time slots does not exceed budget $B$.
      \item \textbf{Feasibility constraints} (\ref{eq:offline-con-feas-1})$\sim$(\ref{eq:offline-con-feas-2}). As shown by (\ref{eq:offline-con-feas-1}), the workers are selected only when the sensing task is conducted. Additionally, we introduce constraint (\ref{eq:offline-con-feas-2}) to ensure the solvability of our problem by confining its dimension without sacrificing generality; it is assumed that we successively select workers across time span $\mathcal{T}$.
      \item \textbf{Truthfulness} (\ref{eq:offline-con-true}). Each worker $i$ has to communicate its private bid truthfully (see \textbf{Definition}~\ref{def:truth}).
      \item \textbf{Individual rationality} (\ref{eq:offline-con-rat}). Our mechanism has to be individually rational (see \textbf{Definition}~\ref{def:ration}).
    \end{itemize}

    If the costs of the workers were not private information, our $K$-WIN problem $\mathfrak{P}_{\mathsf{off}}$ would be reduced to the knapsack problem which is NP-hard, and we thus should be content with approximation algorithms where the quality-to-cost ratios of the workers are usually utilized. The algorithms for the classic knapsack problems actually provide inspirations to us to design incentive mechanisms for our $K$-WIN problem especially when the sensing abilities of the workers, i.e., $\mathbf m = \{\mu_i\}_{i \in \mathcal N}$, are assumed to be known. As will be shown in Sec.~\ref{sec:baseline}, one possible incentive solution to the off-line $K$-WIN problem is that, we first sort the workers in decreasing order according to $\rho_i = {\mu_i}/{b_i}$; we then select the top-$K$ workers $\widetilde{\mathcal{N}} \subseteq \mathcal{N}$ and pay ${\mu_i}/{\rho_{K+1}}$ to each of $\widetilde{\mathcal{N}} $ according to the principle of second price auction in each time slot, until the total budget is exhausted. In this baseline mechanism where the workers' sensing abilities are known, we still have to incentivize the workers to bid truthfully~\footnote{We will introduce the detail of the baseline mechanism and analyze its truthfulness and individual rationality in Sec.~\ref{sec:baseline}. }. Unfortunately, as mentioned in Sec.~\ref{sec:intro}, $\mathbf m$ is not always available in practice. When $\mathbf m$ is unknown, one popular choice to address the $K$-WIN problem is to utilize the CMAB framework~\cite{BiswasJMN-AAMAS15,XiaoWZG-ICDCS20,GaoHXWSZ-INFOCOM21}. Nevertheless, the standard CMAB framework leverages the exploitation-exploration trade-off among the unknown workers individually, the expense of learning (i.e., the cost to recruit the workers) may not be affordable, especially when the number of workers is huge whereas the given budget is rather limited.

  \vspace{-4ex}
  \subsection{On-Line Problem Formulation} \label{ssec:onlinemodel}
    We now consider an on-line scenario where workers may join in or depart from the crowdsensing platform in an on-line manner. Let $\mathcal{N}^{[t]}$ denote the set of ${N}^{[t]}$ available worker in time slot $t$. In time slot $t$, each worker $i \in \mathcal{N}^{[t]}$ first privately communicates its bid $b_i$ to task assigner; then the assigner decides which workers to recruit and how much to pay them. Suppose $\mathbf{b}^{[t]} = (b_1, b_2, \cdots, b_{N^{[t]}})$ denotes the bid vector of the workers $\mathcal{N}^{[t]}$. Similarly, let $x^{[t]}_i \in \{0,1\}$ denote the binary variable indicating if worker $i$ is selected and $p^{[t]}_i$ represent the payment to the worker $i$. Then, our on-line selection policy and payment policy in time slot $t$ can be represented by $\mathsf{X}^{[t]}_{\mathsf{on}} = ( x^{[t]}_i )_{i \in \mathcal{N}^{[t]}}$ and $\mathsf{P}^{[t]}_\mathsf{on} = ( p^{[t]}_i )_{i \in \mathcal{N}^{[t]}}$, respectively.
    We formulate our \textit{on-line} $K$-WIN problem as follows
    \begin{align}
      \mathfrak{P}_{\mathsf{on}}: ~\max& ~R (\mathsf{F_{on}}) = \mathbb E \left[ \sum^T_{t=1}\sum_{i\in \mathcal{N}^{[t]}}  x^{[t]}_{i} r^{[t]}_{i} \right]  \label{eq:online-obj}\\
      \mathrm{s.t.}& ~\sum_{i\in \mathcal{N}^{[t]}} x^{[t]}_i \geq K \cdot z^{[t]}, ~\forall t \in \mathcal{T}   \label{eq:online-con-k}\\
      & ~ \sum^T_{t=1} \sum_{i \in \mathcal{N}^{[t]}} p^{[t]}_i \leq B  \label{eq:online-con-budget}\\
      & ~ x^{[t]}_i \leq  z^{[t]}, ~\forall i \in \mathcal{N}^{[t]}, t \in \mathcal{T}  \label{eq:online-con-feas-1}\\
      & ~ z^{[t]} \geq z^{[t+1]}, ~\forall t \in \mathcal{T}  \label{eq:online-con-feas-2}\\
      & ~ \phi^{[t]}_i \left( x^{[t]}_i, p^{[t]}_i; c_i, \mathbf{b}_{-i} \right) \geq {\phi}^{[t]}_i \left( x^{[t]}_i, p^{[t]}_i; b_i, \mathbf{b}_{-i} \right) \nonumber\\
      & ~ \forall i \in \mathcal{N}^{[t]}, \forall t \in \mathcal{T}, \forall b_i \in [b_{min}, b_{max}], \nonumber\\
      & ~ \forall \mathbf{b}_{-i} \in [b_{min}, b_{max}]^{N-1}  \label{eq:online-con-true}\\
      & ~ \phi^{[t]}_i \left( x^{[t]}_i, p^{[t]}_i \right) \geq 0, ~\forall i \in \mathcal{N}^{[t]}, t \in \mathcal{T}  \label{eq:online-con-rat}\\
      & ~ x^{[t]}_i, z^{[t]} \in \{0,1\}, ~\forall i \in \mathcal{N}^{[t]}, t \in \mathcal{T} \label{eq:online-con-var}
    \end{align}
    where $\mathsf{F_{on}} = ( \mathsf{X}^{[t]}_{\mathsf{on}}, \mathsf{P}^{[t]}_\mathsf{on} )_{t \in \mathcal T}$ denotes our on-line selection-and-payment policy across the time span $\mathcal T$. As shown in the above formulation, the objective function and the constraints in $\mathfrak{P}_{\mathsf{on}}$ actually are very similar to the ones in $\mathfrak{P}_{\mathsf{off}}$. We re-define them by taking into account the dynamic arrival and departure of the on-line workers. Specifically, in each time slot $t$, we select at least $K$ workers from the given available workers $\mathcal N^{[t]}$ to ensure the sensing robustness, and the truthfulness and the individual rationality should be guaranteed for each of $\mathcal N^{[t]}$.

    Likewise, if the costs of the workers were not private information, the on-line $K$-WIN problem would be reduced to a series of knapsack problems across different time slots. Therefore, in light of the reduction, the basic idea of the baseline mechanism for our off-line $K$-WIN problem can be extended to address the on-line version, as will be shown in Sec.~\ref{sec:online-caci}. Furthermore, in the above on-line $K$-WIN problem, when the sensing abilities of the workers in each time slot are unknown, the challenges are two-fold as mentioned in Sec.~\ref{sec:intro}: on one hand, similar to its off-line counterpart, learning the workers' sensing abilities is confined by limited budget; on the other hand, even with sufficient budget, since the available workers may be different over time, the knowledge about the sensing abilities of unavailable workers could not be applicable for learning the ones of available workers.

\vspace{-2ex}
\section{Baseline Mechanisms} \label{sec:baseline}
  In this section, we introduce baseline mechanisms for the above two optimization problems $\mathfrak{P}_{\mathsf{off}}$ and $\mathfrak{P}_{\mathsf{on}}$, respectively. In fact, The baseline mechanisms not only inspire the design of our on-line/off-line CACI mechanisms, but also can be used to evaluate the CACI mechanisms mainly from theoretical perspective. In the following, we first introduce the design of the off-line baseline mechanism in Sec.~\ref{ssec:bl-mechdesign}, where a fixed group of workers are given and their sensing abilities are known. We then analyze its truthfulness and individual rationality in Sec.~\ref{ssec:bl-truth}. We finally show in Sec~\ref{ssec:extonline} how to extend the basic idea of the off-line mechanism to address the on-line $K$-WIN problem.

  \subsection{Off-Line Mechanism Design} \label{ssec:bl-mechdesign}
    To inspire the design of our off-line and on-line CACI mechanisms, we first consider a baseline mechanism $\mathsf{F}^*(\mathbf{m}, \mathbf{b}, B)$, to address the $K$-WIN problem $\mathfrak{P}_{\mathsf{off}}$ (see (\ref{eq:offline-obj})$\sim$(\ref{eq:offline-con-var})) with the workers' sensing abilities $\mathbf{m} = (\mu_i)_{i \in \mathcal N}$ known as prior knowledge. The pseudo-code of the baseline mechanism is given in \textbf{Algorithm}~\ref{alg:baseline}. We first sort the workers in $\mathcal N$ in decreasing order according to its \textit{Quality-to-Bid Ratio} (QBR) $\rho^*_i = {\mu_i}/{b_i}$ such that $\rho^*_1 \geq \rho^*_i \geq \cdots \geq \rho^*_N$ (see Line~\ref{ln:bl-sort}) and let $\widetilde{\mathcal N}^*$ be the set of top-$K$ workers in the ordered $\mathcal N$ (see Line~\ref{ln:bl-topK}). For $\forall i \in \widetilde{\mathcal N}^*$, we calculate the payment as follows
    \begin{equation}  \label{eq:bl-payment}
      \tilde{p}^*_i = \min \left\{ \frac{\mu_i}{\rho^*_{K+1}}, b_{max} \right\}
    \end{equation}
    as illustrated in Line~\ref{ln:bl-payment}. We denote by $B^{[t]}$ the residual budget in time slot $t$, which are initialized by $B^{[t]} = B$ (see Lines~\ref{ln:bl-initslot}$\sim$\ref{ln:bl-initbudget}). In each time slot, if there is sufficient residual budget (i.e., $B^{[t]} \geq \sum_{i \in \widetilde{\mathcal N}^*} \tilde{p}^*_i$),
    %
    %\footnote{We hereby terminate our mechanism when the residual budget is not sufficient for us to go through all the candidate workers $\widetilde{\mathcal N}^*$, so as to facilitate our theoretic analysis. In fact, we can relax the condition on the residual budget such that our mechanism continues until we cannot select any worker in $\widetilde{\mathcal N}^*$ using the residual budget, and all our theoretical results still hold if we take this relaxation.}, 
    %
    we select the $K$ workers $\widetilde{\mathcal N}^*$ to conduct the sensing task such that $x^{[t]}_i = 1$ for $\forall i \in \widetilde{\mathcal N}^*$ and $x^{[t]}_i = 0$ for the others (see Line~\ref{ln:bl-select}), and we pay each worker $i \in \widetilde{\mathcal N}^*$ such that $p^{[t]}_i = \tilde{p}^*_i$ for for $\forall i \in \widetilde{\mathcal N}^*$ and $p^{[t]}_i = 0$ for the others (see Line~\ref{ln:bl-pay}). We finally update $B^{[t+1]}$ accordingly (see Line~\ref{ln:bl-budget}) and move on to the next time slot (see Line~\ref{ln:bl-update}).
    \begin{algorithm}[htb!]
      \KwIn{Workers $\mathcal N$, sensing ability vector $\mathbf{m}$, bid vector $\mathbf b$, and total budget $B$.}
      \KwOut{Selection policy $\mathsf{X}^*_{\mathsf{off}}$ and payment policy $\mathsf{P}^*_{\mathsf{off}}$.}
      %
      %$\rhd$ \textit{Initialization phase}: \\
      %
      %$\rhd$ \textit{Selection-and-payment phase}: \\
      %
      Sort the workers in $\mathcal N$ in decreasing order according to $\rho^*_i = {\mu_i}/{b_i}$ such that $\rho^*_1 \geq \rho^*_i \geq \cdots \geq \rho^*_N$;  \label{ln:bl-sort}\\
      Let $\widetilde{\mathcal N}^* \subseteq \mathcal{N}$ be the set of top-$K$ workers;  \label{ln:bl-topK} \\
      Let $\tilde{p}^*_i = \min \left\{ {\mu_i}/{\rho^*_{K+1}}, b_{max} \right\}$ for $\forall i \in \widetilde{\mathcal N}^*$;  \label{ln:bl-payment}\\
      $t=1$; \label{ln:bl-initslot}\\
      $B^{[t]} = B$; \label{ln:bl-initbudget}\\
      \While{$B^{[t]} \geq \sum_{i \in \widetilde{\mathcal N}^*} \tilde{p}^*_i$}{  \label{ln:bl-start}
        Select $\widetilde{\mathcal N}^*$ to conduct the sensing task such that $x^{[t]}_i = 1$ for $\forall i \in \widetilde{\mathcal N}^*$ and $0$ for $\forall i \in \mathcal{N} / \widetilde{\mathcal{N}}^*$; \label{ln:bl-select}\\
        Pay the selected workers $\widetilde{\mathcal N}^*$ such that $p^{[t]}_i = \tilde{p}^*_i$ for $\forall i \in \widetilde{\mathcal N}^*$ and $0$ for $\forall i \in \mathcal{N} / \widetilde{\mathcal{N}}^*$; \label{ln:bl-pay}\\
        $B^{[t+1]} = B^{[t]} - \sum_{i \in \widetilde{\mathcal N}^*} \tilde{p}^*_i$; \label{ln:bl-budget}\\
        $t = t+1$;  \label{ln:bl-update}\\
      } \label{ln:bl-end}
    \caption{A baseline mechanism $\mathsf{F}^*_{\mathsf{off}}$} 
    \label{alg:baseline}
    \end{algorithm}
    %\vspace{-2ex}  

  \vspace{-4ex}
  \subsection{Truthfulness and Individual Rationality} \label{ssec:bl-truth}
    According to \textbf{Algorithm}~\ref{alg:baseline}, it is apparent that both the budget constraint and the feasibility constraint are respected in the baseline mechanism. Furthermore, we ensure the truthfulness and the individual rationality, as shown in \textbf{Proposition}~\ref{prop:bl-truthfulness} and \textbf{Proposition}~\ref{prop:bl-rationality}, respectively. Due to the space limit, the proofs are given in the supplementary material.
    \begin{proposition} \label{prop:bl-truthfulness}
      Our off-line baseline mechanism is truthful.
    \end{proposition}
    \begin{proposition} \label{prop:bl-rationality}
    %\vspace{-1ex}
      Our off-line baseline mechanism is individually rational.
    \end{proposition}

    As shown above, the baseline mechanism with $\mathbf{m} = \{\mu_i\}_{i \in \mathcal N}$ known as prior knowledge has its truthfulness and individual rationality ensured. Furthermore, according to \cite{Lawler-SFCS77,Vazirani-book01}, it is actually a $2$-approximation algorithm. Unfortunately, it is usually very difficult to pre-profile a huge number of workers; therefore, $\mathbf{m}$ may not always be available as known, which makes our problem much more difficult. Although some state-of-the-art proposals apply the CMAB framework to tackle the uncertainty of the workers' sensing abilities~\cite{BiswasJMN-AAMAS15,XiaoWZG-ICDCS20,XiaoAWGZW-TMC21}, they leverage the exploration-exploitation trade-off among the individual workers, resulting in significant overhead. For example, the total budget even may not be sufficient to explore each of the workers once with the truthfulness and individual rationality guaranteed.

  \subsection{Extension to On-Line Workers} \label{ssec:extonline}
    It is worthy to note that the baseline mechanism can be extended to address the on-line $K$-WIN problem $\mathfrak{P}_{\mathsf{on}}$ with the on-line workers' sensing abilities known as prior knowledge. The pseudo-code of our on-line baseline mechanism $\mathsf{F}^*_{\mathsf{on}}$ is given in \textbf{Algorithm}~\ref{alg:onbaseline}. Specifically, in each time slot $t$, we first sort the workers available in time slot $t$, i.e., ${\mathcal N}^{[t]}$, in decreasing order according to $\rho^* = \mu_i / b_i$ such that $\rho^*_1 \geq \rho^*_2 \geq \cdots \geq \rho^*_{N^{[t]}}$ (see Line~\ref{ln:onbl-sort}). We let $\widetilde{\mathcal N}^{*[t]} \subseteq \mathcal{N}^{[t]}$ be the set of top-$K$ workers, and calculate the payment $\tilde{p}^*_i$ for each worker $i \in \widetilde{\mathcal N}^{*[t]}$, as illustrated in Lines~\ref{ln:onbl-topK} and \ref{ln:onbl-payment}. If there is sufficient budget to pay $\widetilde{\mathcal N}^{*[t]}$, we select and pay the workers $\widetilde{\mathcal N}^{*[t]}$ to conduct the sensing task (see Lines~\ref{ln:onbl-select} and \ref{ln:onbl-pay}), and update the residual budget $B^{[t+1]} = B^{[t]} - \sum_{i \in \widetilde{\mathcal N}^{*[t]}} p^{[t]}_i$ (see Line~\ref{ln:onbl-budget}).
    \begin{algorithm}[htb!]
      \KwIn{On-line workers $\mathcal{N}^{[t]}$, sensing ability vector $\mathbf{m}^{[t]}$ and bid vector $\mathbf{b}^{[t]}$ with $t=1,2,\cdots,T$, and total budget $B$.}
      \KwOut{Selection policy $\mathsf{X}^*_{\mathsf{on}}$ and payment policy $\mathsf{P}^*_{\mathsf{on}}$.}
      %
      %$\rhd$ \textit{Initialization phase}: \\
      %
      $B^{[1]} = B$; \\
      %
      %$\rhd$ \textit{Selection-and-payment phase}: \\
      %
      \For{$t=1,2,\cdots,T$}{
        Sort workers $\mathcal{N}^{[t]}$ in decreasing order according to $\rho^*_i = {\mu_i}/{b_i}$, such that $\rho^*_1 \geq \rho^*_i \geq \cdots \geq \rho^*_{N^{[t]}}$;  \label{ln:onbl-sort}\\
        Let $\widetilde{\mathcal N}^{*[t]} \subseteq \mathcal{N}^{[t]}$ be the set of top-$K$ workers;  \label{ln:onbl-topK} \\
        Let $\tilde{p}^*_i = \min \left\{ {\mu_i}/{\rho^*_{K+1}}, b_{max} \right\}$ for $\forall i \in \widetilde{\mathcal N}^{*[t]}$;  \label{ln:onbl-payment}\\
        \If{$B^{[t]} \geq \sum_{i \in \widetilde{\mathcal N}^{*[t]}} p^{[t]}_i$}{
          Select $\widetilde{\mathcal N}^{*[t]}$ to conduct the sensing task such that $x^{[t]}_i = 1$ for $\forall i \in \widetilde{\mathcal N}^{*[t]}$ and $x^{[t]}_i = 0$ for $\forall i \in \mathcal{N}^{[t]} / \widetilde{\mathcal{N}}^{*[t]}$; \label{ln:onbl-select}\\
          Pay the selected workers $\widetilde{\mathcal N}^{*[t]}$ such that $p^{[t]}_i = \widetilde{p}^*_i$ for $\forall i \in \widetilde{\mathcal{N}}^{*[t]}$ and $p^{[t]}_i = 0$ for $\forall i \in \mathcal{N}^{[t]} / \widetilde{\mathcal{N}}^{*[t]}$; \label{ln:onbl-pay}\\
          $B^{[t+1]} = B^{[t]} - \sum_{i \in \widetilde{\mathcal N}^{*[t]}} p^{[t]}_i$; \label{ln:onbl-budget}\\
        }
      }
    \caption{An on-line baseline mechanism $\mathsf{F}^*_{\mathsf{on}}$} 
    \label{alg:onbaseline}
    \end{algorithm}

    In fact, the above on-line baseline mechanism is ``similar'' to its off-line counterpart given in \textbf{Algorithm}~\ref{alg:baseline}. In each time slot, it selects $K$ individuals with higher QBRs from the given available workers and pays them according to the second price principle. Therefore, the on-line mechanism inherits the optimality as well as the truthfulness and individual rationality of its off-line counterpart. Due to the space limit, we hereby do not present the detailed proof. Nevertheless, when the sensing abilities of the on-line workers are unknown, the standard CMAB framework is not feasible even if we have sufficient budget, since it learns the sensing abilities of the workers individually, and we cannot estimate the sensing ability of a new worker according to what we have learnt for the others, as mentioned in Sec.~\ref{sec:intro}. As will be shown in Sec.~\ref{sec:online-caci}, the context information of the workers is utilized in our on-line mechanism. By learning the mapping between context space and sensing ability, given any worker in any time slot, we can accurately estimate its sensing ability according to its context, even the worker never appears theretofore.

\section{Our Off-Line CACI Mechanism} \label{sec:offline}
  In this section, we first introduce the details of our off-line CACI mechanism in Sec.~\ref{ssec:offmechdesign}. Using the above off-line baseline mechanism as reference, we reveal the upper bound on the difference between the baseline mechanism and our off-line CACI mechanism in terms of the expected cumulative sensing revenue in Sec.~\ref{ssec:offregret}. We finally discuss about the truthfulness and individual rationality in Sec.~\ref{ssec:offtruth}.

  \vspace{-2ex}
  \subsection{CACI Mechanism Design} \label{ssec:offmechdesign}
    As mentioned in Sec.~\ref{sec:intro}, our mechanism is motivated by the fact that workers with similar contexts usually have similar sensing abilities. Hence, we first divide the context space $\mathcal S$ into a group of $d^M$ disjoint hypercubes where $d$ represents the granularity of partitioning \footnote{We will explain the definition of $d$ in detail in Sec.~\ref{ssec:offregret}}. The hypercubes have identical size ${1}/{d^M}$. We denote by $\mathcal Q$ the set of hypercubes and let $Q(i)$ represent the hypercube such that $s_i \in Q(i)$. If worker $i$ is selected in time slot $t$, it is also said that $Q(i)$ is selected in time slot $t$. With a fine tuned $d$, the workers in the same hypercube have similar sensing abilities, and we estimate the sensing ability of any worker $i$ as the one of $Q(i)$. The idea of our mechanism is to learn the ``sensing abilities'' of the hypercubes by leveraging the exploration-exploitation trade-off among them and estimate the ones of the unknown workers according to their contexts.
    \begin{algorithm}[htb!]
      \KwIn{Unknown workers $\mathcal N$ with context vector $\mathbf{s}$ and bid vector $\mathbf{b}$, and total budget $B$.}
      \KwOut{Selection policy $\mathsf{X_{off}}$ and payment policy $\mathsf{P_{off}}$.}
      $\rhd$ \textit{Initialization}: \\
      $t=0$; \label{ln:off-init-slot}\\
      $B^{[t+1]} = B$; \label{ln:off-init-budget}\\
      $\bar{r}^{[t]}(Q) = 0$ and $\lambda^{[t]}(Q) = 0$ for $\forall Q \in \mathcal{Q}$; \label{ln:off-init-rlambda}\\
      $\rhd$ \textit{Exploration phase}: \\
      ${B}^\# = \left( \frac{b_{max}}{\mu^2_{max}} \right)^\frac{1}{3} \hspace{-1ex} d^\frac{M}{3} B^\frac{2}{3} \ln^\frac{1}{3} B $; \label{ln:off-explore-budget}\\
      \For{$t = 1, 2, \cdots, \left\lfloor \frac{B^\#}{K b_{max}} \right\rfloor$}{  \label{ln:off-explore-start}
        $\widetilde{\mathcal{N}}^{[t]} = \emptyset$; \label{ln:off-explore-init}\\
        \For{$k = 1, 2, \cdots, K$}{ \label{ln:off-explore-comp-n-1}
          $\ell = ((t-1)K + k)~\mathrm{mod}~d^M$; \\
          Uniformly choose a worker (e.g. worker $i$) in the $\ell$-th hypercube; \\
          $\widetilde{\mathcal{N}}^{[t]} \leftarrow \widetilde{\mathcal{N}}^{[t]} \bigcup \{i\}$; \\
        }  \label{ln:off-explore-comp-n-2}
        Select workers $\widetilde{\mathcal{N}}^{[t]}$ to conduct the sensing task such that $x^{[t]}_i = 1$ for $\forall i \in \widetilde{\mathcal{N}}^{[t]}$ and $x^{[t]}_i = 0$ for $\forall i \in \mathcal{N} / \widetilde{\mathcal{N}}^{[t]}$; \label{ln:off-explore-select}\\
        Pay workers $\widetilde{\mathcal{N}}^{[t]}$ such that $p^{[t]}_i = b_{max}$ for $\forall i \in \widetilde{\mathcal{N}}^{[t]}$ and $p^{[t]}_i = 0$ for $\forall i \in \mathcal{N} / \widetilde{\mathcal{N}}^{[t]}$;  \label{ln:explore-pay}  \label{ln:off-explore-pay}\\
        Observe the reward feedback $r^{[t]}_i$ for $\forall i \in \widetilde{\mathcal{N}}^{[t]}$; \label{ln:off-explore-reward}\\
        Update $\lambda^{[t]}(Q)$ and $\bar{r}^{[t]}(Q)$ for $\forall Q \in \mathcal{Q}$ according to (\ref{eq:offcaci-uplambda}) and (\ref{eq:offcaci-upavgr}), respectively;   \label{ln:off-explore-update-lambda-r}\\
        $B^{[t+1]} = B^{[t]} - K \cdot b_{max}$; \label{ln:off-explore-upbudget}\\
      } \label{ln:off-explore-end}
      $\rhd$ \textit{Exploitation phase}:\\
      $t \leftarrow t+1$; \label{ln:off-exploit-start}\\
      Let $u_i = \bar r^{[t-1]}(Q(i)) + \sqrt{\frac{\ln B}{\lambda^{[t-1]}(Q(i))}}$ for $\forall i\in \mathcal N$; \label{ln:off-exploit-ucb}\\
      Sort workers $\mathcal N$ in decreasing order according to $\rho_i = {u_i}/{b_i}$ such that $\rho_1 \geq \rho_2 \geq \cdots \geq \rho_N$; \label{ln:sort}  \label{ln:off-explore-sort}\\
      Let $\widetilde{\mathcal N}$ be the set of top-$K$ workers in the ordered $\mathcal N$; \label{ln:off-explore-topk}\\
      Let $\tilde p_i = \min\left\{ {u_i}/{\rho_{K+1}}, b_{max} \right\}$ for $\forall i\in \widetilde{\mathcal N}$; \label{ln:off-explore-comppay}\\
      \While{$B^{[t]} \geq \sum_{i \in \widetilde{\mathcal N}} \tilde p_i$}{ \label{ln:off-exploit-select-and-pay}
        Select workers $\widetilde{\mathcal N}^{[t]} = \widetilde{\mathcal N}$ to conduct the sensing task such that $x^{[t]}_i = 1$ for $i \in \widetilde{\mathcal N}^{[t]}$ and $x^{[t]}_i = 0$ for $\forall i \in \mathcal{N}/\widetilde{\mathcal N}^{[t]}$; \label{ln:off-exploit-select}\\
        Pay each worker $i \in \widetilde{\mathcal N}^{[t]}$ such that $p^{[t]}_i = \tilde{p}_i$ for $i \in \widetilde{\mathcal N}^{[t]}$ and $p^{[t]}_i = 0$ for $\forall i \in \mathcal{N} / \widetilde{\mathcal N}^{[t]}$; \label{ln:off-exploit-pay}\\
        $B^{[t+1]} = B^{[t]} - \sum_{i \in \widetilde{\mathcal N}^{[t]}} p^{[t]}_i$; \label{ln:off-exploit-upbudget}\\
        $t \leftarrow t+1$; \label{ln:off-exploit-upslot}
      } \label{ln:off-exploit-end}
    \caption{Our Off-line CACI mechanism $\mathsf{F_{off}}$.} 
    \label{alg:offline-caci}
    \end{algorithm}

    The pseudo-code of our mechanism is given in \textbf{Algorithm}~\ref{alg:offline-caci}. Let $\widetilde{\mathcal N}^{[t]}$ denote the set of $K$ selected workers in time slot $t$ and recall that $B^{[t]}$ is the residual budget we can use to recruit workers in time slot $t$. Let $\lambda^{[t]}(Q)$ and $\bar{r}^{[t]}(Q)$ denote the total number of times $Q$ is selected and the average cumulative reward obtained by selecting $Q$ till time slot $t$, respectively. Specifically, for $\forall Q \in \mathcal{Q}$ in time slot $t$, we have 
    \begin{equation}  \label{eq:offcaci-uplambda}
      \lambda^{[t]}(Q) = \lambda^{[t-1]}(Q) + \sum_{i \in \widetilde{\mathcal N}^{[t]}} \mathbb{I} \left( s_i \in Q \right)
    \end{equation}
    and
    \begin{equation}  \label{eq:offcaci-upavgr}
      \bar{r}^{[t]}(Q) = \frac{\bar{r}^{[t-1]}(Q) \cdot \lambda^{[t-1]}(Q) + \sum_{i \in \widetilde{\mathcal N}^{[t]}} \mathbb{I} \left( s_i \in Q \right) \cdot r^{[t]}_i}{\lambda^{[t]}(Q)}
    \end{equation}
    where $\mathbb{I}: \{ \mathrm{True}, \mathrm{False} \} \rightarrow \{1, 0\}$ represents an indicator function. As shown in Lines \ref{ln:off-init-slot}$\sim$\ref{ln:off-init-rlambda}, $B^{[t]}$ is initialized by $B^{[1]}=B$, and $\lambda^{[t]}(Q)$ and $\bar{r}^{[t]}(Q)$ are initialized by $\lambda^{[0]}(Q)=0$ and $\bar{r}^{[0]}(Q)=0$, respectively.

    The algorithm is divided into two phases: \textit{exploration} phase (see Lines~\ref{ln:off-explore-budget}$\sim$\ref{ln:off-explore-end}) and \textit{exploitation} phase (see Lines~\ref{ln:off-exploit-start}$\sim$\ref{ln:off-exploit-end}). In the first exploration phase, we uniformly select among the hypercubes (and thus the unknown workers) at random and pay the selected workers with budget $B^\#$, mainly for the purpose of estimating the ``sensing abilities'' of the hypercubes (and thus the ones of the workers). In the following exploitation phase, we select among the workers according to the sensing ability estimates learnt in the above exploration phase and pay the selected workers with the remaining budget. Specifically, as shown in Line~\ref{ln:off-explore-budget}, the budget in the exploration phase is 
    \begin{align} \label{eq:off-explorebudget}
      %B^\# \hspace{-1ex}= \left( \frac{b_{max}}{\mu^2_{max}} \right)^\frac{1}{3} \hspace{-1ex} d^\frac{M}{3} B^\frac{2}{3} \ln^\frac{1}{3} B  =  \left( \frac{2^M b_{max}}{4 \mu^2_{max}} \right)^{\frac{1}{3}} \hspace{-1ex} B^{\frac{2\alpha+M}{3\alpha+M}} \ln^{\frac{1}{3}} B
      %
      B^\# \hspace{-1ex}= \left( \frac{b_{max}}{\mu^2_{max}} \right)^\frac{1}{3} \hspace{-1ex} d^\frac{M}{3} B^\frac{2}{3} \ln^\frac{1}{3} B 
    \end{align}
    Using the budget, the exploration phase consists of $\left\lfloor \frac{B^\#}{K b_{max}} \right\rfloor$ time slots (see Line~\ref{ln:off-explore-start}). In each time slot of the exploration phase $t$, we select $K$ hypercubes in a round robin fashion and then one random worker in each of the $K$ hypercubes (see Lines~\ref{ln:off-explore-init}$\sim$\ref{ln:off-explore-comp-n-2}). The selected $K$ workers, i.e., $\widetilde{\mathcal{N}}^{[t]}$, are then recruited and each of them are paid $b_{max}$. After observing the reward feedback $r^{[t]}_i$ of any $i \in \widetilde{\mathcal{N}}^{[t]}$, we update $\lambda^{[t]}(Q)$ and $\bar{r}^{[t]}(Q)$ for $\forall Q \in \mathcal{Q}$ according to (\ref{eq:offcaci-uplambda}) and (\ref{eq:offcaci-upavgr}), respectively (see Lines~\ref{ln:off-explore-reward}$\sim$\ref{ln:off-explore-update-lambda-r}). In the end, we update the residual budget by $B^{[t+1]} = B^{[t]} - K b_{max}$, as shown in Line~\ref{ln:off-explore-upbudget}).

    In the exploitation phase, we first estimate the sensing abilities of the workers based on the observations obtained in the above exploration phase. As demonstrated in Line~ \ref{ln:off-exploit-ucb}, for each worker $i$, we calculate an \textit{Upper Confidence Bound} (UCB) on its sensing ability as
    \begin{equation} \label{eq:ucb}
      u_i = \bar r^{[t-1]}(Q(i)) + \sqrt{\frac{\ln B}{\lambda^{[t-1]}(Q(i))}}
    \end{equation}
    We then define a weight for each worker $i$ as $\rho_i = {u_i}/{b_i}$ and sort the workers in decreasing order according to $\{\rho_i\}_{i\in \mathcal N}$ (see Line~\ref{ln:sort}). In fact, $\rho_i$ is an estimate on $\rho^*_i$. Let $\widetilde{\mathcal N} \subseteq \mathcal N$ be the set of top-$K$ workers in the ordered $\mathcal N$ (see Line~\ref{ln:off-explore-topk}). As shown in Line~\ref{ln:off-explore-comppay}, the payment to any worker $i\in \widetilde{\mathcal N}$ can be calculated by
    \begin{equation} \label{eq:estpayment}
      \tilde p_i = \min \left\{ \frac{u_i}{\rho_{K+1}}, b_{max} \right\}
    \end{equation}
    As shown in Lines~\ref{ln:off-exploit-select-and-pay}$\sim$\ref{ln:off-exploit-end}, in each time slot of the exploitation phase, we select the workers $\widetilde{\mathcal N}$ to conduct the sensing task and pay each of them $\tilde p_i$, till the budget is exhausted.

  \subsection{Regret Analysis} \label{ssec:offregret}
    Considering the baseline mechanism has a constant approximation ratio with respect to the optimal one (see Sec.~\ref{ssec:offmechdesign}), we hereby evaluate our mechanism by showing how the expected cumulative reward of our off-line CACI mechanism approaches the one of the baseline mechanism. Therefore, we define the regret of our mechanism by
    \begin{eqnarray} \label{eq:regret}
      && \mathsf{Regret} \left( \mathsf{F}^*_{\mathsf{off}}(\mathbf{m}, \mathbf{b}, B), \mathsf{F_{off}}(\mathbf{s}, \mathbf{b}, B) \right)  \nonumber\\
      &=& R(\mathsf{F}^*_{\mathsf{off}}(\mathbf{m}, \mathbf{b}, B)) - R(\mathsf{F_{off}}(\mathbf{s}, \mathbf{b}, B))
    \end{eqnarray}
    where 
    \begin{equation} \label{eq:bl-creward}
      R(\mathsf{F}^*_{\mathsf{off}}(\mathbf{m}, \mathbf{b}, B)) = \left\lfloor \frac{B}{\sum_{i \in \widetilde{\mathcal N}^*} \tilde{p}^*_i} \right\rfloor \sum_{i \in \widetilde{\mathcal N}^*} \mu_i  \leq \frac{B\sum_{i \in \widetilde{\mathcal N}^*} \mu_i}{\sum_{i \in \widetilde{\mathcal N}^*} \tilde{p}^*_i}
    \vspace{-3ex}
    \end{equation}
    and
    \begin{align} \label{eq:creward}
      R(\mathsf{F}_{\mathsf{off}}(\mathbf{s}, \mathbf{b}, B)) \geq& \mathbb E \left[ \left\lfloor \frac{B-B^\#}{\sum_{i \in \widetilde{\mathcal N}} \tilde{p}_i} \right\rfloor \sum_{i \in \widetilde{\mathcal N}} \mu_i \right]  \nonumber\\
      %
      %\geq& \mathbb E \left[  \left( \frac{B-B'}{\sum_{i \in \widetilde{\mathcal N}^*} \tilde{p}_i} - 1 \right) \sum_{i \in \widetilde{\mathcal N}} \mu_i \right]  \nonumber\\
      %
      \geq& \mathbb E \left[ \frac{B-B^\#}{\sum_{i \in \widetilde{\mathcal N}} \tilde{p}_i}  \sum_{i \in \widetilde{\mathcal N}} \mu_i \right] - K \mu_{max}
    \end{align}
    denote the expected cumulative rewards yielded by the off-line baseline mechanism $\mathsf{F}^*_{\mathsf{off}}(\mathbf{m}, \mathbf{b}, B)$ and our off-line CACI mechanism $\mathsf{F}_{\mathsf{off}}(\mathbf{s}, \mathbf{b}, B)$, respectively. The regret function actually measures the price we have to pay for learning the sensing abilities of the workers.

    As discussed in Section~\ref{ssec:offmodel}, our mechanism is based on the fact that workers with similar context have comparable sensing abilities. This is formulated in \textbf{Assumption}~\ref{as:holder}.
    \begin{assumption}[H$\ddot{\mathrm{o}}$lder Condition]  \label{as:holder}
      If there exist $L>0$ and $\alpha>0$ such that for any contexts $s, s' \in \mathcal S$, it holds that
      \begin{equation} \label{eq:holder}
        \left\vert \mathbb{E}[r(s)] - \mathbb{E}[r(s')] \right\vert \leq L \|s - s'\|^\alpha
      \end{equation}
    \end{assumption}
    The H$\ddot{\mathrm{o}}$lder condition on the context space actually is the common basis of the studies on Lipschitz bandit~\cite{KleinbergSU-STOC08} and continuum bandit~\cite{Kleinberg-NIPS04}. It also should be noted that our mechanism still works even if the assumption does not hold. Nevertheless, the regret might not be bounded if the assumption was violated. 
    \begin{lemma} \label{le:intracube}
      Given any hypercube $Q \in \mathcal Q$, we have
      \begin{equation} \label{eq:intracube}
        \left| \mathbb E[r(s)] - \mathbb E[r(s')] \right| \leq \delta
      \end{equation}
      for $\forall s, s' \in Q$, where $\delta = L\left(M^{\frac{1}{2}} d^{-1}\right)^\alpha$.
    \end{lemma}
    \begin{proof}
    Since we evenly partition the context space into a group of $d^M$ hypercubes with identical size $1/d^M$ (see Sec.~\ref{ssec:offmechdesign}), we have $\|s - s'\| \leq M^{\frac{1}{2}} d^{-1}$ for $\forall s, s' \in Q$. By considering the H$\ddot{\mathrm{o}}$lder Condition (see \textbf{Assumption}~\ref{as:holder}), we then have $\vert \mathbb E[r(s)] - \mathbb E[r(s')]\vert \leq L \|s - s'\|^\alpha =  L\left(M^{\frac{1}{2}} d^{-1}\right)^\alpha$.
    \end{proof}
      The above lemma implies that, for any workers $i$ and $i'$ such that $Q(i) = Q(i')$, we have $|\mu_i - \mu_{i'}| \leq \delta$. Furthermore, for each hypercube $Q \in \mathcal Q$, we denote by $\mu_Q$ the expected reward yielded by any worker $i$ with $s_i \in Q$. It then follows $|\mu_i - \mu_{Q(i)}| \leq \delta$ for $\forall i \in \mathcal N$. Therefore, $\mu_{Q(i)}$ can be used as an accurate estimate on $\mu_i$, inspired by which, we leverage the CMAB framework to learn the  ``sensing abilities'' of a finite number of hypercubes rather than the ones of the huge number of individual workers, as shown in \textbf{Algorithm}~\ref{alg:offline-caci}.

      According to \textbf{Algorithm}~\ref{alg:baseline} and \textbf{Algorithm}~\ref{alg:offline-caci}, the reasons for the regret are two-fold: one one hand, we utilize the sensing abilities of the hypercubes $\overline{\mathbf{m}} = \{ \mu_{Q} \}_{Q \in \mathcal Q}$ to estimate the ones of the workers $\mathbf{m} = \{\mu_i\}_{i \in \mathcal N}$ and such an estimation may result in loss of reward even $\overline{\mathbf{m}}$ is known; on the other hand, since it is difficult to pre-profile the huge number of workers, $\overline{\mathbf{m}}$ actually is unknown such that we have to utilize a CMAB framework to learn it, while the learning process could lead to the loss of reward, especially considering the limited budget may considerably restrain the learning accuracy. Therefore, we decompose the regret function (\ref{eq:regret}) as follows
      \begin{eqnarray} \label{eq:decompose}
        &\mathsf{Regret} \left( \mathsf{F}^*_{\mathsf{off}}(\mathbf{m}, \mathbf{b}, B), \mathsf{F}_{\mathsf{off}}(\mathbf{s}, \mathbf{b}, B) \right)  \nonumber\\
        =& R(\mathsf{F}^*_{\mathsf{off}}(\mathbf{m}, \mathbf{b}, B)) - R(\mathsf{F}^*_{\mathsf{off}}(\overline{\mathbf{m}}, \mathbf{b}, B)) \nonumber\\
        &+ R(\mathsf{F}^*_{\mathsf{off}}(\overline{\mathbf{m}}, \mathbf{b}, B))  - R(\mathsf{F}_{\mathsf{off}}(\mathbf{m}, \mathbf{b}, B))
      \end{eqnarray}
      where 
      \begin{equation}
        R(\mathsf{F}^*_{\mathsf{off}}(\overline{\mathbf{m}}, \mathbf{b}, B)) = \left\lfloor \frac{B}{\sum_{i \in \widetilde{\mathcal N}^\dagger} \tilde{p}^\dagger_i} \right\rfloor \sum_{i \in \widetilde{\mathcal N}^\dagger} \mu_i
      \end{equation}
      %
      %$R(\mathsf{F}^*_{\mathsf{off}}(\overline{\mathbf{m}}, \mathbf{b}, B)) = \left\lfloor {B}/{\sum_{i \in \widetilde{\mathcal N}^\dagger} \tilde{p}^\dagger_i} \right\rfloor \sum_{i \in \widetilde{\mathcal N}^\dagger} \mu_i$
      %
      %
      denotes the expected cumulative reward obtained by applying the baseline mechanism $\mathsf{F}^*_{\mathsf{off}}$ to select and pay the workers with $\overline{\mathbf{m}}$ known. Therein, the workers $\mathcal N$ are sorted in decreasing order according to $\overline{\mathbf{m}}$ such that $\rho_{1^\dagger} \geq \rho_{2^\dagger} \geq \cdots \geq \rho_{N^\dagger}$ where $\rho_{i^\dagger} = {\mu_{Q(i^\dagger)}}/{b_{i^\dagger}}$ is an approximation QBR for the $i$-th worker in the ordered $\mathcal N$. Let $\widetilde{\mathcal N}^\dagger \subseteq \mathcal N$ denote the subset of top-$K$ workers. The payment to each worker $i \in \widetilde{\mathcal N}^\dagger$ is defined by 
      \begin{equation} \label{eq:paymenthybpercube}
        \tilde{p}^\dagger_i = \min \left\{ \frac{\mu_{Q(i)}}{\rho_{(K+1)^\dagger}}, b_{max} \right\}
      \end{equation}
      where $(K+1)^\dagger$ is the $(K+1)$-th worker in the ordered $\mathcal N$. We then have 
      \begin{equation} \label{eq:boundsubbaseline}
        \frac{B \sum_{i \in \widetilde{\mathcal N}^\dagger} \mu_i}{\sum_{i \in \widetilde{\mathcal N}^\dagger} \tilde{p}^\dagger_i}   - K\mu_{max}  \leq R(\mathsf{F}^*_{\mathsf{off}}(\overline{\mathbf{m}}, \mathbf{b}, B)) \leq   \frac{B\sum_{i \in \widetilde{\mathcal N}^\dagger} \mu_i}{\sum_{i \in \widetilde{\mathcal N}^\dagger} \tilde{p}^\dagger_i}  
      \end{equation}

      By combining (\ref{eq:bl-creward})$\sim$(\ref{eq:boundsubbaseline}), the regret of our off-line CACI mechanism can be decomposed into the following two sub-regret functions
      \begin{align} \label{eq:regret1-1}
        &\mathsf{Regret}(\mathsf{F}^*_{\mathsf{off}}(\mathbf{m}, \mathbf{b}, B), \mathsf{F}^*_{\mathsf{off}}(\overline{\mathbf{m}}, \mathbf{b}, B))   \nonumber\\
        =& R(\mathsf{F}^*_{\mathsf{off}}(\mathbf{m}, \mathbf{b}, B)) -R(\mathsf{F}^*_{\mathsf{off}}(\overline{\mathbf{m}}, \mathbf{b}, B))   \nonumber\\
        \leq& B \left( \frac{\sum_{i \in \widetilde{\mathcal N}^*} \mu_i}{\sum_{i \in \widetilde{\mathcal N}^*} \tilde{p}^*_i} - \frac{\sum_{i \in \widetilde{\mathcal N}^\dagger} \mu_i}{\sum_{i \in \widetilde{\mathcal N}^\dagger} \tilde{p}^\dagger_i} \right)  + K\mu_{max}
      \end{align}
      and
      \begin{align}  \label{eq:regret2-1}
        &\mathsf{Regret} \left( \mathsf{F}^*_{\mathsf{off}}(\overline{\mathbf{m}}, \mathbf{b}, B), \mathsf{F}_{\mathsf{off}}(\mathbf{s}, \mathbf{b}, B) \right)   \nonumber\\
        =& R(\mathsf{F}^*_{\mathsf{off}}(\overline{\mathbf{m}}, \mathbf{b}, B)) -  R(\mathsf{F}_{\mathsf{off}}(\mathbf{s}, \mathbf{b}, B))  \nonumber\\
        \leq& \frac{B\sum_{i \in \widetilde{\mathcal N}^\dagger} \mu_i}{\sum_{i \in \widetilde{\mathcal N}^\dagger} \tilde{p}^\dagger_i}  - \mathbb E \left[ \frac{B-B^\#}{\sum_{i \in \widetilde{\mathcal N}} \tilde{p}_i}  \sum_{i \in \widetilde{\mathcal N}} \mu_i \right] + K \mu_{max}  \nonumber\\
        \leq &  \frac{B^\# \mu_{Q((K+1)^\dagger)}}{b_{(K+1)^\dagger}} + \frac{B\delta}{b_{min}} +  K \mu_{max}  \nonumber\\
        & + (B-B^\#) \mathbb E \left[ \frac{\mu_{Q((K+1)^\dagger)}}{b_{(K+1)^\dagger}} - \frac{\sum_{i\in \widetilde{\mathcal N}} \mu_i}{\sum_{i\in \widetilde{\mathcal N}} \tilde{p}_i}  \right]  \nonumber\\
        \leq&  \frac{B^\# \mu_{max}}{b_{min}} + \frac{B\delta}{b_{min}}  +  K \mu_{max}  \nonumber\\
        & + (B-B^\#) \mathbb E \left[ \frac{\mu_{Q((K+1)^\dagger)}}{b_{(K+1)^\dagger}} - \frac{\sum_{i\in \widetilde{\mathcal N}} \mu_i}{\sum_{i\in \widetilde{\mathcal N}} \tilde{p}_i}  \right] 
      \end{align}

      In the following, we first show in \textbf{Theorem}~\ref{thm:regret} that we fine tune the granularity of the context space partitioning to fully utilizing the budget for the exploration-exploitation trade-off, such that the regret is properly bounded. We then give the corresponding proof by refining the bounds for the above two sub-regret functions $\mathsf{Regret}(\mathsf{F}^*_{\mathsf{off}}(\mathbf{m}, \mathbf{b}, B), \mathsf{F}^*_{\mathsf{off}}(\overline{\mathbf{m}}, \mathbf{b}, B))$ and $\mathsf{Regret} \left( \mathsf{F}^*_{\mathsf{off}}(\overline{\mathbf{m}}, \mathbf{b}, B), \mathsf{F}_{\mathsf{off}}(\mathbf{s}, \mathbf{b}, B) \right)$

      \begin{theorem} \label{thm:regret}
        Assume the $M$-dimensional context space is partitioned by a granularity $d = \left\lceil B^{\frac{1}{3\alpha+M}} \right\rceil$ and the budget for exploration is $B^\# = b^{\frac{1}{3}}_{max} \mu^{-\frac{2}{3}}_{max} d^{\frac{M}{3}} B^{\frac{2}{3}} \ln^{\frac{1}{3}}B$, the regret of our off-line CACI mechanism (\ref{eq:regret}) is bounded by
        \begin{equation} \label{eq:offregret}
          \frac{3 ( 2^M b_{max} \mu_{max} )^{\frac{1}{3}}}{b_{min}} B^{\frac{2\alpha+M}{3\alpha+M}} \ln^{\frac{1}{3}}B + \frac{4L M^{\frac{\alpha}{2}}}{b_{min}} B^{\frac{2\alpha+M}{3\alpha+M}} + 2K\mu_{max} 
        \end{equation}
        with probability at least $1 - \frac{2d^M}{B^2}$. It is implied that, the regret of our mechanism can be bounded by $\mathcal O \left( B^{\frac{2\alpha+M}{3\alpha+M}} \ln^{\frac{1}{3}}B \right)$
      \end{theorem}
      \begin{proof}
        We first refine the upper bounds (\ref{eq:regret1-1}) and (\ref{eq:regret2-1}) in \textbf{Proposition}~\ref{prop:subregret1} and \textbf{Proposition}~\ref{prop:subregret2}, respectively. Due to the space limit, the detailed proofs of the above two propositions are given in the supplementary material.
        \begin{proposition} \label{prop:subregret1}
          Recalling $\mathsf{F}^*_{\mathsf{off}}(\mathbf{m}, \mathbf{b}, B)$ and $\mathsf{F}^*_{\mathsf{off}}(\overline{\mathbf{m}}, \mathbf{b}, B)$ denote the solutions where the workers are selected and paid according to $\mathbf{m}$ and $\overline{\mathbf{m}}$, respectively, we have
          \begin{equation}
            \mathsf{Regret}(\mathsf{F}^*_{\mathsf{off}}(\mathbf{m}, \mathbf{b}, B), \mathsf{F}^*_{\mathsf{off}}(\overline{\mathbf{m}}, \mathbf{b}, B)) \leq \frac{2B\delta}{b_{min}} + K\mu_{max}
          \end{equation}
        \end{proposition}
        \begin{proposition} \label{prop:subregret2}
          For $\mathsf{F}^*_{\mathsf{off}}(\overline{\mathbf{m}}, \mathbf{b}, B)$ and $\mathsf{F}_{\mathsf{off}}(\mathbf{s}, \mathbf{b}, B)$, we have
          \begin{align*} 
            &\mathsf{Regret} \left( \mathsf{F}^*_{\mathsf{off}}(\overline{\mathbf{m}}, \mathbf{b}, B), \mathsf{F}_{\mathsf{off}}(\mathbf{s}, \mathbf{b}, B) \right) \nonumber\\
            \leq& \frac{B^\# \mu_{max} + 2B\delta}{b_{min}} + \frac{2B}{b_{min}}\sqrt{\frac{d^M b_{max} \ln B}{B^\#}} + K \mu_{max} 
          \end{align*}
          hold with probability at least $1 - \frac{2d^M}{B^2}$.
        \end{proposition}

        We combine \textbf{Proposition}~\ref{prop:subregret1} and \textbf{Proposition}~\ref{prop:subregret2} and thus have
        \begin{align}
          &\mathsf{Regret}(\mathsf{F}^*_{\mathsf{off}}(\mathbf{m}, \mathbf{b}, B), \mathsf{F}_{\mathsf{off}}(\mathbf{s}, \mathbf{b}, B))   \nonumber\\
          %
          %\leq& \frac{(B+1)\delta}{c_{min}} + 2\tau\mu_{max} + \frac{B' \mu_{max}}{b_{min}}  \nonumber\\
          %
          %& + (B-B') \left[ \frac{2}{b_{min}}  \sqrt{\frac{d^M b_{max} \ln B}{B'}} + \frac{\delta}{b_{min}}  \right]  \nonumber\\
          %
          %\leq& \frac{2(B-B')}{b_{min}} \sqrt{\frac{d^M b_{max} \ln B}{B'}} + \frac{\mu_{max}}{b_{min}}B' + \frac{4B\delta}{b_{min}} + 2\tau \mu_{max} \nonumber\\
          %
          \leq& \frac{2B}{b_{min}} \sqrt{\frac{d^M b_{max} \ln B}{B^\#}} + \frac{\mu_{max}}{b_{min}}B^\# + \frac{4B\delta}{b_{min}}   + 2 K \mu_{max}
        \end{align}
        By minimizing the rightmost side of the above inequality, we find that, when $B^\# = \left(\frac{b_{max}}{\mu^2_{max}}\right)^{\frac{1}{3}} d^{\frac{M}{3}} B^{\frac{2}{3}} \ln^{\frac{1}{3}}B$, we can tighten the upper bound of the regret as follows
        \begin{align*}
          &\mathsf{Regret}(\mathsf{F}^*_{\mathsf{off}}(\mathbf{m}, \mathbf{b}, B), \mathsf{F}_{\mathsf{off}}(\mathbf{s}, \mathbf{b}, B)) \nonumber\\
          \leq & \frac{3 (b_{max} \mu_{max})^{\frac{1}{3}}}{b_{min}} d^{\frac{M}{3}} B^{\frac{2}{3}} \ln^{\frac{1}{3}}B + \frac{4L M^{\frac{\alpha}{2}}}{b_{min}} d^{-\alpha} B + 2K\mu_{max} \nonumber\\
          \leq& \left( \frac{3 ( 2^M b_{max} \mu_{max} )^{\frac{1}{3}}}{b_{min}}  \ln^{\frac{1}{3}}B  + \frac{4L M^{\frac{\alpha}{2}}}{b_{min}} \right) B^{\frac{2\alpha+M}{3\alpha+M}} + 2K\mu_{max} 
        \end{align*}
        where we have the second line by letting $d = \left\lceil B^{\frac{1}{3\alpha+M}} \right\rceil$ such that $B^{\frac{1}{3\alpha+M}} \leq d \leq 2B^{\frac{1}{3\alpha+M}}$.
      \end{proof}

      As demonstrated earlier, the upper bound on the regret for our offline CACI mechanism can be written as $\mathcal{O} \left( d^{\frac{M}{3}} B^{\frac{2}{3}} \ln^{\frac{1}{3}}B \right)$. This can be further represented as $\mathcal O \left( B^{\frac{2\alpha+M}{3\alpha+M}} \ln^{\frac{1}{3}}B \right)$ by substituting $d = \left\lceil B^{\frac{1}{3\alpha+M}} \right\rceil$. It is evident that $d$ is a crucial factor influencing the upper bound. Numerous existing studies have explored incentive mechanisms based on the conventional CMAB framework, with an upper bound of $\mathcal O \left( N^\frac{1}{3} B^\frac{2}{3} \ln^\frac{1}{3}B \right)$ on their regrets \cite{BiswasJMN-AAMAS15,XiaoWZG-ICDCS20}. In our CACI mechanism, we can adjust the granularity of partitioning, i.e., $d$, such that $d^\frac{M}{3} \ll N^\frac{1}{3}$. This results in a much tighter bound, especially when $B$ is significantly limited and $N$ is considerably large.

  \subsection{Truthfulness and Individual Rationality}  \label{ssec:offtruth}
    Given the good regret performance of our off-line CACI mechanism, we continue to show its truthfulness and individual rationality in \textbf{Theorem}~\ref{thm:offcaci-truth}. Detailed proof can be found in the supplementary material.
    \begin{theorem}  \label{thm:offcaci-truth}
      Our off-line CACI mechanism is truthful and individually rational.
    \end{theorem}

\section{Our On-Line CACI Mechanism} \label{sec:online-caci}
  In the above off-line CACI mechanism, we spend $B^\#$ on exploring a given (fixed) set of unknown workers and utilize the rest of the budget to exploit the workers according to the experience obtained in the exploration stage. Nevertheless, when the available workers are time-varying, we cannot properly leverage the exploration-exploitation trade-off by dividing the selection-and-payment process into two ``disjoint'' phases. In this section, we reveal how to design an efficient mechanism to incentivize on-line workers. We first present our mechanism design in Sec.~\ref{ssec:online-mech}. We then analyze the regret of the proposed mechanism in Sec.~\ref{ssec:onregret}, and finally discuss its truthfulness and individual rationality in Sec.~\ref{ssec:ontruth}.

  \vspace{-2ex}
  \subsection{Design of On-Line CACI Mechanism}  \label{ssec:online-mech}
    As mentioned in Sec.~\ref{ssec:onlinemodel}, the workers available in each time slot may be different, such that learning the sensing abilities of the individual workers is not feasible. Fortunately, since the context space of the volatile workers is fixed, we can instead learn the mapping between the context and the sensing abilities and thus select and pay the elite workers.

    The pseudo-code of our mechanism is given in \textbf{Algorithm}~\ref{alg:online-caci}. In the initialization phase (see Lines~\ref{ln:on-ini}$\sim$\ref{ln:on-init-update-B}), we uniformly select one worker, e.g., worker $i$, in each hypercube $Q$ such that $s_i \in Q$, to conduct the sensing task (see Line~\ref{ln:on-init-select}) and pay the selected worker $p^{[t]}_i = b_{max}$ (see Line~\ref{ln:on-init-pay}). We then observe the reward feedback $r^{[t]}_i$, according to which, we update $\bar{r}^{[t]}(Q)$ and  $\lambda^{[t]}(Q)$, as shown in Lines~(\ref{ln:on-init-reward})$\sim$(\ref{ln:on-init-update-lambda}). We also deduct the payment to the $d^M$ selected workers from the total budget (see Line~\ref{ln:on-init-update-B}). In a nutshell, we select $d^M$ workers (i.e., one worker in each hypercube) in the initialization phase to initialize $\bar{r}^{[t]}(Q)$ and  $\lambda^{[t]}(Q)$ for $\forall Q \in \mathcal{Q}$. 
    In each iteration $t$ of the next exploration-and-exploitation phase, when the residual budget is sufficient (i.e., $B^{[t]} \geq K b_{max}$), we first calculate an UCB index (\ref{eq:onucb}) for each worker $i \in \widetilde{\mathcal N}^{[t]}$ as shown in Line~\ref{ln:on-calucb}
    \begin{equation} \label{eq:onucb}
      u^{[t]}_i = \bar{r}^{[t-1]}(Q(i)) + \sqrt{\frac{(K+1)\ln t}{\lambda^{[t-1]}(Q(i))}}
    \end{equation}
    We then sort the workers $\widetilde{\mathcal N}^{[t]}$ according to $\rho_i = u^{[t]}_i / b_i$ such that $\rho_1 \geq \rho_2 \geq \cdots \geq \rho_{N^{[t]}}$ (see Line~\ref{ln:on-sort}). The top-$K$ workers $\widetilde{\mathcal N}^{[t]} \subseteq \mathcal{N}^{[t]}$ are selected to conduct the sensing task (see Line~\ref{ln:on-select}), and each of the selected workers is paid by $p^{[t]}_i = \min\{ u^{[t]}_i/\rho_{K+1}, b_{max}\}$ (see Line~\ref{ln:on-pay2}). After observing the corresponding reward feedbacks of the selected $K$ workers, we update $\lambda^{[t]}(Q)$ and $\bar{r}^{[t]}(Q)$ for $\forall Q \in \mathcal{Q}$ according to (\ref{eq:offcaci-uplambda}) and (\ref{eq:offcaci-upavgr}), respectively, as demonstrated in Lines \ref{ln:on-reward}$\sim$\ref{ln:on-update-lambda-r}. We finally update the residual budget $B^{[t+1]} = B^{[t]} - \sum_{i \in \widetilde{\mathcal N}^{[t]}} \tilde{p}_i$, as shown in Line~\ref{ln:on-update-budget}.
    \begin{algorithm}[htb!]
      \KwIn{On-line unknown workers $\mathcal N^{[t]}$ and their context vector $\mathbf{s}^{[t]}$ and bid vector $\mathbf{b}^{[t]}$ with $t=1,2,\cdots$, and total budget $B$.}
      \KwOut{Selection policy $\mathsf{X}^{[t]}_{\mathsf{on}}$ and payment policy $\mathsf{P}^{[t]}_{\mathsf{on}}$ where $t=1,2,\cdots$}
      $\rhd$ \textit{Initialization phase}: \label{ln:on-init-start}\\
      $t=1$;  $B^{[t]} = B$; \label{ln:on-ini}\\
      \For{ $\mathbf{each}~Q \in \mathcal Q$}{
        Select worker $i \in \mathcal{N}^{[t]}$ with $s_i \in \mathcal Q$ at random such that $x^{[t]}_i =1$; \label{ln:on-init-select}\\
        Pay worker $i$ such that $p^{[t]}_i = b_{max}$;  \label{ln:on-init-pay}\\
        Observe $r^{[t]}_i$, i.e., worker $i$'s reward feedback $r^{[t]}_i$;  \label{ln:on-init-reward}\\
        $\bar{r}^{[t]}(Q)=r^{[t]}_i$; \label{ln:on-init-update-r}\\
        $\lambda^{[t]}(Q) = 1$; \label{ln:on-init-update-lambda}\\
      }  \label{ln:on-init-end}
      $B^{[t+1]} = B^{[t]}-d^M b_{max}$;  \label{ln:on-init-update-B}\\
      $\rhd$ \textit{Exploration-and-exploitation phase}: \\
      $t \leftarrow t+1$; \label{ln:on-update-t}\\
      \While{$B^{[t]} \geq K b_{max}$}{
        $u^{[t]}_i = \bar{r}^{[t-1]}(Q(i)) + \sqrt{\frac{(K+1)\ln t}{\lambda^{[t-1]}(Q(i))}}$ for $\forall i \in \mathcal{N}^{[t]}$;  \label{ln:on-calucb}\\
        Sort workers $\mathcal N^{[t]}$ in decreasing order according to $\rho_i = {u^{[t]}_i}/{b_i}$ such that $\rho_1 \geq \rho_2 \geq \cdots, \geq \rho_{N^{[t]}}$; \label{ln:on-sort}\\
        Select top-$K$ workers $\widetilde{\mathcal{N}}^{[t]} \subseteq \mathcal{N}^{[t]}$ to conduct the sensing task such that $x_i = 1$ for $\forall i\in \widetilde{\mathcal{N}}^{[t]}$ while $x_i = 0$ for others; \label{ln:on-select}\\
        % \begin{equation*}
        %   x^{[t]}_i = \begin{cases}
        %     1, ~\forall i\in \widetilde{\mathcal{N}}^{[t]} \\
        %     0, ~\forall i\in \mathcal{N}^{[t]} / \widetilde{\mathcal{N}}^{[t]} 
        %   \end{cases}
        % \end{equation*} \label{ln:on-select}\\
        %
        %
        Pay each worker $i \in \widetilde{\mathcal N}^{[t]}$ such that $p^{[t]}_i = \min\left\{{u_i}/{\rho_{K+1}}, b_{max}\right\}$ for $\forall i \in \widetilde{\mathcal N}^{[t]}$ and $p^{[t]}_i = 0$ for $\forall i \in \mathcal{N}^{[t]} /\ \widetilde{\mathcal N}^{[t]}$; \label{ln:on-pay2}\\
        %
        % \begin{equation*}
        %   p^{[t]}_i = \begin{cases}
        %     \min \left\{ \frac{u_i}{\rho_{K+1}}, b_{max} \right\}, ~\forall i \in \widetilde{\mathcal N}^{[t]} \\
        %     %
        %     0, ~\forall i \in \widetilde{\mathcal N}^{[t]} / \widetilde{\mathcal N}^{[t]}
        %   \end{cases}
        % \end{equation*} \label{ln:on-pay}\\
        %
        %
        %
        Observe the reward feedback $r^{[t]}_i$ for $\forall i \in \widetilde{\mathcal{N}}^{[t]}$; \label{ln:on-reward}\\
        %
        %$\tilde{p}_i = \min\left\{\frac{u_i}{\rho_{K+1}}, b_{max}\right\}$ for any $i \in {\mathcal{N}}^{[t]}$; \label{ln:on-pay1}\\
        %
        %
        Update $\lambda^{[t]}(Q)$ and $\bar{r}^{[t]}(Q)$ for $\forall Q \in \mathcal{Q}$ according to (\ref{eq:offcaci-uplambda}) and (\ref{eq:offcaci-upavgr}), respectively; \label{ln:on-update-lambda-r}\\
        $B^{[t+1]} = B^{[t]} - \sum_{i \in \widetilde{\mathcal{N}}^{[t]}} p^{[t]}_i$; \label{ln:on-update-budget}\\
        $t \leftarrow t+1$; \label{ln:on-update-slot}
      }
    \caption{Our on-line CACI mechanism $\mathsf{F_{on}}$.} 
    \label{alg:online-caci}
    \end{algorithm}

  \vspace{-2ex}
  \subsection{Regret Analysis of On-Line CACI} \label{ssec:onregret}
    The regret of our on-line CACI mechanism is defined by the gap between the expected cumulative reward of our on-line mechanism $\mathsf{F_{on}} ( (\mathbf{s}^{[t]}, \mathbf{b}^{[t]} )_{t \in \mathcal T}, B )$ and the one of its baseline counterpart $\mathsf{F^*_{on}} ( (\mathbf{m}^{[t]}, \mathbf{b}^{[t]} )_{t \in \mathcal T}, B )$. Let $\widetilde{T}^*$ be the number of time steps the on-line baseline mechanism proceeds and $\widetilde{T}$ be the number of the ones our on-line CACI mechanism proceeds. The regret function is defined by
    \begin{align} \label{eq:onregret-def}
      &\mathsf{Regret} \left( \mathsf{F^*_{on}} \left( (\mathbf{m}^{[t]}, \mathbf{b}^{[t]} )_{t \in \mathcal T}, B \right), \mathsf{F}_\mathsf{on} \left( (\mathbf{s}^{[t]}, \mathbf{b}^{[t]} )_{t \in \mathcal T}, B \right) \right)  \nonumber\\
      =&  \mathbb{E} \left[ \sum^{\widetilde{T}^*}_{t=1} \sum_{i \in \widetilde{\mathcal N}^{*[t]}} \mu_i - \sum^{\widetilde{T}}_{t=1} \sum_{i \in \widetilde{\mathcal N}^{[t]}} \mu_i \right]
    \end{align}
    Before presenting the upper bound of the regret function (\ref{eq:onregret-def}) in \textbf{Theorem}~\ref{thm:onregretbd}, we first give some notions as follows
    \begin{align}
    \begin{cases}
      %\Delta_{min} = \min_t \left\{ \sum_{i \in \widetilde{\mathcal N}^{\dagger[t]}} \frac{\mu_{Q(i)}}{b_i} - \sum_{i \in \widetilde{\mathcal N}^{[t]}} \frac{\mu_{Q(i)}}{b_i}, \forall \widetilde{\mathcal N}^{[t]} \neq \widetilde{\mathcal N}^{\dagger[t]} \right\}  \vspace{1ex}\\
      %
      \Delta_{min} = \min_t \Big\{ \sum_{i \in {\mathcal N}_0} \frac{\mu_{Q(i)}}{b_i} - \sum_{i \in {\mathcal N}_1} \frac{\mu_{Q(i)}}{b_i}, \\
      \forall \mathcal{N}_0, \mathcal{N}_1 \subseteq \mathcal{N}^{[t]}~\text{such that}~|\mathcal{N}_0| = |\mathcal{N}_1| = K, \mathcal{N}_0 \neq \mathcal{N}_1 \Big\} \vspace{1ex}\\
      \nabla_{max} = \max_{Q, Q' \in \mathcal{Q}} \left| \mu_Q - \mu_{Q'} \right|
    \end{cases}
    \end{align}
    In particular, as demonstrated in \textbf{Theorem}~\ref{thm:onregretbd}, a fine tuned granularity of partitioning is necessitated to ensure that the regret function (\ref{eq:onregret-def}) can be bounded. It is worthy to note that the upper bound of the regret of our on-line CACI mechanism is $\mathcal{O}(B)$, which is slightly worse than the one of the off-line counterpart; nevertheless, this is the price we have to pay for incentivizing on-line unknown workers.
    \begin{theorem} \label{thm:onregretbd}
      Let $d = \left\lceil B^{\frac{1}{3 \alpha + M}} \right\rceil$. The upper bound of the regret of our on-line CACI mechanism (\ref{eq:onregret-def}) is
      \begin{align} \label{eq:onregretbd}
        &\varepsilon_0 B + \frac{4L M^{\frac{\alpha}{2}}}{b_{min}} B^{\frac{2 \alpha + M}{3 \alpha + M}} + \varepsilon_1 \left( \frac{K \pi^2}{3} + 2 \right) B^{\frac{M}{3\alpha+M}}   \nonumber\\
        &+ \varepsilon_2 K^2(K+1) B^{\frac{M}{3\alpha+M}} \ln\frac{B}{b_{min} K}
      \end{align} 
      where
      \begin{equation}
      \begin{cases}
        \varepsilon_0 = \frac{b_{max}\mu^2_{max}}{b^2_{min}\mu_{min}} + \frac{\mu_{max}}{b_{min}}  \\
        \varepsilon_1 = 2^M \left(  \frac{b_{max} \mu_{max}}{b_{min}} +  \nabla_{max} \right) \\
        \varepsilon_2 = \frac{8^M (b_{max} \mu_{max} + b_{min} \nabla_{max})}{b^3_{min} \Delta^2_{min}}  
      \end{cases}
      \end{equation}
    \end{theorem}
    \begin{proof}
      We borrow the basic idea of analyzing the regret of the off-line CACI mechanism, we introduce mechanism $\mathsf{F^*_{on}}((\overline{\mathbf{m}}^{[t]}, \mathbf{b}^{[t]} )_{t \in \mathcal T}, B )$ to decompose the above regret function (\ref{eq:onregret-def}), where $\overline{\mathbf{m}}^{[t]} = \{ \mu_{Q(i)} \}_{i \in \mathcal{N}^{[t]}}$. Specifically, in $\mathsf{F^*_{on}}( (\overline{\mathbf{m}}^{[t]}, \mathbf{b}^{[t]} )_{t \in \mathcal T}, B )$, we adopt $\mu_{Q(i)}$ as the estimate of the sensing ability of worker $i$, and call \textbf{Algorithm}~\ref{alg:onbaseline} to incentivize the on-line workers. Assume $\widetilde{T}^\dagger$ denotes the number of time slots $\mathsf{F^*_{on}}( (\overline{\mathbf{m}}^{[t]}, \mathbf{b}^{[t]} )_{t \in \mathcal T}, B )$ proceeds, and $\widetilde{\mathcal{N}}^{\dagger[t]} \subseteq \widetilde{\mathcal{N}}^{[t]}$ represents the set of the workers selected by $\mathsf{F^*_{on}}( (\overline{\mathbf{m}}^{[t]}, \mathbf{b}^{[t]} )_{t \in \mathcal T}, B )$ in time slot $t$. We then decompose the regret function (\ref{eq:onregret-def}) as follows
      %
      %\vspace{-2ex}
      \begin{align} \label{eq:onregret-dec}
        &\mathsf{Regret} \left( \mathsf{F^*_{on}} \left( (\mathbf{m}^{[t]}, \mathbf{b}^{[t]} )_{t \in \mathcal T}, B \right), \mathsf{F}_\mathsf{on} \left( (\mathbf{s}^{[t]}, \mathbf{b}^{[t]} )_{t \in \mathcal T}, B \right) \right)  \nonumber\\
        =& \mathbb{E} \left[ \sum^{\widetilde{T}^*}_{t=1} \sum_{i \in \widetilde{\mathcal N}^{*[t]}} \mu_i - \sum^{\widetilde{T}}_{t=1} \sum_{i \in \widetilde{\mathcal N}^{\dagger[t]}} \mu_i \right]  \nonumber\\
        & +\mathbb{E} \left[ \sum^{\widetilde{T}}_{t=1} \sum_{i \in \widetilde{\mathcal N}^{\dagger[t]}} \mu_i - \sum^{\widetilde{T}}_{t=1} \sum_{i \in \widetilde{\mathcal N}^{[t]}} \mu_i \right]
      \end{align}
      The bounds of the two items in the regret function (\ref{eq:onregret-dec}) are given in \textbf{Proposition}~\ref{prop:on-subregret1} and \textbf{Proposition}~\ref{prop:on-subregret2}, respectively. Due to the space limit, the detailed proofs of the two proposition are shown in the supplementary material.
      \begin{proposition} \label{prop:on-subregret1}
        Recall that $\widetilde{\mathcal N}^{*[t]}$ and $\widetilde{\mathcal N}^{\dagger[t]}$ denote the workers selected by $\mathsf{F^*_{on}} \left( (\mathbf{m}^{[t]}, \mathbf{b}^{[t]} )_{t \in \mathcal T}, B \right)$ and $\mathsf{F^*_{on}}((\overline{\mathbf{m}}^{[t]}, \mathbf{b}^{[t]} )_{t \in \mathcal T}, B )$ in time slot $t$, respectively, and $\widetilde{T}^*$ and $\widetilde{T}$ denote the numbers of time slots $\mathsf{F_{on}} \left( (\mathbf{m}^{[t]}, \mathbf{b}^{[t]} )_{t \in \mathcal T}, B \right)$ and $\mathsf{F_{on}}(({\mathbf{s}}^{[t]}, \mathbf{b}^{[t]} )_{t \in \mathcal T}, B )$ proceed, respectively. We have
        \begin{align} \label{eq:on-subregret1}
          & \mathbb{E} \left[ \sum^{\widetilde{T}^*}_{t=1} \sum_{i \in \widetilde{\mathcal N}^{*[t]}} \mu_i - \sum^{\widetilde{T}}_{t=1} \sum_{i \in \widetilde{\mathcal N}^{\dagger[t]}} \mu_i \right]  \nonumber\\
          \leq& \left( \frac{b_{max}\mu^2_{max}}{b^2_{min}\mu_{min}} + \frac{\mu_{max}}{b_{min}} \right) B  +  \left( \frac{b_{max}\mu_{max}}{b^2_{min} \mu_{min}} + \frac{2}{b_{min}} \right)\delta B \nonumber\\
          & + \frac{4 b_{max} \mu_{max} K^2 (K+1) d^M}{b^3_{min} \Delta^2_{min}} \ln \frac{B}{K b_{min}}     \nonumber\\
          & + \frac{b_{max} \mu_{max}}{b_{min}} \left( \frac{K \pi^2}{3} + 2 \right) d^M
        \end{align}
      \end{proposition}
      \begin{proposition} \label{prop:on-subregret2}
        Recall that $\widetilde{\mathcal N}^{\dagger[t]}$ and $\widetilde{\mathcal N}^{[t]}$ denote the workers selected by $\mathsf{F_{on}}((\overline{\mathbf{m}}^{[t]}, \mathbf{b}^{[t]} )_{t \in \mathcal T}, B )$ and $\mathsf{F_{on}}(({\mathbf{s}}^{[t]}, \mathbf{b}^{[t]} )_{t \in \mathcal T}, B )$ in time slot $t$, respectively, and $\widetilde{T}^*$ and $\widetilde{T}$ denote the numbers of time slots $\mathsf{F_{on}} \left( (\mathbf{m}^{[t]}, \mathbf{b}^{[t]} )_{t \in \mathcal T}, B \right)$ and $\mathsf{F_{on}}(({\mathbf{s}}^{[t]}, \mathbf{b}^{[t]} )_{t \in \mathcal T}, B )$ proceed, respectively. We have
        \begin{align}  \label{eq:on-subregret2}
          & \mathbb{E} \left[ \sum^{\widetilde{T}}_{t=1} \sum_{i \in \widetilde{\mathcal N}^{\dagger[t]}} \mu_i - \sum^{\widetilde{T}}_{t=1} \sum_{i \in \widetilde{\mathcal N}^{[t]}} \mu_i \right]  \nonumber\\
          \leq& \left( \frac{4K^2 (K+1)}{(b_{min} \Delta_{min})^2} \ln\frac{B}{K b_{min}} + \frac{K \pi^2}{3} + 1 \right) \nabla_{max} d^M  + \frac{2\delta B}{b_{min}}
        \end{align}
      \end{proposition}
      By letting $d = \left\lceil B^{\frac{1}{3\alpha+M}} \right\rceil$ such that $B^{\frac{1}{3\alpha+M}} \leq d \leq 2B^{\frac{1}{3\alpha+M}}$, ( \ref{eq:on-subregret1}) and (\ref{eq:on-subregret2}) can be re-written as
      \begin{align}
        & \mathbb{E} \left[ \sum^{\widetilde{T}^*}_{t=1} \sum_{i \in \widetilde{\mathcal N}^{*[t]}} \mu_i - \sum^{\widetilde{T}}_{t=1} \sum_{i \in \widetilde{\mathcal N}^{\dagger[t]}} \mu_i \right]  \nonumber\\
        \leq& \left( \frac{b_{max}\mu^2_{max}}{b^2_{min}\mu_{min}} + \frac{\mu_{max}}{b_{min}} \right) B  \nonumber\\
        & + \left( \frac{b_{max}\mu_{max}}{b^2_{min} \mu_{min}} + \frac{2}{b_{min}} \right) L M^{\frac{\alpha}{2}} B^{\frac{2\alpha+M}{3\alpha + M}}    \nonumber\\
        & + \frac{2^{2+M} b_{max} \mu_{max} K^2 (K+1)}{b^3_{min} \Delta^2_{min}} B^{\frac{M}{3\alpha+M}} \ln \frac{B}{K b_{min}}    \nonumber\\
        & + \frac{2^M b_{max} \mu_{max}}{b_{min}} \left( \frac{K \pi^2}{3} + 2 \right) B^{\frac{M}{3\alpha+M}}
      \end{align}
      and
      \begin{align}
        & \mathbb{E} \left[ \sum^{\widetilde{T}}_{t=1} \sum_{i \in \widetilde{\mathcal N}^{\dagger[t]}} \mu_i - \sum^{\widetilde{T}}_{t=1} \sum_{i \in \widetilde{\mathcal N}^{[t]}} \mu_i \right] \nonumber\\
        \leq& \frac{2^{2+M} \nabla_{max} K^2 (K+1)}{(b_{min} \Delta_{min})^2} B^{\frac{M}{3\alpha+M}} \ln\frac{B}{K b_{min}}     \nonumber\\
        & + 2^M \left( \frac{K \pi^2}{3} + 1 \right)\nabla_{max}  B^{\frac{M}{3\alpha+M}} + \frac{2L M^{\frac{\alpha}{2}}}{b_{min}} B^{\frac{2\alpha+M}{3\alpha+M}}
      \end{align}
      by combining which, we complete the proof of \textbf{Theorem}~\ref{thm:onregretbd}.
    \end{proof}

    %

  %\vspace{-6ex}
  \subsection{Truthfulness and Individual Rationality of On-Line CACI} \label{ssec:ontruth}
    We continue to show the truthfulness and individual rationality of our on-line CACI mechanism. Since the proof of the above theorem is similar to the ones of \textbf{Theorems}~\ref{thm:offcaci-truth}, we omit the detailed proof, especially considering the space limit.
    \begin{theorem} \label{thm:on-truth-ration}
      Our on-line CACI mechanism is of truthfulness and individual rationality.
    \end{theorem}

\section{Experiments} \label{sec:exp}
  We first introduce the reference algorithms in Sec.~\ref{ssec:refalgo} and then compare them with our off-line CACI mechanism on both synthetic and real datasets in Sec.~\ref{ssec:exp-offcaci}. We also extend the reference algorithms to deal with on-line workers and present the comparison results with our on-line CACI mechanism in Sec.~\ref{ssec:exp-oncaci}.

  \subsection{Reference Algorithms}  \label{ssec:refalgo}
    We mainly compare our CACI mechanism with the following state-of-the-art ones which can be adapted to address our $K$-WIN problems including the off-line version $\mathfrak{P}_{\mathsf{off}}$ (see Sec.~\ref{ssec:offprobform}) and the on-line counterpart $\mathfrak{P}_{\mathsf{on}}$ (see \ref{ssec:onlinemodel}).
    \begin{itemize}
      \item \textbf{Baseline}: As demonstrated in \textbf{Algorithm}~\ref{alg:baseline}, it is assumed in the baseline mechanism that the sensing ability of each worker is known as prior knowledge. Through the off-line baseline mechanism, we get a nearly optimal solution for $\mathfrak{P}_{\mathsf{off}}$, while the mechanism ensures the truthfulness and the individual rationality, as demonstrated in Sec.~\ref{sec:offline}, respectively. Furthermore, we can extend the off-line baseline mechanism to address the on-line version of our $K$-WIN problem, as illustrated in Sec.~\ref{ssec:extonline}. The baseline mechanisms are actually used to calculate the regrets of our CACI mechanisms and the other reference ones.
      \item \textbf{CMAB-based}: In \cite{BiswasJMN-AAMAS15}, a CMAB-based mechanism is proposed to crowdsource so-called ``time critical'' sensing tasks in face of a fixed group of workers. Although the context of the problem is different from ours, the proposed mechanism can be used to address our off-line $K$-WIN problem. It can be considered as a degeneration of our off-line CACI mechanism where the context space is sufficiently partitioned such that each hypercube contains only one worker. In another word, we leverage the CMAB method to learn the sensing abilities of the individual workers. In fact, similar idea is also adapted in \cite{XiaoWZG-ICDCS20}. Unfortunately, when the available workers in each time slot are volatile, the traditional CMAB-based mechanism is not applicable; therefore, these approaches based on standard CMAB framework cannot be adapted to address our on-line $K$-WIN problem 
      \item \textbf{$\varepsilon$-first}: We adapt the $\varepsilon$-first mechanism ~\cite{TranCMRJ-AAAI10,ThanhSRJ-AI14} to address our off-line $K$-WIN problem by allocating $\varepsilon$-fraction of the total budget (e.g., $\varepsilon=0.3, 0.5$ in our experiments) for exploration and using the residual budget for exploitation. In a nutshell, on one hand, the $\varepsilon$-first mechanism is similar to the traditional CMAB-based one in the sense that the workers are explored and exploited individually; on the other hand, instead of elaborately budgeting for the exploration-exploitation trade-off (see (\ref{eq:off-explorebudget}) in Sec.~\ref{ssec:offmechdesign}), it determines the budget for exploration in a ``reckless'' manner. In fact, the the $\varepsilon$-first mechanism can be adapted to adress the on-line version of the $K$-WIN problem, by harnessing the context information of the on-line workers. We first divide the context space into a set of disjoint hypercubes, and then utilize the principle of $\varepsilon$-first to explore and exploit the hypercubes instead of the individual volatile workers. 
    \end{itemize}

  \subsection{Experiment Results of Off-Line CACI Mechanism}  \label{ssec:exp-offcaci}
    In this section, our off-line CACI mechanism is evaluated on both the synthetic and the real datasets.
    \subsubsection{Evaluation with Synthetic Data}  \label{ssec:exp-offcaci-syn}
      We assume there are $10^5$ worker and let $K=150$. Each worker $i$ has its true cost $c_i$ uniformly distributed over $[0.2,1]$ and randomly chooses a value from $[c_i, 1]$ as its bid $b_i$. We suppose the context space $\mathcal S$ has $M = 2$ dimensions and each dimension is normalized to $[0,1]$ as mentioned in Sec.~\ref{ssec:offmodel}. The workers have their contexts uniformly distributed in the context space $\mathcal S$. We also randomly set the workers' sensing abilities such that the H$\ddot{\mathrm{o}}$lder condition holds for $\alpha=1$ for the purpose of quantitative analysis. Note that all our empirical analysis still holds when the constant parameters take another values. For each reported data sample, we repeat the experiments for ten times and take average over the results.

\begin{figure}
      \centering
        \subfigure[Cumulative reward vs. budget]{\label{fig:offsyn-reward-budget}\includegraphics[width=.49\columnwidth]{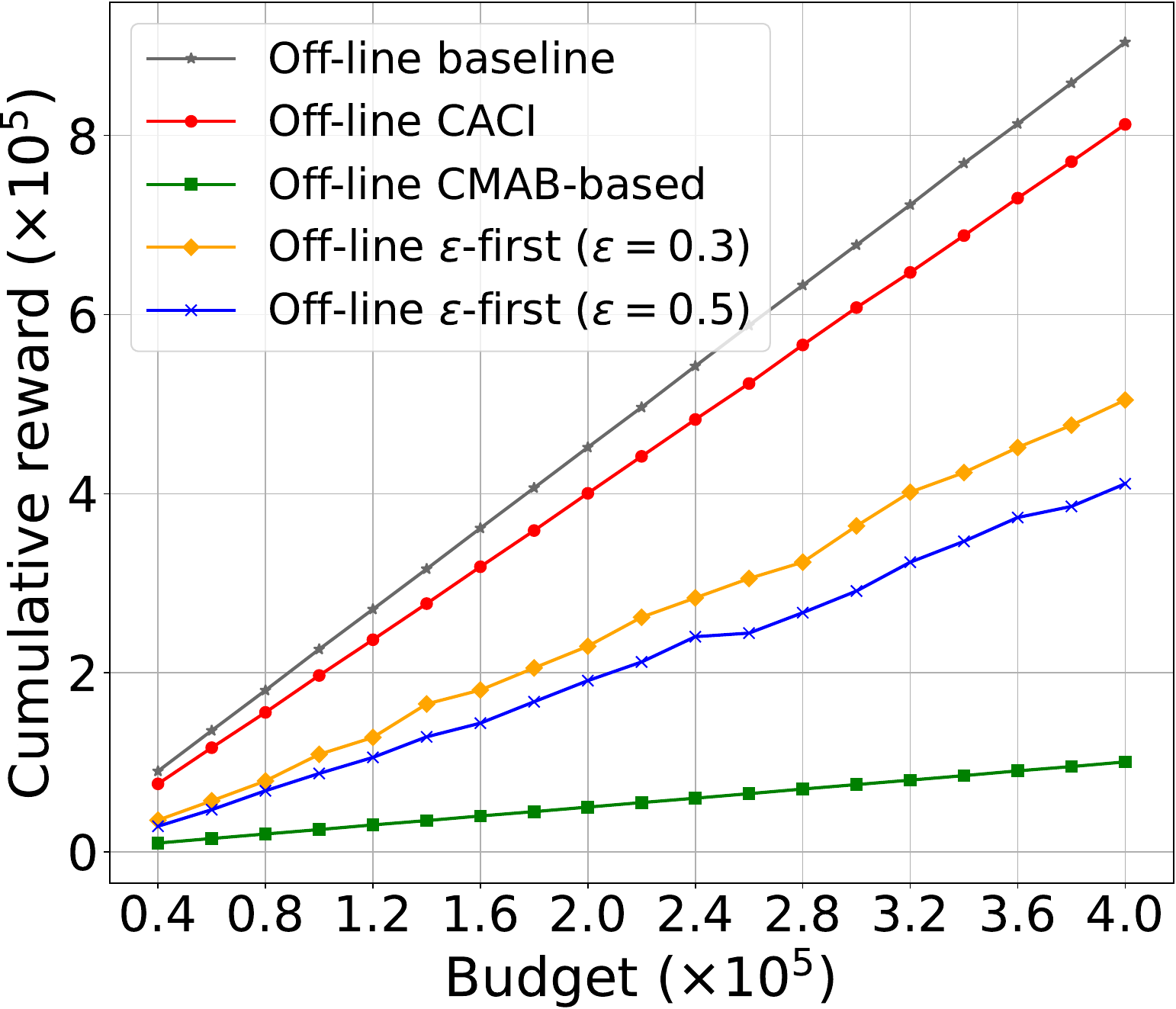}}
        \subfigure[Regret vs. budget]{\label{fig:offsyn-regret-budget}\includegraphics[width=.49\columnwidth]{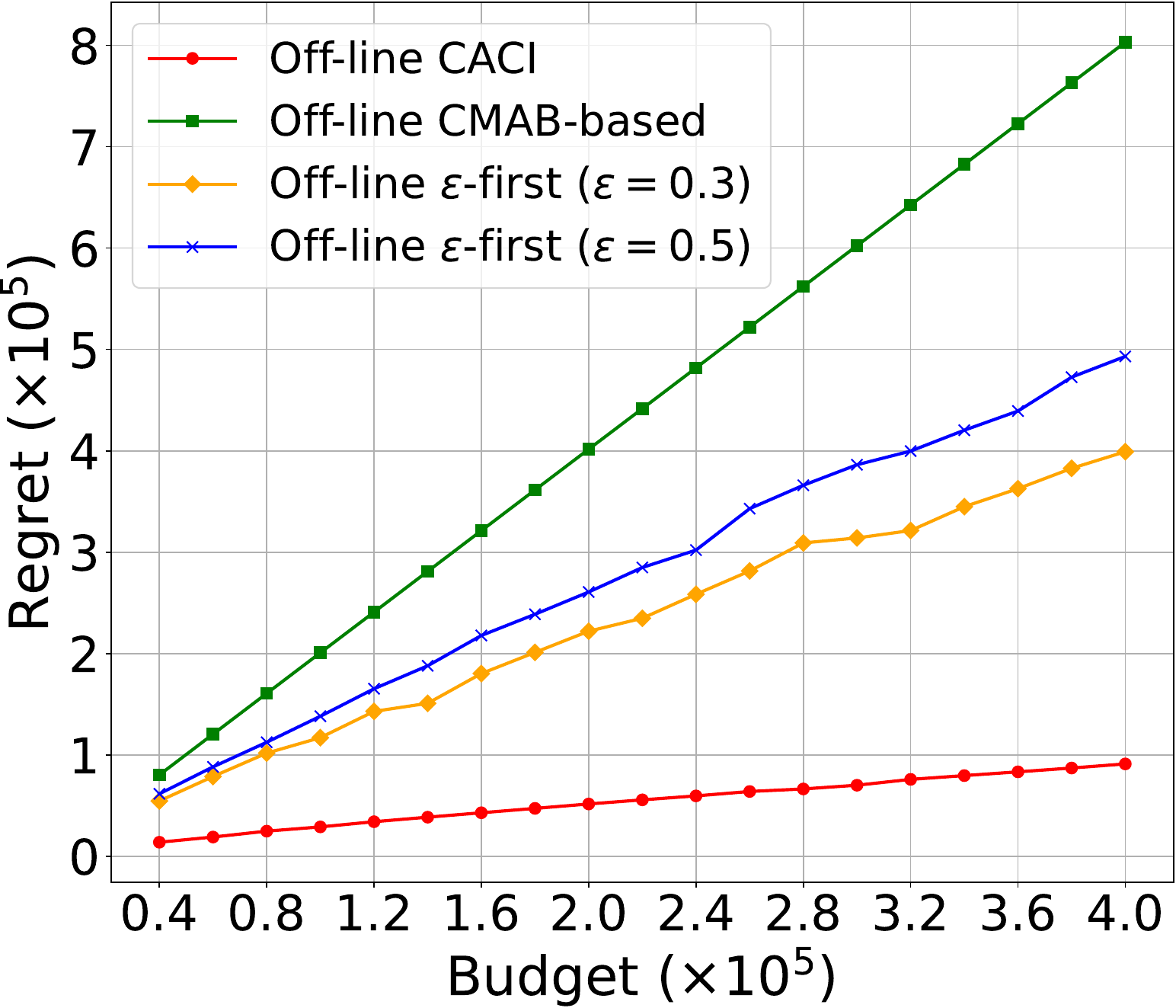}}
      \caption{Comparison results under different budgets on synthetic dataset.}
      \label{fig:offsyn-budget}
      \vspace{-3ex}
      \end{figure}

      We first study the performance of the different mechanisms in terms of cumulative reward (the total actual reward obtained in practice) and report the results in Fig.~\ref{fig:offsyn-budget} (a). We vary the budget from $4 \time 10^4$ to $4 \times 10^5$ with s step size of $2 \times 10^4$. It is worthy to notice that the budget is only four times higher (or even much smaller) than the number of workers. It is shown in Fig.~\ref{fig:offsyn-budget} (a) that, we obtain much more cumulative reward through our CACI mechanism than the reference ones (including the CMAB-based and the $\varepsilon$-first mechanisms), since the contexts of the workers are fully utilized to estimate the their sensing abilities even we do not have sufficient budget. Compared with the reference mechanisms, the cumulative reward yielded by CACI is very close to the one obtained through the ideal basedline mechanism, especially when the budget is more limited. By increasing the budget gradually, although the budget is still significantly limited, the advantage of our mechanism over the others becomes more oblivious, since we are allowed to partition the context space with a finer gradunarity under the increased budget and thus more accurately estimate the workers' sensing abilities according to their contexts. We also report the regrets of the different mechanisms in Fig.~\ref{fig:offsyn-budget} (b). It is apparent that, our CACI mechanism has much lower regret than the other three alternatives. When the budget is increased, our algorithm proceeds more iterations such that the regret is increased but at a very low rate, which is consistent with our theoretical result in \textbf{Theorem}~\ref{thm:regret}.

\begin{figure}
      \centering
        \subfigure[Cumulative reward vs. \# of workers]{\label{fig:offsyn-reward-numwk}\includegraphics[width=.49\columnwidth]{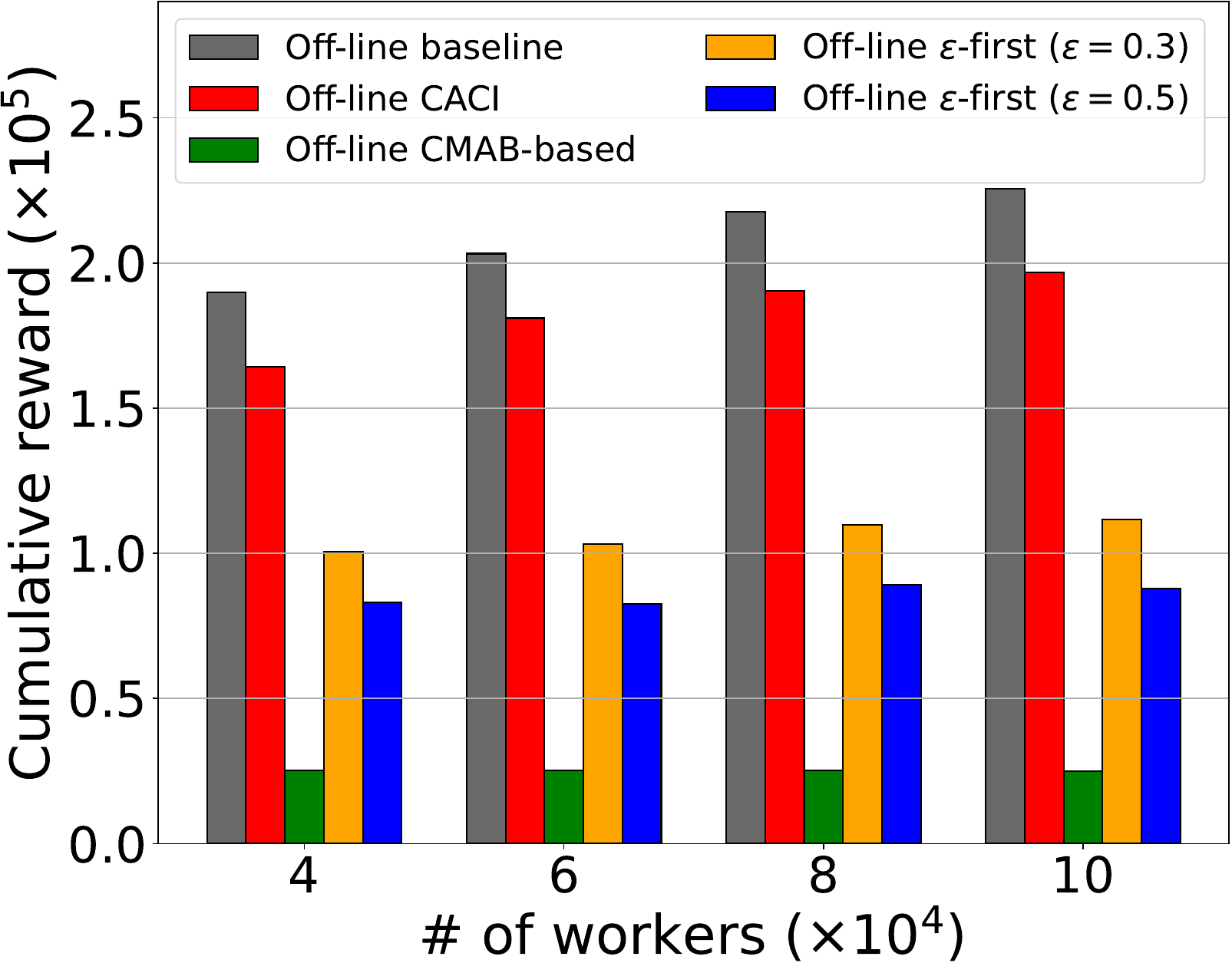}}
        \subfigure[Regret vs. \# of workers]{\label{fig:offsyn-regret-numwk}\includegraphics[width=.49\columnwidth]{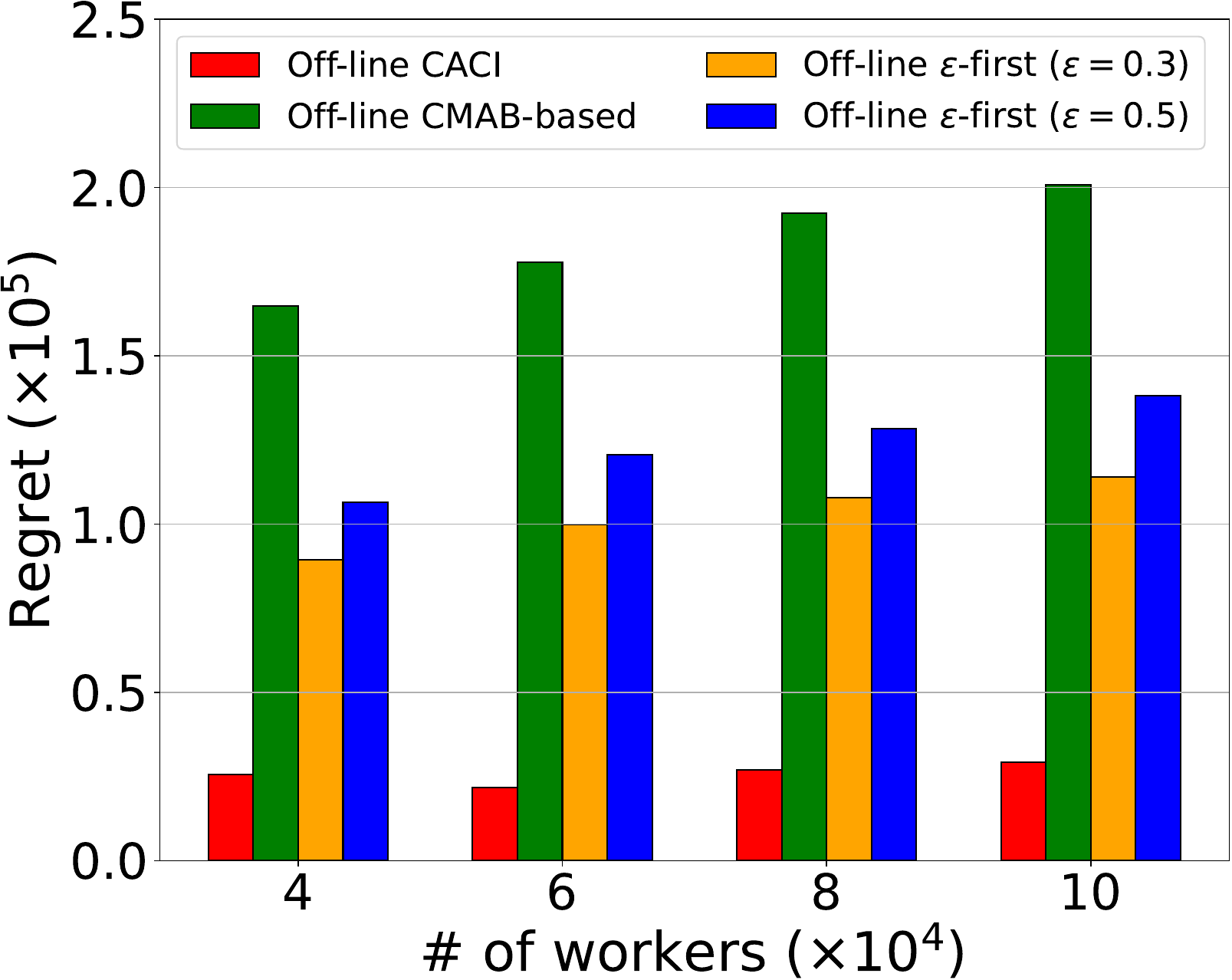}}
      \caption{Comparison results with different numbers of workers on synthetic dataset.}
      \label{fig:offsyn-numwk}
      \vspace{-3ex}
      \end{figure}

      We then fix the total budget $B=10^5$ and vary the number of workers $N=4, 6, 8, 10 \times 10^4$ to show the performance of the different algorithms with respect to an increasing number of workers. In Fig.~\ref{fig:offsyn-numwk}, we report the cumulative rewards and the regrets of the different mechanisms. It is revealed in Fig.~\ref{fig:offsyn-numwk}~(a) that, no matter how many unknown workers are given, our CACI mechanism yields much more cumulative reward. Specifically, when there are $10^5$ workers, the cumulative reward of our CACI mechanism is around eight times higher than the one of the CMAB-based one and two times higher than the $\varepsilon$-first mechanism. As there may be more elite workers
      participating in the sensing task when the total number of the workers is increased, the basline mechanism fully exploits the elite workers and thus results in more cumulative reward. Since our CACI mechanism can utilize limited budget to learn the sensing abilities of the workers, it also yields more cumulative reward under the increased budget, such that the cumulative reward of the CACI mechanism keeps very close to the one of the baseline mechanism.
      Furthermore, as revealed in Fig.~\ref{fig:offsyn-numwk}~(b), the regret of our CACI mechanism is much smaller than the ones of the two reference mechanisms across the different numbers of unknown workers. As the number of workers is increased, the regret of our mechanism is stable (only a very slight increase is observed), since the performance of our mechanism mainly depends on the budget as shown in \textbf{Theorem}~\ref{thm:regret}, while the ones of the other reference mechanisms are increased obliviously, because they all leverage the exploration-exploitation trade-off among the increased number of individual workers whereas the total budget is rather limited.
      % %
      % %

\begin{figure}[htb!] 
      \begin{center}
        \parbox{.24\textwidth}{\center\includegraphics[width=.24\textwidth]{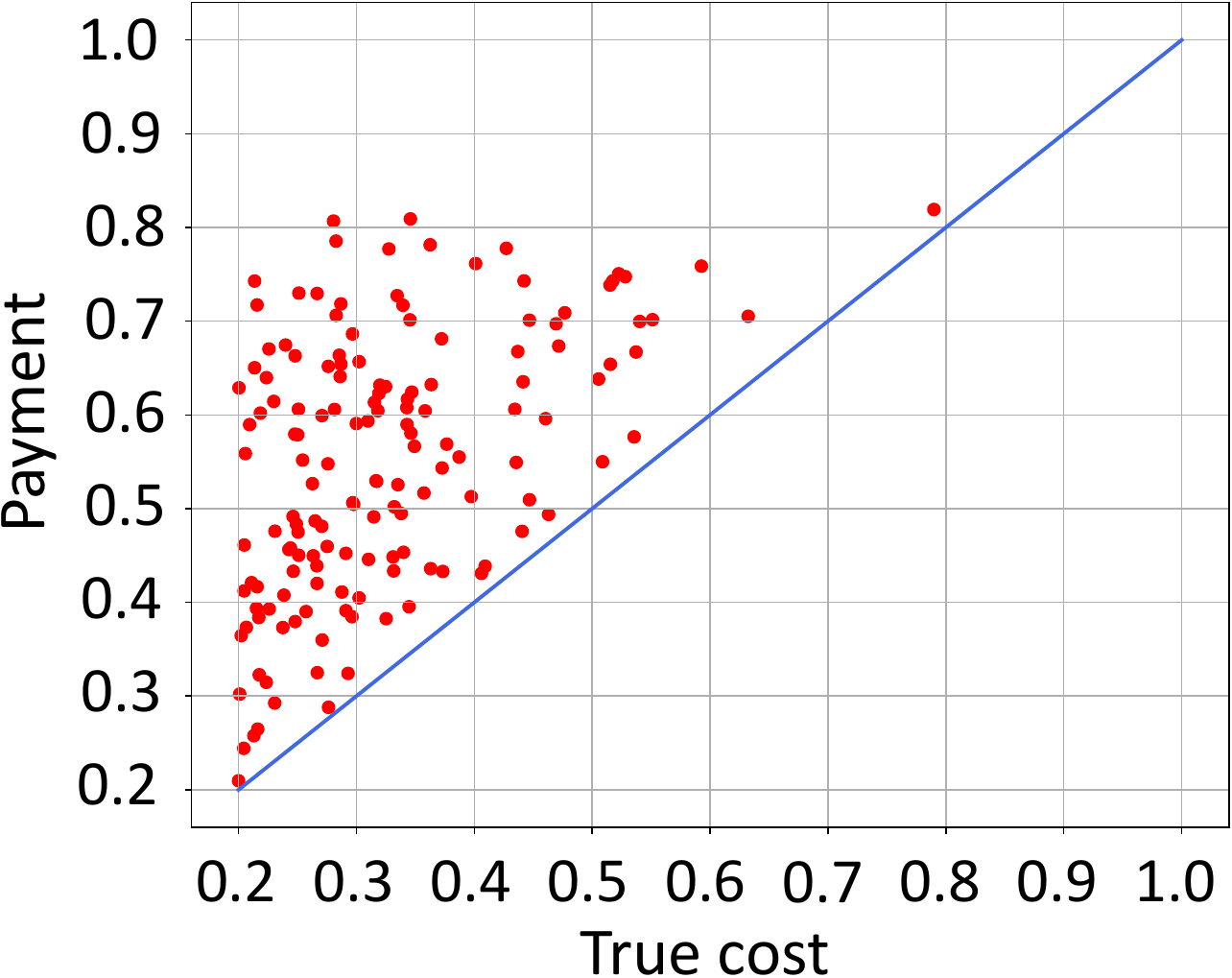}}
        \parbox{.24\textwidth}{\center\includegraphics[width=.24\textwidth]{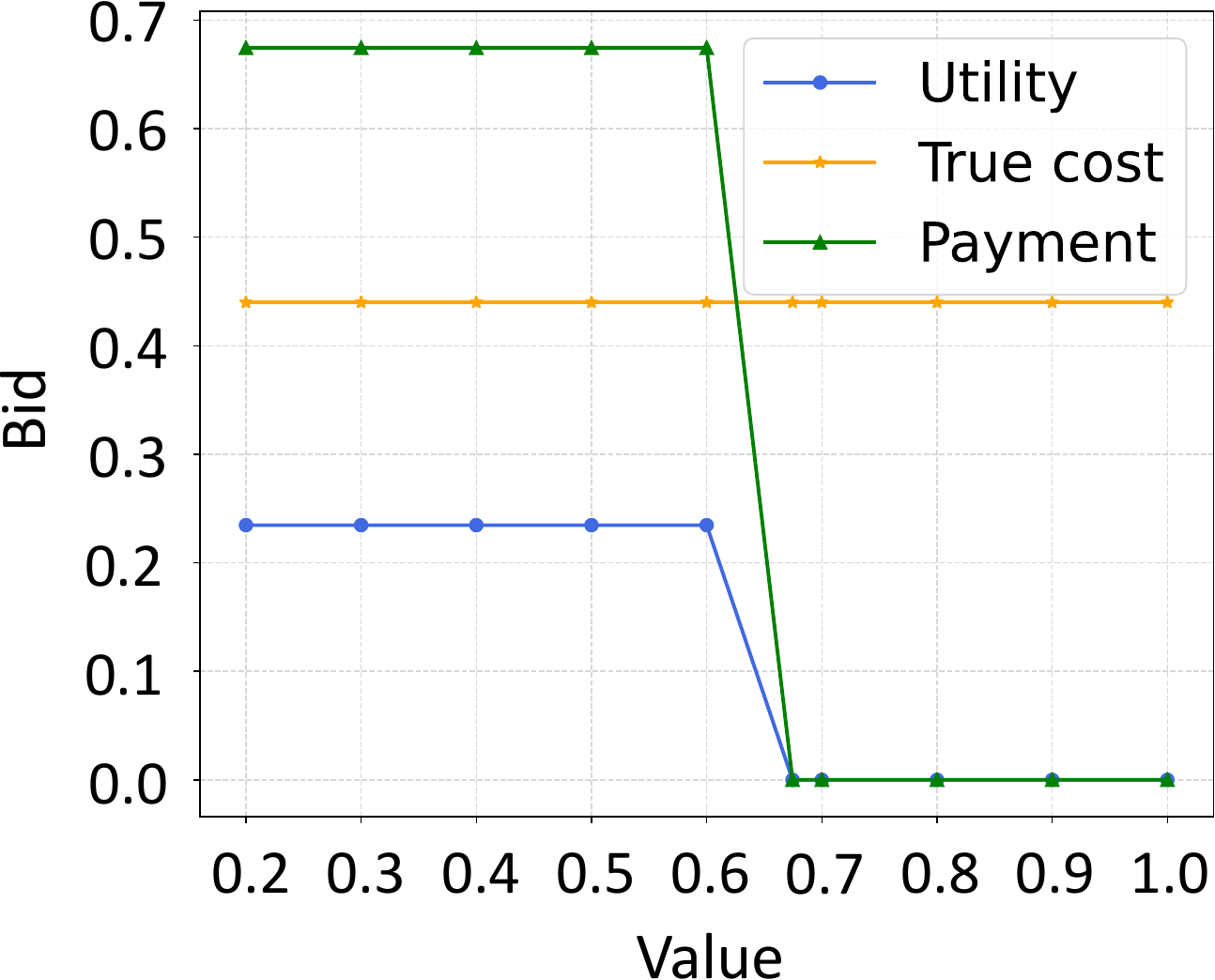}}
        \parbox{.24\textwidth}{\center\scriptsize(a) Truthfulness}
        \parbox{.24\textwidth}{\center\scriptsize(b) Individual rationality}
      \caption{Truthfulness and individual rationality.}
      \label{fig:incentive}
      \end{center}
      \vspace{-4ex}
      \end{figure}

      We also evaluate the truthfulness and the individual rationality of our mechanism. We first illustrate the truthfulness of our mechanism and present the results in Fig.~\ref{fig:incentive} (a). We randomly choose one worker (whose true cost is $0.44$) and change its bid value from $0.2$ to $1.0$ while keeping all the others' bids unchanged. It is illustrated that, the worker obtains fixed (positive) utility unless it proposes a bid higher than the so-called ``critical payment'' ($0.67$ in our case). Therefore, the worker has no motivation to propose an untruthful bid, since doing so does not let it earn more utility. We then evaluate the individual rationality of our mechanism in Fig.~\ref{fig:incentive} (b) where we plot all the payments obtained by each worker. It is found that, for each selected worker in each time slot, its payment is always higher than its true cost. In the following, we do not evaluate the truthfulness and individual rationality, since the results obtained on the real data are similar to the above one.

    \subsubsection{Evaluation with Real Data}  \label{ssec:exp-offcaci-real}
      We hereby implement our off-line CACI mechanism on two real datasets, to reveal its applicability in real scenarios. We first conduct our experiments on a vehicular trajectory dataset, which includes $13,798$ taxicabs' GPS coordinates collected in eight days in Shenzhen, China~\cite{YangWYJ-TITS19,LiKWJ-TITS21}. Each data sample in the dataset consists of a vehicle's license plate number, longitude location, latitude location, etc. We choose the center of the whole area as the venue to conduct sensing tasks. We also choose a time window of five minutes, within which, there are $7,365$ vehicles reporting $54,395$ GPS coordinates. Since distance is usually a main concern for location-based sensing tasks (e.g., air pollution surveillance or noise monitoring) while drivers (or workers) carrying abundantly powered sensor devices (e.g., mobile phones) may more prefer to conduct sensing tasks~\cite{GaoWXC-INFOCOM20,MullerTSK-TON18}, we take into account a 2D context space involving \textit{distance to task spot} and \textit{battery state} as dimensions. In particular, each worker estimates its prospective trajectory in our time window and report the center of the trajectory to the crowdsensing platform. We suppose the battery state of each worker obeys a uniform distribution in $[0,1]$. Given a worker with context $s = (s_0, s_1)$ where $s_0$ and $s_1$ denote the distance to the task spot and the battery state, we define its sensing ability as $\mathbb{E}[r(s)] = \frac{1}{\sigma}\sqrt{\frac{s_1}{2\pi}} \exp \left( -\frac{s^2_0}{2\sigma^2} \right)$, by borrowing the idea from \cite{MullerTSK-TON18}. We let $\sigma=1$ such that the H$\ddot{\text{o}}$lder condition holds for $\alpha=3.5$ to facilitate our quantitative analysis. We then normalize the workers' sensing abilities into $[0,1]$. Additionally, we let $K=100$ and adopt the same setting for the true costs and bids of the workers as before.

\begin{figure}
      \centering
        \subfigure[Cumulative reward vs. budget]{\label{fig:offtaxi-reward-udget}\includegraphics[width=.49\columnwidth]{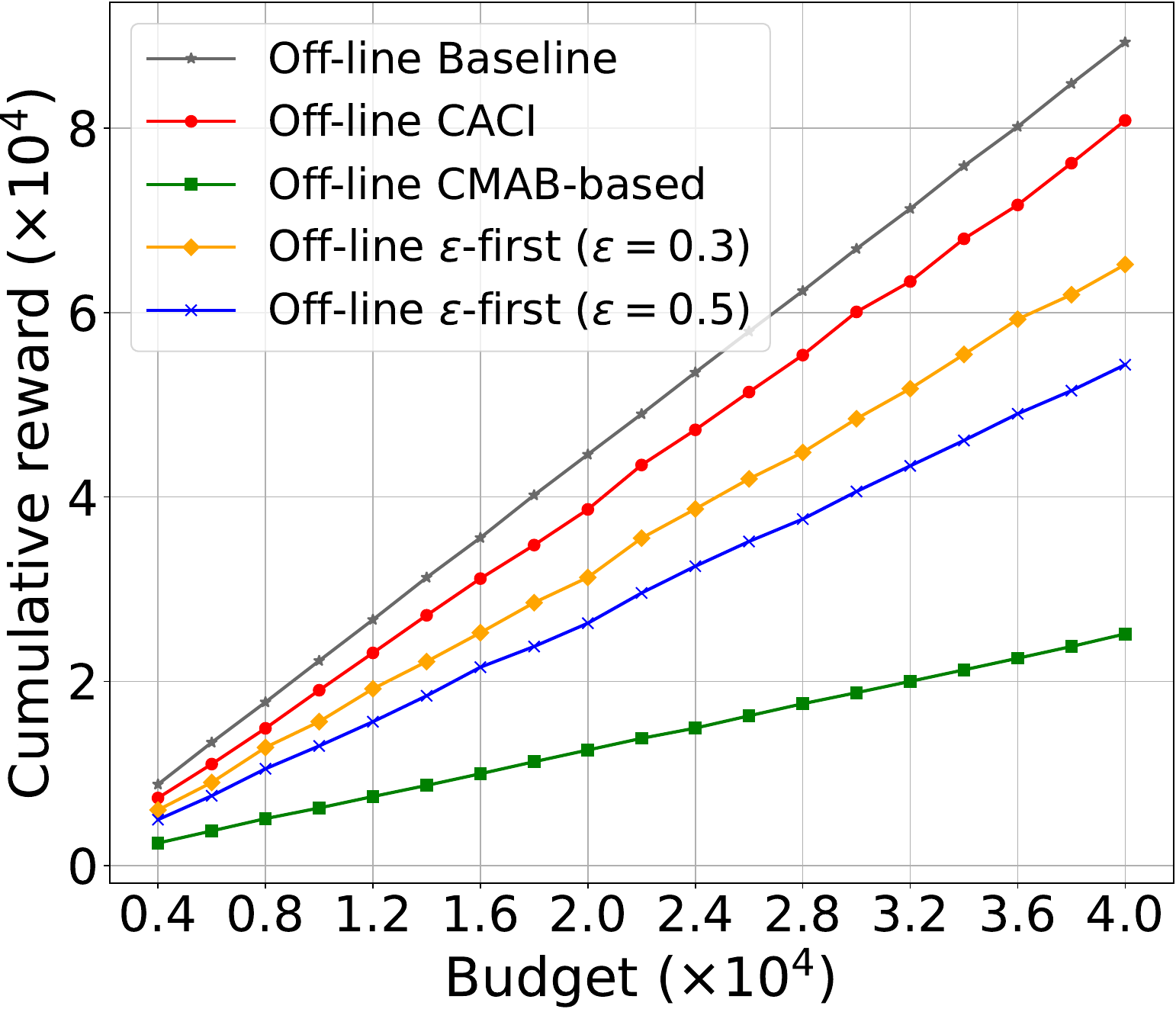}}
        \subfigure[Regret vs. budget]{\label{fig:offtaxi-regret-budget}\includegraphics[width=.49\columnwidth]{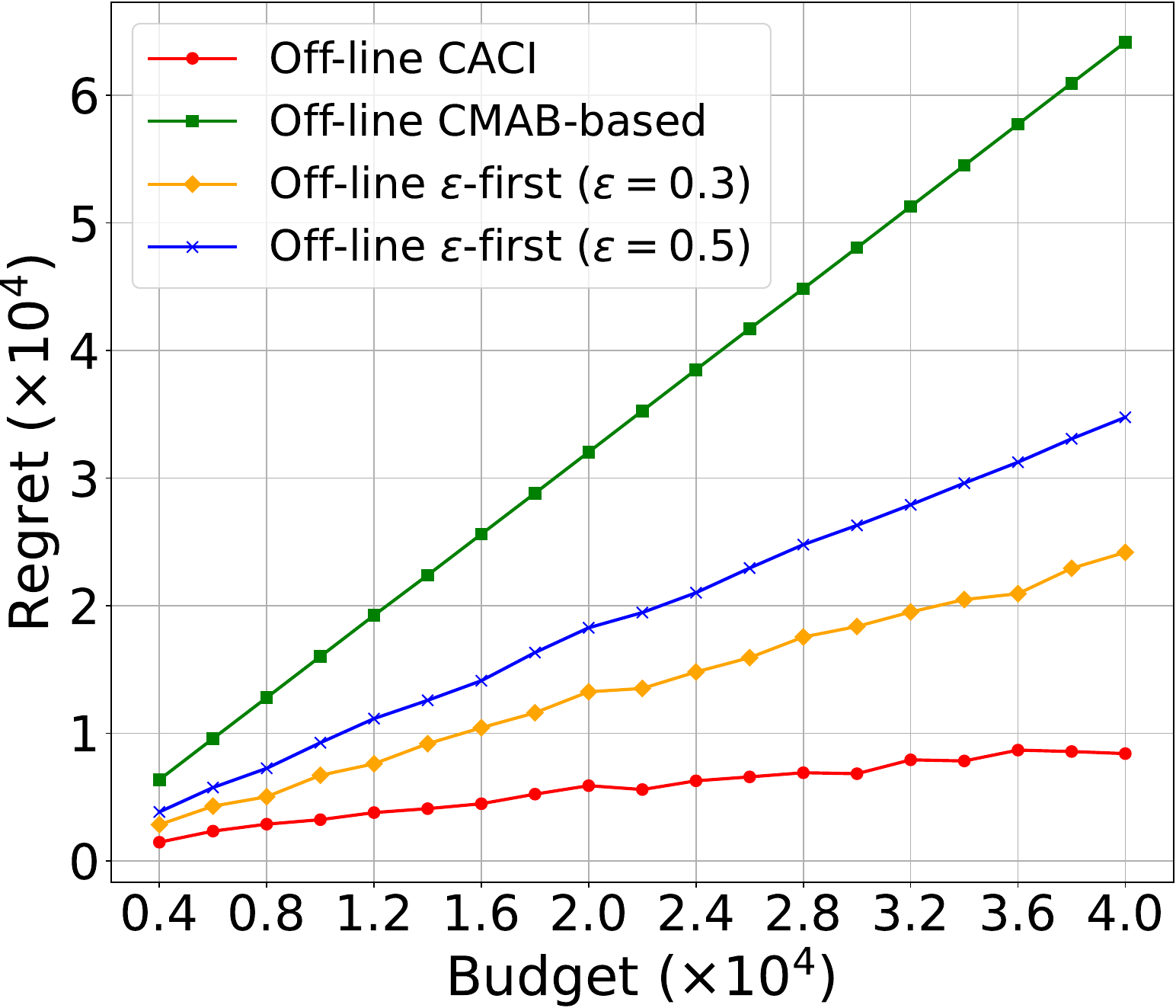}}
      \caption{Comparison results under different budgets on vehicular trajectory dataset.}
      \label{fig:offtaxi-budget}
      \end{figure}

\begin{figure}
      \centering
        \subfigure[Cumulative reward vs. \# of workers]{\label{fig:offtaxi-reward-nwk}\includegraphics[width=.49\columnwidth]{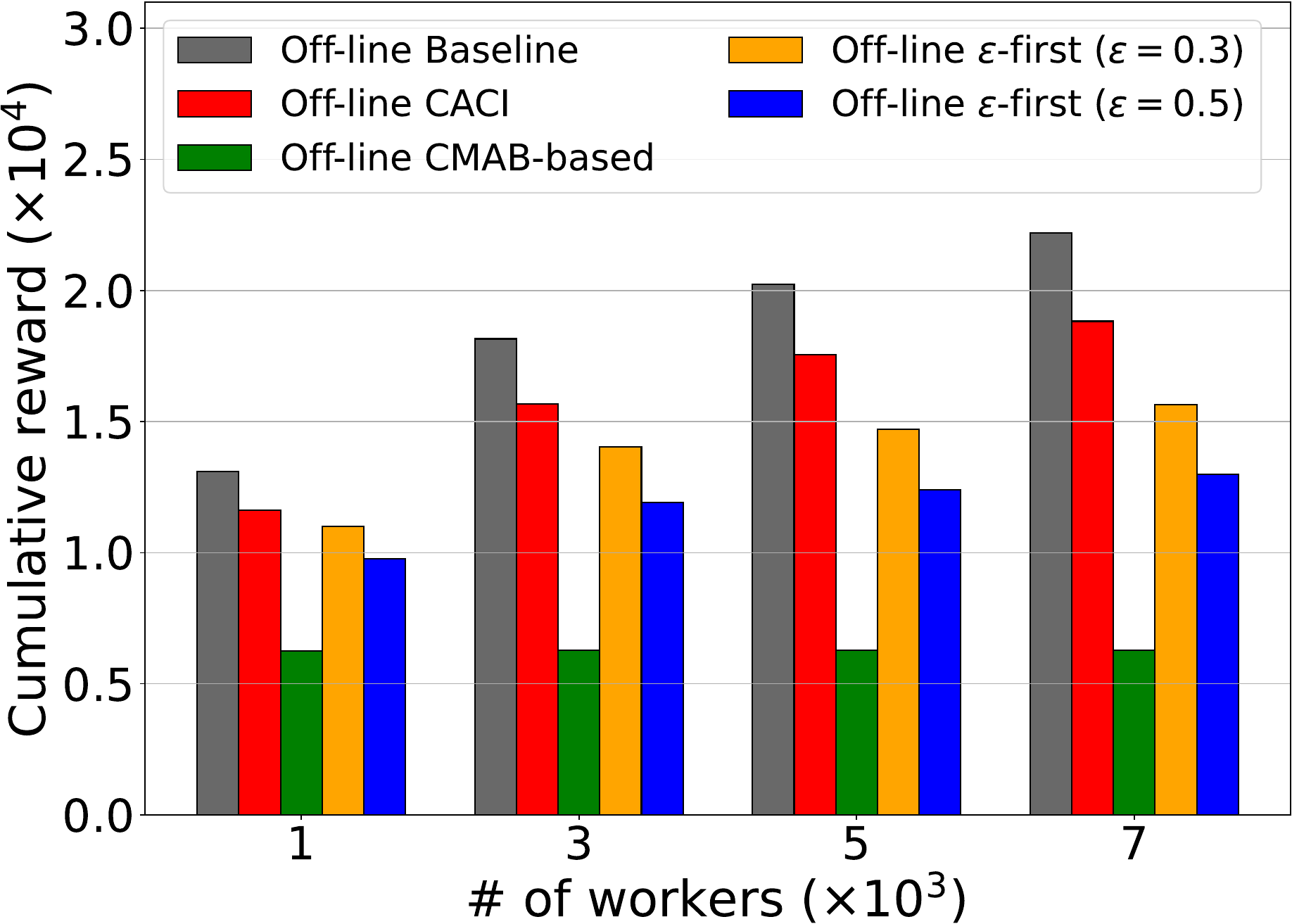}}
        \subfigure[Regret vs. \# of workers]{\label{fig:offtaxi-regret-nwk}\includegraphics[width=.49\columnwidth]{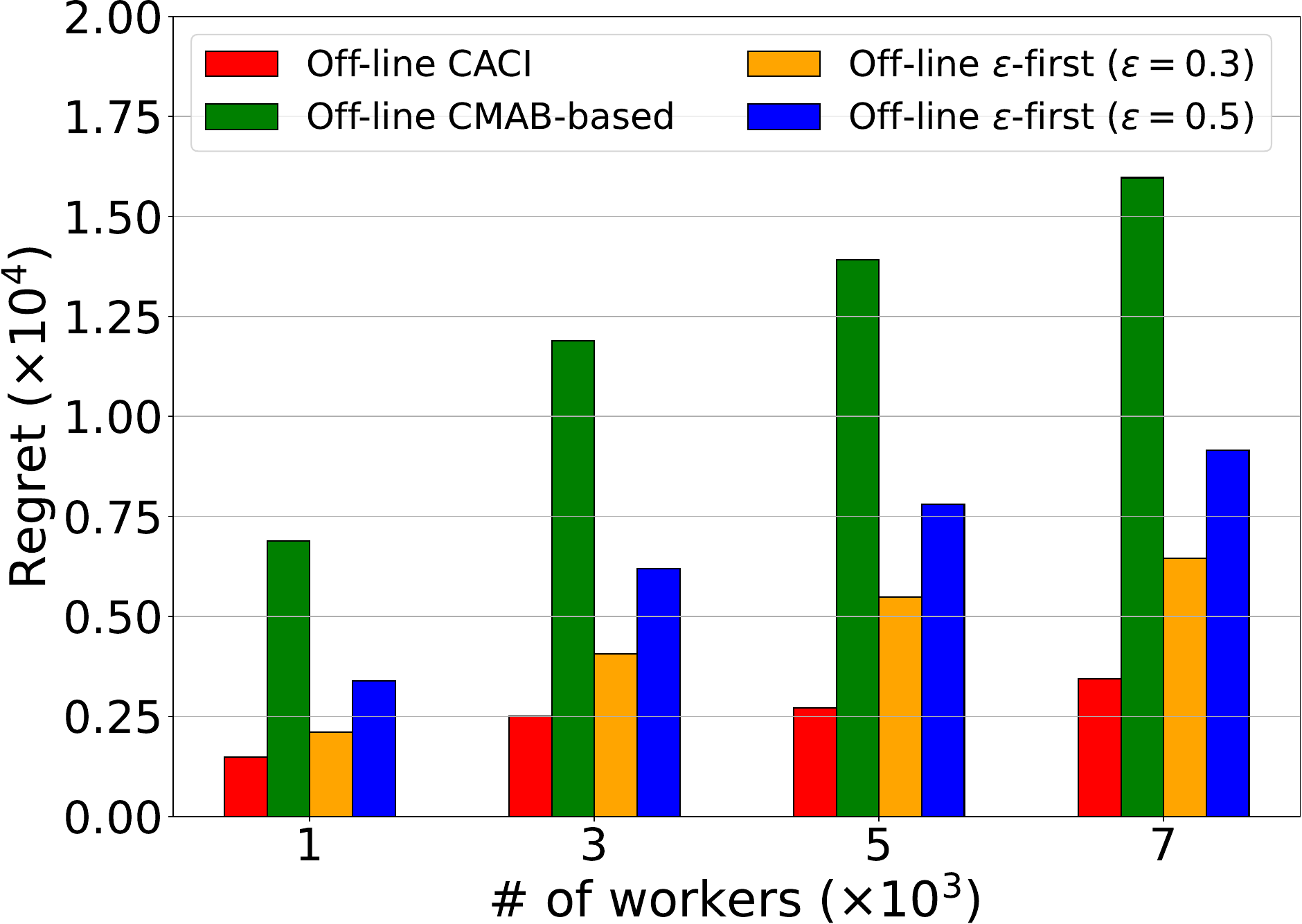}}
      \caption{Comparison results with different numbers of workers on vehicular trajectory dataset.}
      \label{fig:offtaxi-nwk}
      \vspace{-2ex}
      \end{figure}

      In Fig.~\ref{fig:offtaxi-budget}, we report the cumulative rewards and the regrets of the different mechanisms. Likewise, the budget is varied from $4 \times 10^3$ to $4 \times 10^4$. It is demonstrated that, our CACI mechanism yields higher cumulative reward and less regret than the other reference ones. We also present the performance of the different mechanisms with different numbers of workers (i.e., $N = 1, 3, 5, 7 \times 10^3$) in Fig.~\ref{fig:offtaxi-nwk} by fixing budget $B=10^4$. Unsurprisingly, regardless of how many unknown workers are given, our CACI mechanism always have better performance (i.e., more cumulative reward and less regret) than the reference ones and its advantage is more oblivious when there are more workers. Moreover, similar to what we have shown in Sec.~\ref{ssec:exp-offcaci-syn}, when the number of workers is increased, our mechanism has ``stable'' performance. Specifically, it has its cumulative reward increased proportionally along with the baseline mechanism and hence its regret grows very slightly.
      % %
      % \begin{figure}[htb!]
      % \begin{center}
      %   \parbox{.8\columnwidth}{\center\includegraphics[width=.75\columnwidth]{fig/offig/reward-budget-taxi-off.pdf}}
      %   \parbox{.8\columnwidth}{\center\scriptsize(a) Cumulative reward vs. budget}
      %   %
      %   \parbox{.8\columnwidth}{\center\includegraphics[width=.75\columnwidth]{fig/offig/regret-budget-taxi-off.pdf}}
      %   \parbox{.8\columnwidth}{\center\scriptsize(b) Regret vs. budget}
      % \caption{Comparison results under different budgets on vehicular trajectory dataset.}
      % \label{fig:offtaxi-budget}
      % \end{center}
      % \end{figure}
      % %
      % %
      % \begin{figure}[htb!]
      % \begin{center}
      %   \parbox{.8\columnwidth}{\center\includegraphics[width=.75\columnwidth]{fig/offig/reward-numwk-taxi-off.pdf}}
      %   \parbox{.8\columnwidth}{\center\scriptsize(a) Cumulative reward vs. \# of workers}
      %     %
      %   \parbox{.8\columnwidth}{\center\includegraphics[width=.75\columnwidth]{fig/offig/regret-numwk-taxi-off.pdf}}
      %   \parbox{.8\columnwidth}{\center\scriptsize(b) Regret vs. \# of workers}
      % \caption{Comparison results with different numbers of workers on vehicular trajectory dataset.}
      % \label{fig:offtaxi-nwk}
      % \end{center}
      % \end{figure}
      % %

      We also evaluate the performance of our mechanism in a crowdsensing application based on the dataset published by Yelp~\cite{yelp}. We randomly choose $10^5$ workers from the dataset. The settings for the workers' true costs and bids are the same as the ones in Sec.~\ref{ssec:exp-offcaci-syn}. In the Yelp dataset, the sensed data (i.e., the reviews of the workers on business) is voted by reviewers. For each of the sensed data, we assume that we get a unit of reward if it receives at least three positive votes. We choose \textit{number of fans}, \textit{number of friends} and \textit{number of years as elite} as the context dimensions, considering their strong correlations to the data quality. In our experiments, we gradually increase the dimensionality of the context space, to evaluate our algorithm. Since the Yelp dataset does not include the true sensing abilities of the workers, we are interested in investigating the performance of the different algorithms in terms of cumulative reward only. For each worker, when it is selected, we randomly choose one from its data samples without replacement to calculate the cumulative reward.

\begin{figure}[htbp]
\centering
\begin{minipage}[t]{0.48\columnwidth}
\centering
\includegraphics[width=0.95\columnwidth]{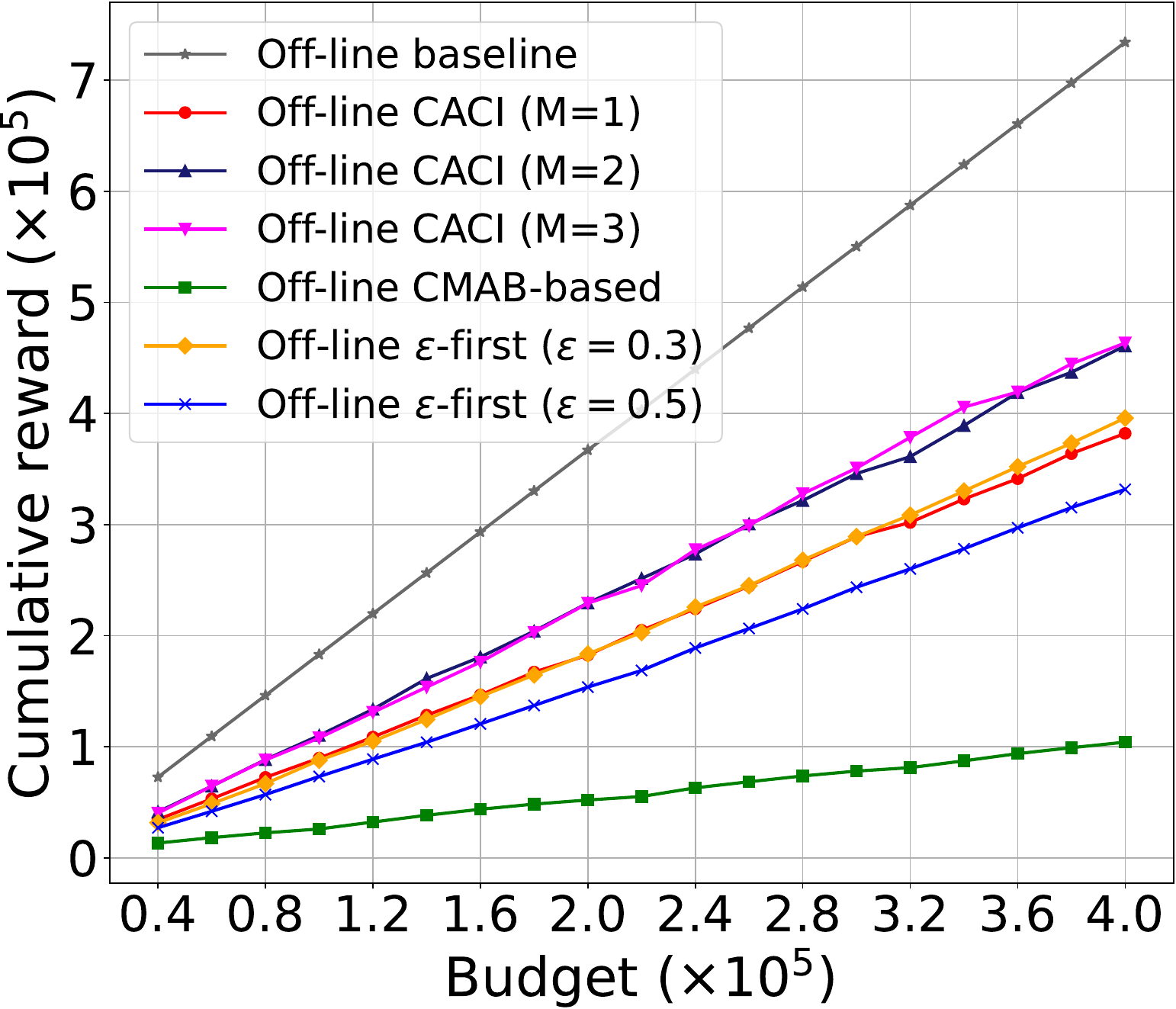}
\caption{Comparison results with different budgets on Yelp dataset.}
\label{fig:offyelp-budget}
\end{minipage}
\begin{minipage}[t]{0.48\columnwidth}
\centering
\includegraphics[width=0.98\columnwidth]{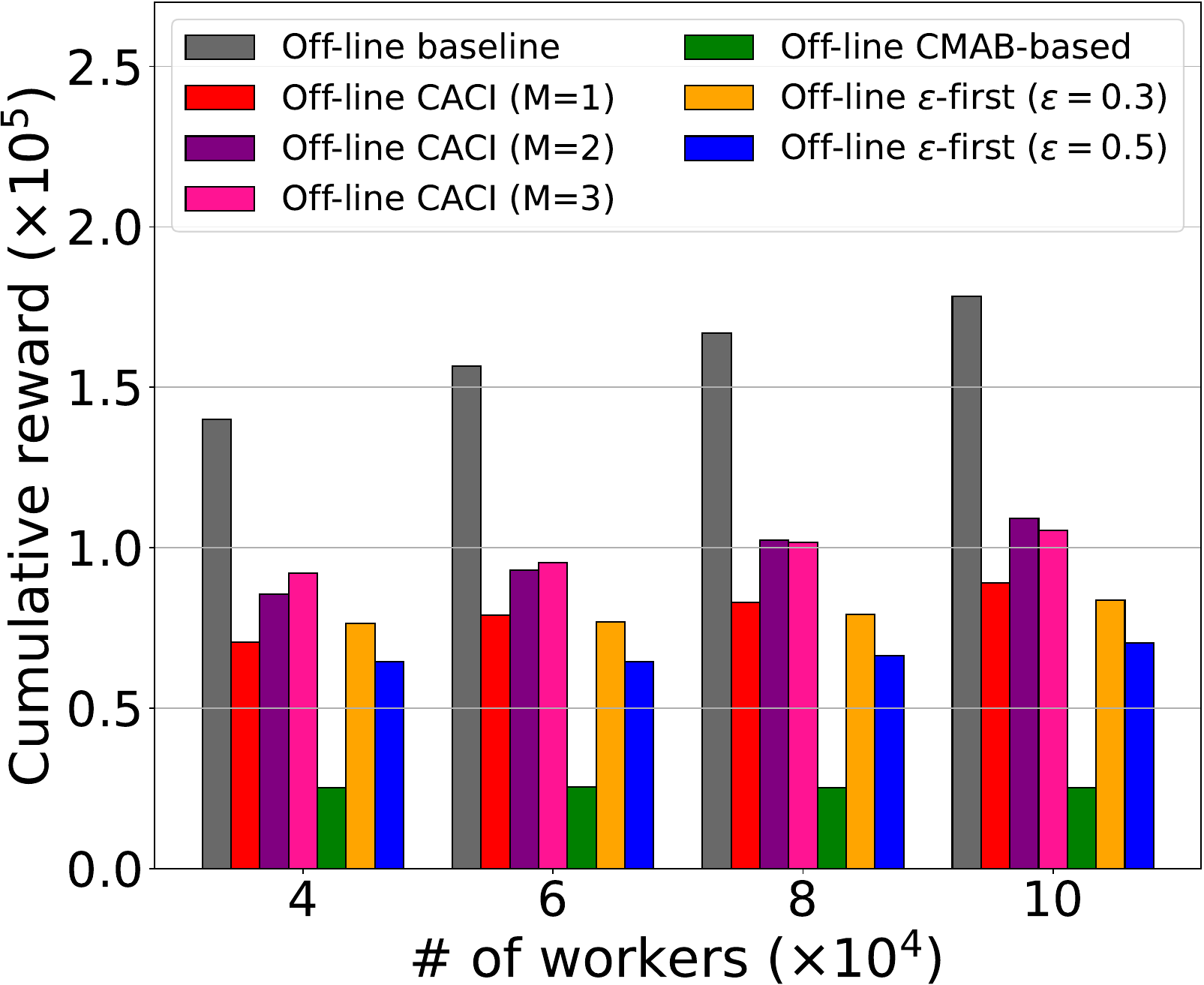}
\caption{Comparison results with different numbers of workers on Yelp dataset.}
\label{fig:offyelp-diffwk}
\end{minipage}
\end{figure}

      In Fig.~\ref{fig:offyelp-budget}, we report the cumulative rewards with the total budget varying from $0.4$ to $4.0 \times 10^5$. It is illustrated that, our mechanism outperforms the others and its performance is closer to the one of the baseline mechanism especially for $M=2,3$. By taking into more relevant dimensions (e.g., by increasing $M$ from $1$ to $3$), our mechanism yields more cumulative reward. Nevertheless, a higher-dimensional context space does not imply much higher cumulative reward. For example, the  cumulative reward obtained by our mechanism in the 3D context space is very close to the one in the 2D context space. We also evaluate the scalability of the mechanisms to the different numbers of workers. We fix budget $B=10^5$ while varying the number of workers from $4$ to $10 \times 10^4$. As demonstrated by the results in Fig.~\ref{fig:offyelp-diffwk}, our CACI mechanism yields much more cumulative reward than the others in all settings. Moreover, similar to the baseline mechanism, even there are more unknown workers participate in the crowdsensing, our mechanism yields more cumulative reward, since we partition the context space with a carefully tuned granularity under limited budget. 
      %
      % \begin{figure}[htb!]
      % \begin{center}
      %   \includegraphics[width=.75\columnwidth]{fig/offig/reward-budget-yelp-off.pdf}
      % \caption{Comparison results with different budgets on Yelp dataset.}
      % \label{fig:offyelp-budget}
      % \end{center}
      % \end{figure}
      % %
      % \begin{figure}[htb!]
      % \begin{center}
      %   \includegraphics[width=.75\columnwidth]{fig/offig/reward-numwk-yelp-off.pdf}
      % \caption{Comparison results with different numbers of workers on Yelp dataset.}
      % \label{fig:offyelp-diffwk}
      % \end{center}
      % \end{figure}
      % %

  \subsection{Evaluation Results of On-Line CACI Mechanism}  \label{ssec:exp-oncaci}
    In this section, we evaluate our on-line CACI mechanism on both the synthetic and the real datasets.
    \subsubsection{Evaluation on Synthetic Data}  \label{sssec:exp-oncaci-syn}
      We assume there are $200$ available workers in each time slot and the workers across different time slots are distinct. We let $K=10$ such that we have to recruit $10$ workers in each time slot. The workers have their contexts uniformly distributed in the context space $\mathcal S \subseteq [0,1]^{M=2}$. We adopt the same settings for the workers' costs and bids. Additionally, we randomly set the workers' sensing abilities such that the H$\ddot{\mathrm{o}}$lder condition holds for $\alpha=3.5$ for the purpose of quantitative analysis. 

\begin{figure}
      \centering
        \subfigure[Cumulative reward vs. budget]{\label{fig:reward-budget-syn-on}\includegraphics[width=.49\columnwidth]{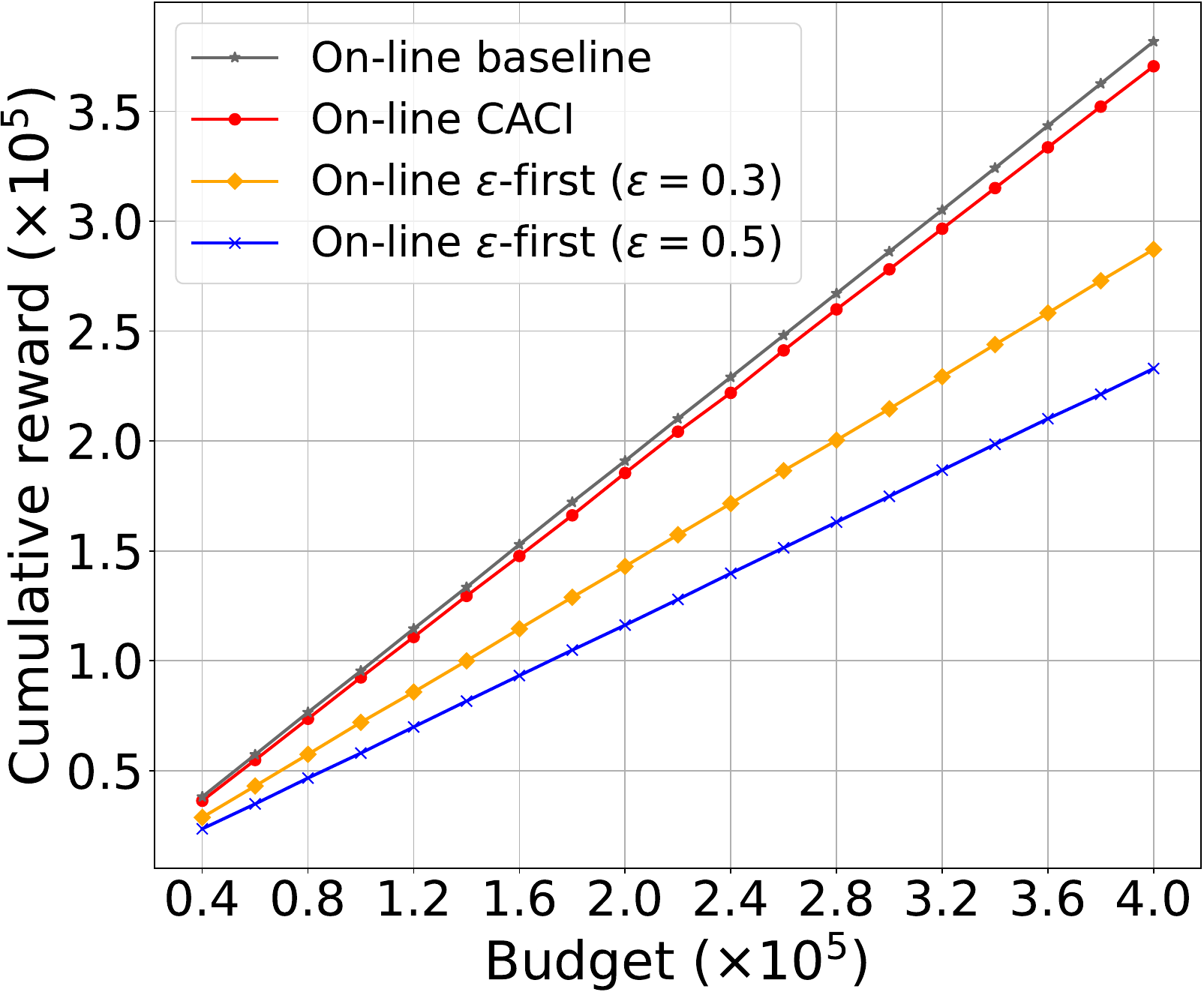}}
        \subfigure[Regret vs. budget]{\label{fig:regret-budget-syn-on}\includegraphics[width=.49\columnwidth]{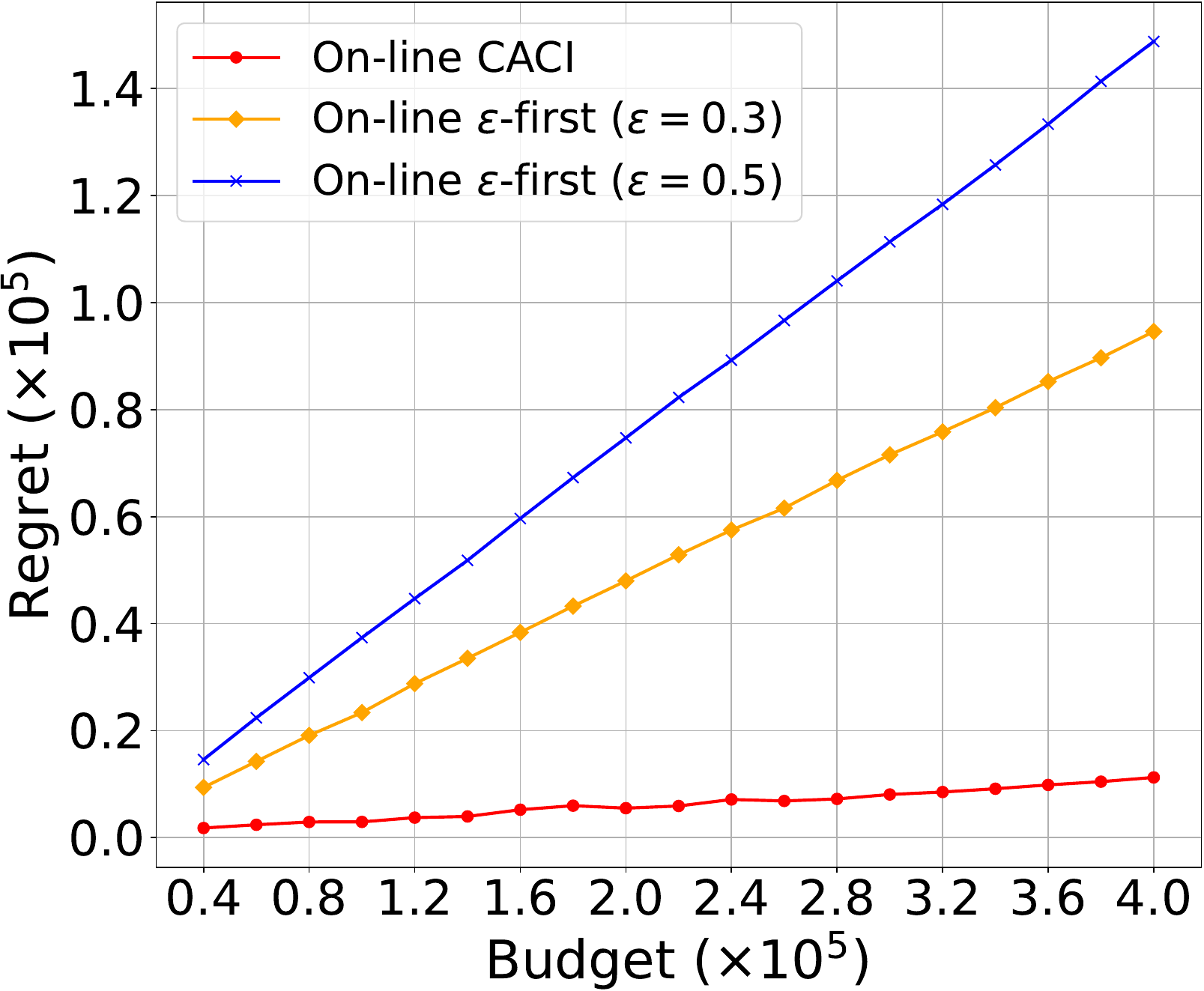}}
      \caption{Comparison results with different budgets on synthetic dataset.}
      \label{fig:onsyn-budget}
      \vspace{-2ex}
      \end{figure}

      The results are reported in Fig.~\ref{fig:onsyn-budget}. Note that the traditional CMAB framework leverages the exploration-exploitation trade-off among the workers individually and thus cannot be used to address our on-line $K$-WIN problem; therefore, we hereby compare our on-line CACI mechanism with the on-line variant of the $\varepsilon$-first mechanism only (where we choose $\varepsilon = 0.3, 0.5$). It is observed in Fig.~\ref{fig:onsyn-budget} that, compared with the $\varepsilon$-first mechanism, our CACI mechanism has much more cumulative reward and thus smaller regret, across the different budgets from $0.4$ to $4 \times 10^5$. It is noticeable that, the cumulative reward yielded by our CACI mechanism is very close to the one of the baseline mechanism, while the regret is increased very slightly when more budget is spent. According to these observations, although \textbf{Theorem}~\ref{thm:onregretbd} implies that the regret of our on-line CACI mechanism is bounded by $O(B)$, the increase of the regret is actually very marginal and thus our on-line CACI mechanism yields cumulative reward comparable to the one of the baseline mechanism.

    \subsubsection{Evaluation on Real Data}  \label{sssec:exp-oncaci-real}
      We first implement our on-line CACI mechanism on the vehicular trajectory dataset. Similarly, we assume the center of the whole area is the venue to conduct sensing tasks and there are around $200$ drivers (or workers) reporting their bids and related contexts (i.e., their distances to the task spot and the battery states of their sensor devices) in each time slot. We also adopt the same function to map the context space to the expected reward (i.e., $\mathbb{E}[r(s)]$ shown in Sec.~\ref{ssec:exp-offcaci-real}). We adopt the same setting for the true costs and bids of the workers as the ones employed in Sec.~\ref{ssec:exp-offcaci-syn}. We let $\sigma=1$ such that the H$\ddot{\text{o}}$lder condition holds for $\alpha=2$ and suppose $K=10$.

\begin{figure}
      \centering
        \subfigure[Cumulative reward vs. budget]{\label{fig:ontaxi-reward-budget}\includegraphics[width=.47\columnwidth]{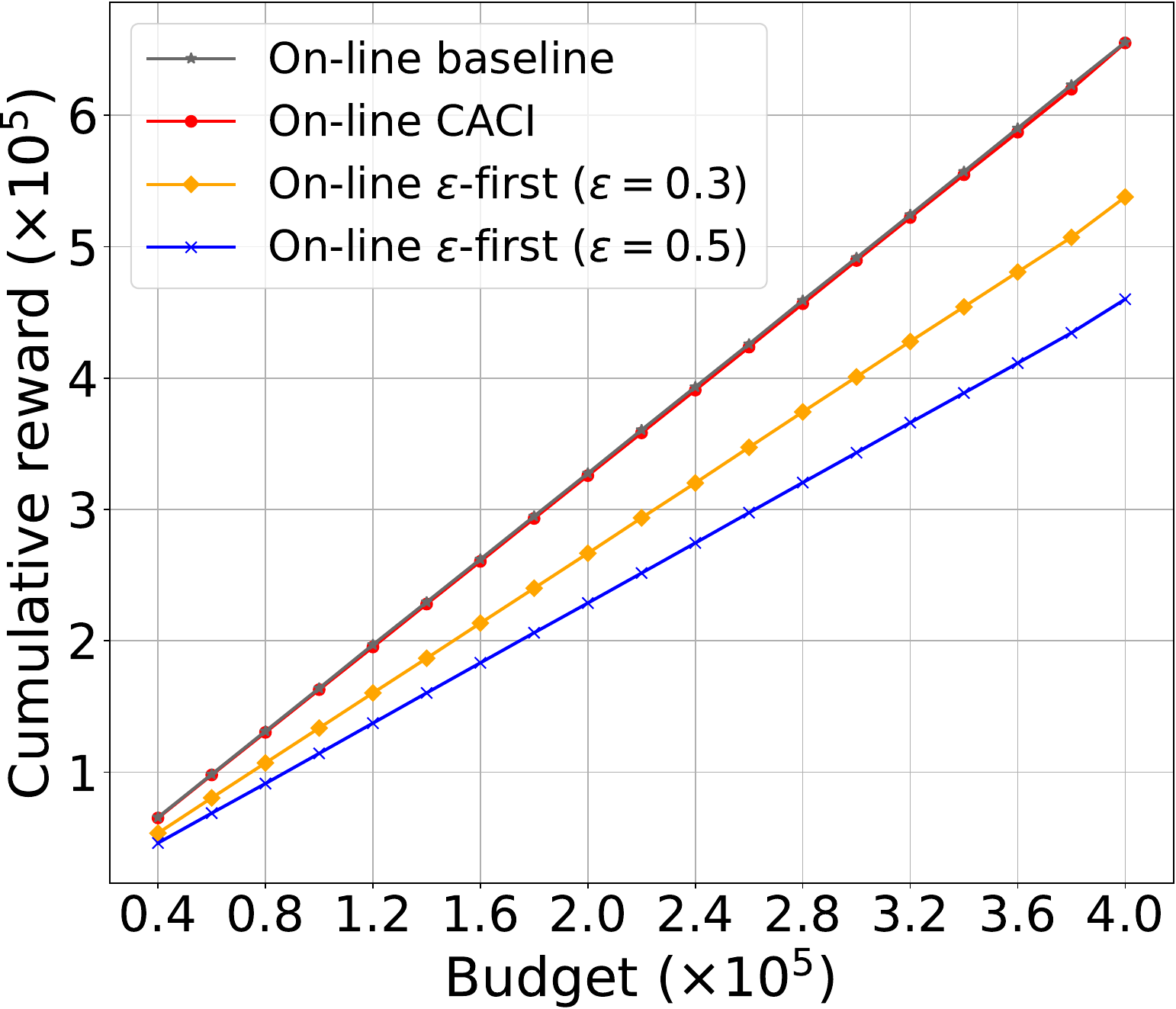}}
        \subfigure[Regret vs. budget]{\label{fig:ontaxi-regret-budget}\includegraphics[width=.49\columnwidth]{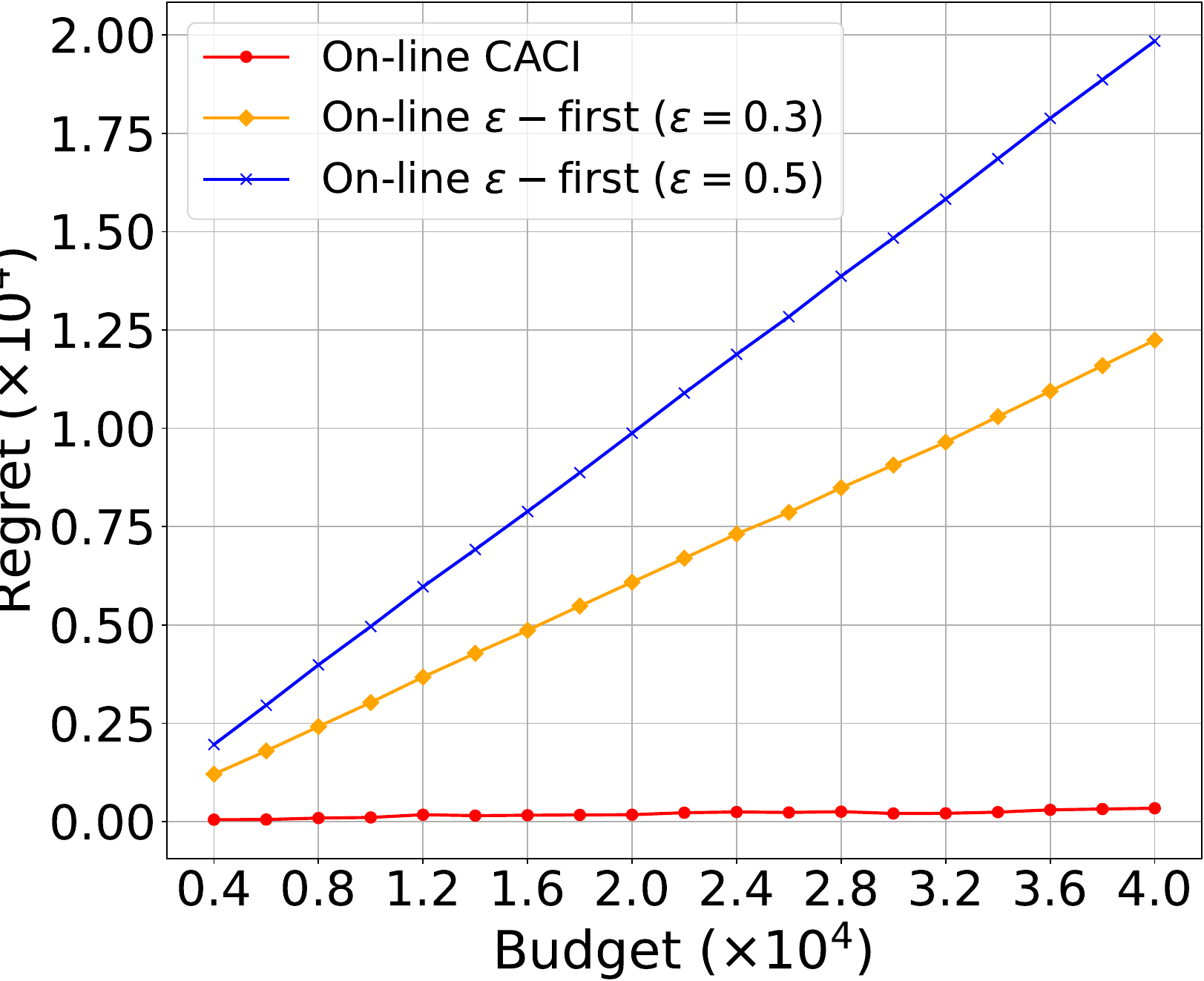}}
      \caption{Comparison results with different budgets on vehicular trajectory dataset.}
      \label{fig:ontaxi-budget}
      \vspace{-3ex}
      \end{figure}

      The performances of our on-line CACI mechanism and the reference ones are presented in Fig.~\ref{fig:ontaxi-budget}, where we vary the budget from $0.4$ to $4 \times 10^4$. It is revealed that, the cumulative reward yielded by our on-line CACI mechanism is almost the same as the one output by the baseline mechanism under the different budgets, and is much higher than the one of the on-line $\varepsilon$-first mechanism. Additionally, the regret of our CACI mechanism across the different budgets is almost zero, whereas the one of the $\varepsilon$-first mechanism grows drastically when the budget is increased.
      % %
      % %

\begin{figure*}[htb!]
      \begin{center}
        \parbox{.32\textwidth}{\center\includegraphics[width=.3\textwidth]{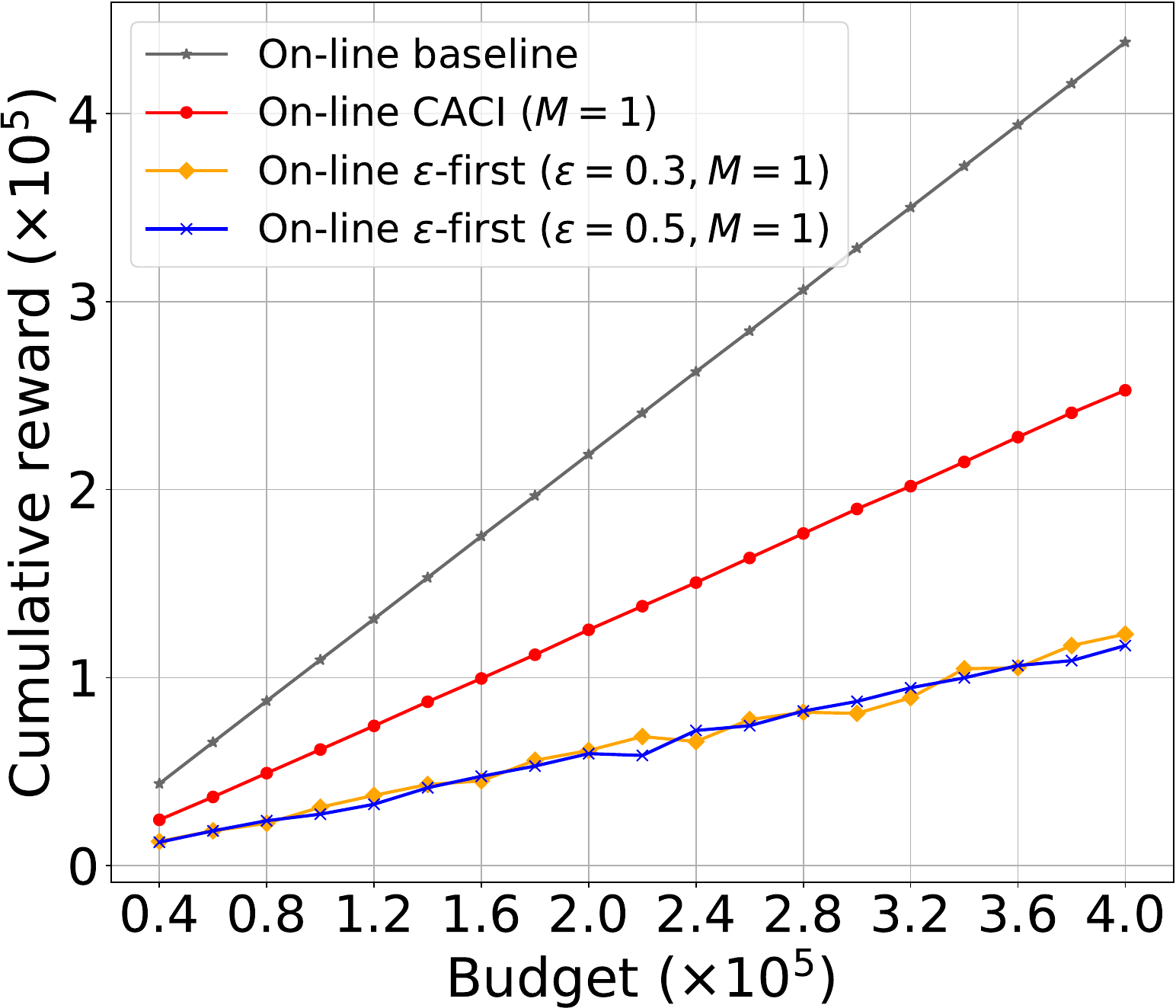}}
        \parbox{.32\textwidth}{\center\includegraphics[width=.3\textwidth]{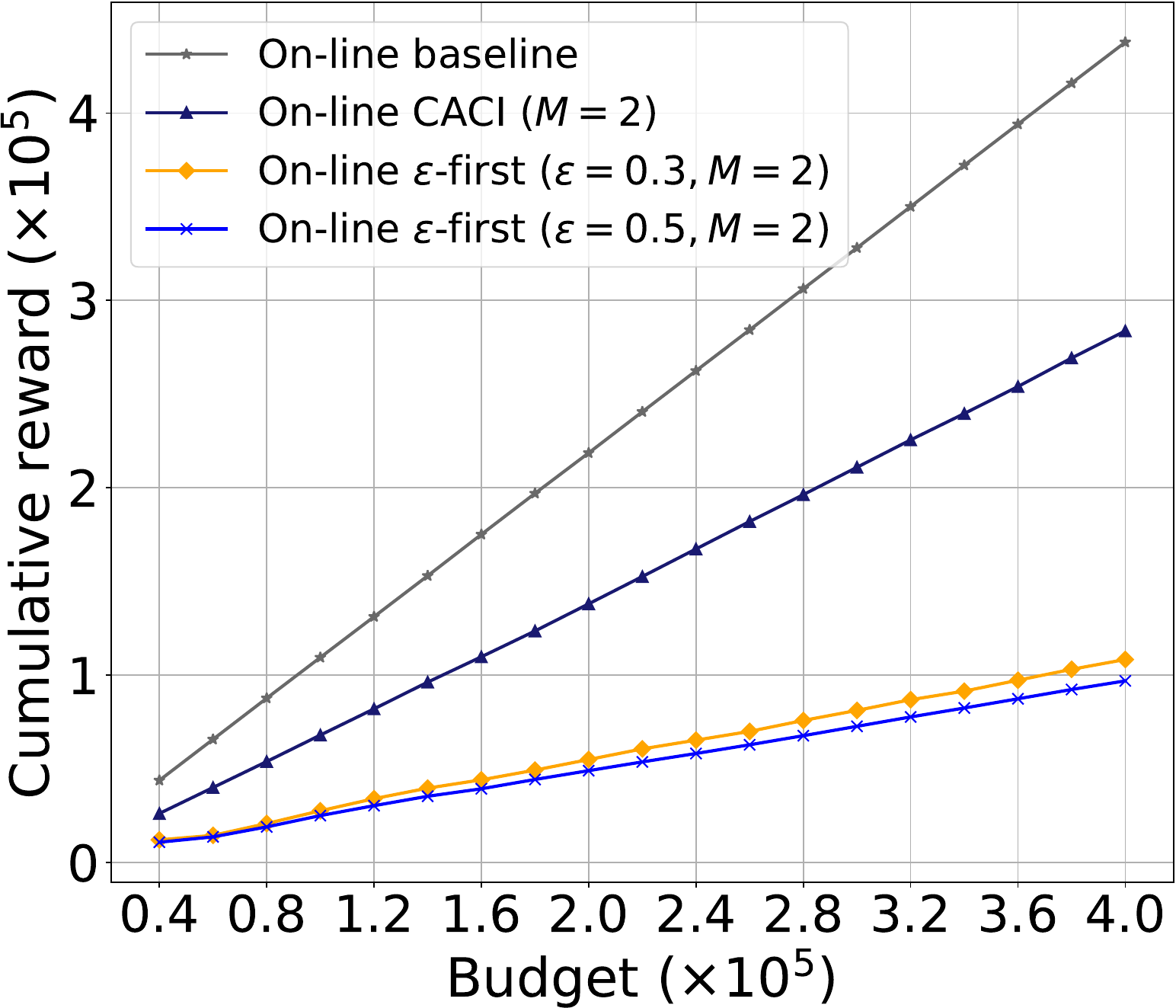}}
        \parbox{.32\textwidth}{\center\includegraphics[width=.3\textwidth]{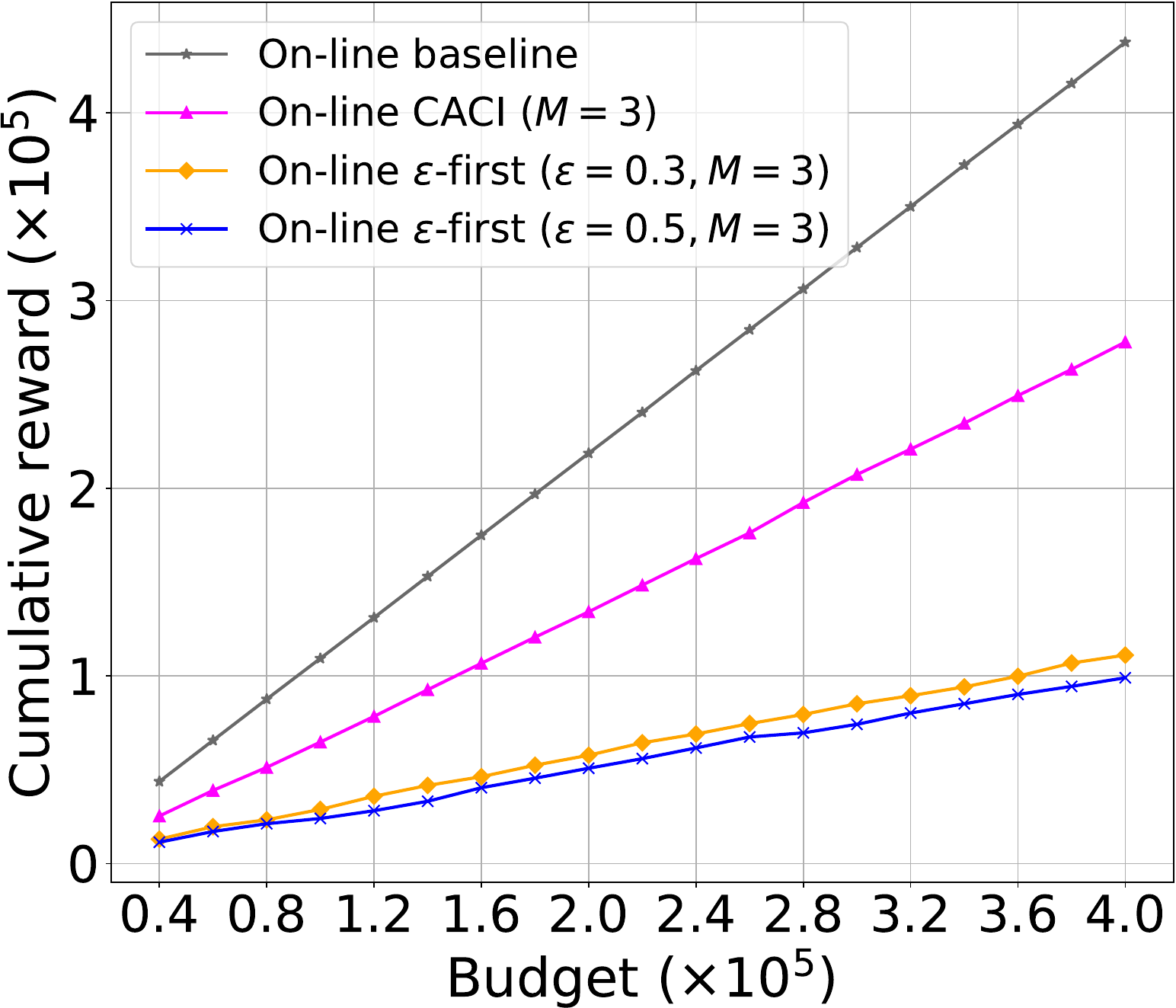}}
        \parbox{.32\textwidth}{\center\scriptsize(a) Cumulative reward vs. budget with $M=1$}
        \parbox{.32\textwidth}{\center\scriptsize(b) Cumulative reward vs. budget with $M=2$}
        \parbox{.32\textwidth}{\center\scriptsize(c) Cumulative reward vs. budget with $M=3$}
        %
        % \parbox{.33\textwidth}{\center\includegraphics[width=.32\textwidth]{fig/onfig/regret-budget-yelp-on-01.pdf}}
        % \parbox{.33\textwidth}{\center\includegraphics[width=.32\textwidth]{fig/onfig/regret-budget-yelp-on-02.pdf}}
        % \parbox{.33\textwidth}{\center\includegraphics[width=.32\textwidth]{fig/onfig/regret-budget-yelp-on-03.pdf}}
        % \parbox{.33\textwidth}{\center\scriptsize(d) Regret vs. budget with $M=1$}
        % \parbox{.33\textwidth}{\center\scriptsize(e) Regret vs. budget with $M=2$}
        % \parbox{.33\textwidth}{\center\scriptsize(f) Regret vs. budget with $M=3$}
      \caption{Comparison results under different budgets on Yelp dataset.}
      \label{fig:onyelp-budget}
      \end{center}
      \vspace{-3ex}
      \end{figure*}

      We also conduct our evaluation on the Yelp dataset. In particular, we randomly choose $200$ workers in each time slot and let $K=10$. We also vary the budget from from $0.4$ to $4.0 \times 10^5$. The other settings are the same as the ones used in Sec.~\ref{sssec:exp-oncaci-real}. The results are presented in Fig.~\ref{fig:onyelp-budget}. It is shown that, our on-line CACI mechanism yields higher cumulative reward than the on-line $\varepsilon$-first mechanism for $M=1,2,3$. Similar to what we have shown in Fig.~\ref{fig:offyelp-budget}, our mechanism yields more cumulative reward, through taking into account more relevant dimensions (e.g., by increasing $M$ from $1$ to $3$). Nevertheless, we may not obtain obvious reward gains in the context space with higher dimension. For example, the cumulative reward obtained in the 3D context space is very close to the one gained in the 2D context space. 

\section{Conclusion}  \label{sec:con}
  In this paper, we have studied how to incentivize a massive number of unknown workers under limited budget. Given a fixed large group of unknown workers, although the standard CMAB framework has been applied to address the uncertainty of the workers by leveraging a exploration-exploitation trade-off among the individual workers, it results in considerable overhead, especially considering the budget may be quite limited whereas the number of the unknown workers may be huge. To address issue, we have proposed our off-line CACI mechanism, which innovates in conducting the exploration-exploitation trade-off in a fine partitioned context space, in stead among the individual workers. We have proved the truthfulness and the individual rationality of our mechanism and revealed an upper bound $\mathcal O \left( B^{\frac{2\alpha+M}{3\alpha+M}} \ln^\frac{1}{3} B \right)$ on its regret. 

  Another challenging issue for incentivizing unknown workers is the dynamics of the workers. They may join in or depart from the crowdsensing platform from time to time, such that learning the sensing abilities of the unknown workers individually through the standard CMAB framework is not feasible. To address the above issue, we have proposed an on-line version of our CACI mechanism. Since correlation between the context and the sensing ability is learnt in our mechanism, we can estimate the sensing ability of any available worker according to its context. Likewise, the truthfulness and the individual rationality of our on-line CACI mechanism can be ensured. We also prove the upper bound of its regret achieves $\mathcal O(B)$, even the available workers across different time slots are distinct. We finally have conducted extensive experiments on both synthetic and real datasets to verify the efficacy of our off-line and on-line mechanisms.

  Our CACI mechanisms entail the context information of workers as input, whereas the context information may be private for the workers. Therefore, we are on the way of developing privacy preserving approaches for our context-aware incentive mechanisms.

\vspace{-2ex}
\bibliographystyle{plain}
\bibliography{incentmab}

%%%%%%%%%%%%%%%%%%%%%%%%%%%%%%%%%%%%%%%%%%%%%%%%%%%%%%%%%%%%%%%%%%%%%%%%%%%%%%%%%%%%%%%%%%%%%%%%

\clearpage

\begin{appendices}
  \section{Proof of Proposition~\ref{prop:bl-truthfulness}} \label{appsec:bl-truthfulness}
    We first show in \textbf{Lemma}~\ref{le:bdpay} that the payment to any selected worker is not less than its proposed bid.
    \begin{lemma} \label{le:bdpay}
      In the baseline mechanism, $\tilde{p}^*_i - b_i \geq 0, ~\forall i \in \widetilde{\mathcal N}^*$.
    \end{lemma}
    \begin{proof}
      According to \textbf{Algorithm}~\ref{alg:baseline}, for the top-$K$ workers in ordered $\mathcal N$, we have $\rho^*_1 \geq \cdots \geq \rho^*_K \geq \rho^*_{K+1}$ (see Line~\ref{ln:bl-sort} in \textbf{Algorithm}~\ref{alg:baseline}). Considering the definition of $\tilde{p}^*_i$ in (\ref{eq:bl-payment}), if $\tilde{p}^*_i = {\mu_i}/{\rho^*_{K+1}}$, we have ${\tilde{p}^*_i}/{b_i} = {\rho^*_{i}}/{\rho^*_{K+1}} \geq 1$ and thus $\tilde{p}^*_i - b_i \geq 0$; if $\tilde{p}^*_i = b_{max}$, we have $\tilde{p}^*_i - b_i = b_{max} - b_i \geq 0$. We complete the proof by combining both the above cases.
    \end{proof}

    Based on \textbf{Lemma}~\ref{le:bdpay}, we now present the proof of \textbf{Proposition}~\ref{prop:bl-truthfulness} as follows. Let $z_i \in \{0,1\}$ and $z'_i \in \{0,1\}$ be indicators specifying if worker $i \in \widetilde{\mathcal N}^*$ is selected by bidding $b_i$ and $c_i$, respectively. According to \textbf{Algorithm}~\ref{alg:baseline}, in each time slot, every worker in $\widetilde{\mathcal N}^*$ will be selected and paid once. Specifically, for each worker $i$, when $z_i = 1$ (resp. $z'_i=1$), it will be paid $\tilde{p}^*_i$ in each time slot by bidding $b_i$ (resp. $c_i$). We consider the following two cases:
    \begin{itemize}
      \item \textbf{Case I}: $b_i > c_i$ such that $z_i \leq z'_i$ \footnote{Raising a higher bid may increase the odds of being not recruited in \textbf{Algorithm}~\ref{alg:baseline}.}. If $z_i = z'_i$, worker $i$ is paid either $\tilde{p}^*_i$ (when $z_i = z'_i = 1$) or $0$ (when $z_i = z'_i = 0$) in each time slot, while the payment is independent of $b_i$. In another word, regardless of what it bids, it obtains the same utility when $z_i = z'_i$. If $z_i = 0$ and $z'_i = 1$, the utility of worker $i$ in each time slot is $\tilde{p}^*_i - c_i \geq 0$ (see \textbf{Lemma}~\ref{le:bdpay}) when bidding $c_i$, or $0$ when bidding $b_i > c_i$. Hence, the best choice for each worker is to bid truthfully.
      \item \textbf{Case II}: $b_i < c_i$ such that $z_i \geq z'_i$. If $z_i = z'_i$, the payment obtained by worker $i$ does not depend on its bid, according to what we have shown above. If $z_i > z'_i$ (i.e., $z_i = 1$ and $z'_i = 0$), we have $b_i \leq \mu_i / \rho^*_{K+1}$ and $c_i \geq \mu_i / \rho^*_{K+1}$. It also implies that, worker $i$ is paid $\tilde{p}^*_i$ when bidding $b_i < c_i$, or $0$ when bidding truthfully. If bidding $b_i < c_i$, the utility of worker in each time slot is $\tilde{p}^*_i - c_i = {\mu_i}/{\rho^*_{K+1}} - c_i \leq 0$. Hence, proposing a smaller bid cannot help worker $i$ to earn more utility.
    \end{itemize}
    Combining the above two cases, we conclude the dominant strategy for each worker is to report its true cost $b_i = c_i$.

  \section{Proof of Proposition~\ref{prop:bl-rationality}}  \label{appsec:bl-rationality}
    For each worker $i$, if $i \notin \widetilde{\mathcal N}^*$, we have $x^{[t]}_i = 0$ and $p^{[t]}_i = 0$ for $\forall t \in \mathcal T$, and the utility of the worker $i$ is thus zero. Hence, we focus on analyzing the payment to the workers in $\widetilde{\mathcal N}^*$. Specifically, according to \textbf{Lemma}~\ref{le:bdpay}, for $\forall i \in \widetilde{\mathcal N}^*$, we have $\tilde{p}^*_i - b_i \geq 0$. Since the truthfulness of our mechanism can be ensured such that $b_i = c_i$ as shown in \textbf{Proposition}~\ref{prop:bl-truthfulness}, we thus conclude that each worker in $\widetilde{\mathcal N}^*$ gets non-negative utility by truthfully participating in the auction.

  % \section{Proof of Lemma~\ref{le:intracube}}
  % % 
  %   Since we evenly partitioning the context space into a group of $d^M$ hypercubes with identical size $1/d^M$ (see Sec.~\ref{ssec:offmechdesign}), we have $\|s - s'\| \leq M^{\frac{1}{2}} d^{-1}$ for $\forall s, s' \in Q$. By considering the H$\ddot{\mathrm{o}}$lder Condition (see \textbf{Assumption}~\ref{as:holder}), we then have $\vert \mathbb E[r(s)] - \mathbb E[r(s')]\vert \leq L \|s - s'\|^\alpha =  L\left(M^{\frac{1}{2}} d^{-1}\right)^\alpha$.

  \section{Proof of Proposition~\ref{prop:subregret1}}  \label{appsec:subregret1}  
    We first continue (\ref{eq:regret1-1}) as follows
    \begin{align} \label{eq:regret1-2}
      &\mathsf{Regret}(\mathsf{F}^*_{\mathsf{off}}(\mathbf{m}, \mathbf{b}, B), \mathsf{F}^*_{\mathsf{off}}(\overline{\mathbf{m}}, \mathbf{b}, B))   \nonumber\\
      %
      %\leq& B \left( \frac{\sum_{i \in \widetilde{\mathcal N}^*} \mu_i}{\sum_{i \in \widetilde{\mathcal N}^*} \tilde{p}^*_i} - \frac{\sum_{i \in \widetilde{\mathcal N}^\dagger} \mu_i}{\sum_{i \in \widetilde{\mathcal N}^\dagger} \tilde{p}^\dagger_i} \right)  + \tau\mu_{max}  \nonumber\\
      %
      \leq& B \left( \frac{\sum_{i \in \widetilde{\mathcal N}^*} \mu_i}{\sum_{i \in \widetilde{\mathcal N}^*} \frac{\mu_i}{\rho_{(K+1)^*}}} - \frac{\sum_{i \in \widetilde{\mathcal N}^\dagger} (\mu_{Q(i)}-\delta)}{\sum_{i \in \widetilde{\mathcal N}^\dagger} \frac{\mu_{Q(i)}}{\rho_{(K+1)^\dagger}}} \right)  + K\mu_{max}  \nonumber\\
      %
      %\leq& B \left( \sum_{i \in \widetilde{\mathcal N}^*} \frac{\mu_{(\tau+1)^*}}{b_{(\tau+1)^*}} - \sum_{i \in \widetilde{\mathcal N}^\dagger} \frac{\mu_{Q(i)}}{b_{(\tau+1)^\dagger}} \right) +  \frac{\tau\delta}{\sum_{i \in \widetilde{\mathcal N}^\dagger}  \tilde{p}^\dagger_i}   + \tau\mu_{max}  \nonumber\\
      %
      \leq& B \left( \frac{\mu_{(K+1)^*}}{b_{(K+1)^*}} -  \frac{\mu_{Q((K+1)^\dagger)}}{b_{(K+1)^\dagger}} \right) +  \frac{B\delta}{b_{min}}   + K\mu_{max}
    \end{align}
    where the first inequality holds since $\mu_i \geq \mu_{Q(i)} - \delta$ for $\forall i \in \mathcal N$ (see \textbf{Lemma}~\ref{le:intracube}), while the second one is due to the fact that $\tilde p^\dagger_i \geq b_{min}$. We then analyze the bound on $\frac{\mu_{(K+1)^*}}{b_{(K+1)^*}} -  \frac{\mu_{Q((K+1)^\dagger)}}{b_{(K+1)^\dagger}}$ by taking into account the following four cases:
    \begin{itemize}
      \item When $\frac{\mu_{(K+1)^*}}{b_{K+1)^*}} > \frac{\mu_{(K+1)^\dagger}}{b_{(K+1)^\dagger}}$ and $\frac{\mu_{Q((K+1)^*)}}{b_{(K+1)^*}} \leq \frac{\mu_{Q((K+1)^\dagger)}}{b_{(K+1)^\dagger}}$, we have $\frac{\mu_{(K+1)^*}}{b_{(K+1)^*}} -  \frac{\mu_{Q((K+1)^\dagger)}}{b_{(K+1)^\dagger}} \leq \frac{\mu_{(K+1)^*}}{b_{(K+1)^*}} - \frac{\mu_{Q((K+1)^*)}}{b_{(K+1)^*}} \leq \frac{\delta}{b_{min}}$.
      \item When $\frac{\mu_{(K+1)^*}}{b_{(K+1)^*}} \leq \frac{\mu_{(K+1)^\dagger}}{b_{(K+1)^\dagger}}$ and $\frac{\mu_{Q((K+1)^*)}}{b_{(K+1)^*}} \leq \frac{\mu_{Q((K+1)^\dagger)}}{b_{(K+1)^\dagger}}$, we have $\frac{\mu_{(K+1)^*}}{b_{(K+1)^*}} - \frac{\mu_{Q((K+1)^\dagger)}}{b_{(K+1)^\dagger}} \leq \frac{\mu_{(K+1)^\dagger}}{b_{(K+1)^\dagger}} - \frac{\mu_{Q((K+1)^\dagger)}}{b_{(K+1)^\dagger}} \leq \frac{\delta}{b_{min}}$.
      \item When $\frac{\mu_{(K+1)^*}}{b_{(K+1)^*}} > \frac{\mu_{(K+1)^\dagger}}{b_{(K+1)^\dagger}}$ and $\frac{\mu_{Q((K+1)^*)}}{b_{(K+1)^*}} > \frac{\mu_{Q((K+1)^\dagger)}}{b_{(K+1)^\dagger}}$, it is implied that worker $(K+1)^* \notin \widetilde{\mathcal N}^*$ is selected in $\widetilde{\mathcal N}^\dagger$ and hence there exist at least one worker $j^* \in \widetilde{\mathcal N}^*$ but $ j^* \notin \widetilde{\mathcal N}^\dagger$. For such a worker, we have $\frac{\mu_{Q(j^*)}}{b_{j^*}} \leq \frac{\mu_{Q((K+1)^\dagger)}}{b_{(K+1)^\dagger}}$ and $\frac{\mu_{j^*}}{b_{j^*}} \geq \frac{\mu_{(K+1)^*}}{b_{(K+1)^*}}$. Therefore, $\frac{\mu_{(K+1)^*}}{b_{(K+1)^*}} -  \frac{\mu_{Q((K+1)^\dagger)}}{b_{(K+1)^\dagger}} \leq \frac{\mu_{j^*}}{b_{j^*}} - \frac{\mu_{Q(j^*)}}{b_{j^*}} \leq \frac{\delta}{b_{min}}$.
      \item When $\frac{\mu_{(K+1)^*}}{b_{(K+1)^*}} \leq \frac{\mu_{(K+1)^\dagger}}{b_{(K+1)^\dagger}}$ and $\frac{\mu_{Q((K+1)^*)}}{b_{(K+1)^*}} > \frac{\mu_{Q((K+1)^\dagger)}}{b_{(K+1)^\dagger}}$, we have $\frac{\mu_{(K+1)^*}}{b_{(K+1)^*}} -  \frac{\mu_{Q((K+1)^\dagger)}}{b_{(K+1)^\dagger}} \leq \frac{\mu_{(K+1)^\dagger}}{b_{(K+1)^\dagger}} - \frac{\mu_{Q((K+1)^\dagger)}}{b_{(K+1)^\dagger}} \leq \frac{\delta}{b_{min}}$.
    \end{itemize}
    Combining all the above cases, we have $\frac{\mu_{(K+1)^*}}{b_{(K+1)^*}} -  \frac{\mu_{Q((K+1)^\dagger)}}{b_{(K+1)^\dagger}} \leq \frac{\delta}{b_{min}}$, by substituting which into (\ref{eq:regret1-2}), we complete the proof.

  \section{Proof of Proposition~\ref{prop:subregret2}}  \label{appsec:subregret2}
    Before diving into the proof of the upper bound of $\mathsf{Regret} \left( \mathsf{F}^*_{\mathsf{off}}(\overline{\mathbf{m}}, \mathbf{b}, B), \mathsf{F}_{\mathsf{off}}(\mathbf{s}, \mathbf{b}, B) \right)$ in \textbf{Proposition}~\ref{prop:subregret2}, we first show in \textbf{Lemma}~\ref{le:bdui} that, for each worker $i$, the upper confidence bound on its sensing ability, i.e., $u_i$, is larger than $\mu_{Q(i)}$, and their difference $u_i - \mu_{Q(i)}$ is bounded with high probability.
    \begin{lemma} \label{le:bdui}
      Through the exploration phase of our CACI mechanism, we have 
      \begin{equation*} \label{eq:bdui}
        \mathbb P \left[ 0 < u_i - \mu_{Q(i)} < 2\sqrt{\frac{d^M b_{max} \ln B}{B^\#}} , \forall i\in\mathcal N \right] \geq 1 - \frac{2d^M}{B^2}
      \end{equation*}
    \end{lemma}
    \begin{proof}
      Let $T^\#_Q$ denote the number of time slots where hypercube $Q$ is explored. According to \textbf{Algorithm}~\ref{alg:offline-caci}, we have $T^\#_Q \geq \left\lfloor \frac{B^\#}{d^M b_{max}} \right\rfloor$. By applying Hoeffding's inequality~\cite{DubhashiP-book09}, we have 
      \begin{align*}
        &\mathbb P \left[ \left| \bar{r}^{\left[ T^\#_Q \right]}(Q) - \mu_{Q} \right| \geq  \sqrt{\frac{d^M b_{max} \ln B}{B^\#}} \right]  \nonumber\\
        %
        %\leq&\mathbb P \left[ \left| \frac{1}{T^\#_{Q}} \sum^{T^\#_{Q}}_{\ell=1} r_\ell(Q) - \mu_{Q} \right| \geq  \sqrt{\frac{\ln B}{T^\#_{Q}}} \right] \leq \frac{2}{B^2}
        %
        \leq& \mathbb P \left[ \left| \bar{r}^{ \left[ T^\#_Q \right]}(Q) - \mu_{Q} \right| \geq  \sqrt{\frac{\ln B}{T^\#_{Q}}} \right] \leq \frac{2}{B^2}
      \end{align*}
      %
      %where $r_\ell(Q)$ denotes the reward obtained by the $\ell$-th selection of $Q$. 
      %
      We then apply union bound such that
      \begin{align*}
        \mathbb P \left[ \bigcap_{Q \in \mathcal Q} \left\{ \left| \bar r^{[T^\#_Q]}(Q) - \mu_{Q} \right| \leq  \sqrt{\frac{d^M b_{max} \ln B}{B^\#}} \right\} \right] \geq& 1 - \frac{2d^M}{B^2}
      \end{align*}
      We finally complete the proof by substituting the definition of $u_i$ (\ref{eq:ucb}) into the above inequality.
    \end{proof}

    Based on \textbf{Lemma}~\ref{le:intracube} and \textbf{Lemma}~\ref{le:bdui}, we then prove the upper bound on the expected difference between $u_i$ and $\mu_i$ in \textbf{Lemma}~\ref{le:ucb}. In another word, $u_i$ can be used as an accurate estimate on $\mu_i$.
    \begin{lemma} \label{le:ucb}
      In our mechanism, for $\forall i\in \mathcal N$, the expected difference between $u_i$ and $\mu_i$ is bounded by
      \begin{equation}
        \mathbb E [ u_i - \mu_i ] \leq 2 \sqrt{\frac{d^M b_{max} \ln B}{B^\#}} + \delta
      \end{equation}
      with probability at least $ 1 - \frac{2 d^M}{B^2}$.
    \end{lemma}
    \begin{proof}
      For each hypercube $Q \in \mathcal Q$, we define $\xi_Q = \min_{s \in Q} \mathbb E[r(s)]$. According to \textbf{Lemma}~\ref{le:intracube}, we have $\mu_Q - \xi_Q \leq \delta$. Furthermore, since $\mu_i \geq \xi_{Q(i)}$, it follows $\mathbb E[u_i - \mu_i] \leq \mathbb E \left[ u_i - \xi_{Q(i)} \right] \leq  \mathbb E \left[ u_i - \mu_{Q(i)} \right] + \delta$. We finally complete the proof by considering \textbf{Lemma}~\ref{le:bdui}.
    \end{proof}

    Based on \textbf{Lemma}~\ref{le:bdui}, we also prove in \textbf{Lemma}~\ref{le:workercube} that, $u_i / b_i \geq \mu_{Q(i^\dagger)} /  b_{i^\dagger}$ holds with high probability, which will be very helpful in the proof of \textbf{Proposition}~\ref{prop:subregret2}.
    \begin{lemma} \label{le:workercube}
      Recall that the workers in $\widetilde{\mathcal N}$ and $\widetilde{\mathcal N}^\dagger$ are sorted in decreasing order according to $\rho_i = {u_i}/{b_i}$ and $\rho^\dagger_i = {\mu_{Q(i)}}/{b_i}$, respectively. For the $i$-th worker in $\widetilde{\mathcal N}$ and the one in $\widetilde{\mathcal N}^\dagger$ (denoted by $i \in \widetilde{\mathcal N}$ and $i^\dagger \in \widetilde{\mathcal N}^\dagger$, respectively), we have 
      \begin{equation} \label{eq:workercube}
        \frac{u_i}{b_{i}} \geq \frac{\mu_{Q(i^\dagger)}}{b_{i^\dagger}}
      \end{equation} 
      hold with probability at least $1-2d^M/B^2$.
    \end{lemma} 
    \begin{proof}
      We take into account the following two different cases, i.e., $i = i^\dagger$ and $i \neq i^\dagger$. If $i = i^\dagger$, the lemma can be proved by directly referring to  \textbf{Lemma}~\ref{le:bdui}; otherwise, we conduct the proof by contradiction as follows. Specifically, we assume $\frac{u_i}{b_{i}} < \frac{\mu_{Q(i^\dagger)}}{b_{i^\dagger}}$ for any worker $i \in \widetilde{\mathcal N}$ and $i^\dagger \in \widetilde{\mathcal N}^\dagger$. Then for $\forall j^\dagger = 1^\dagger,2^\dagger,\cdots,i^\dagger$, we have $\frac{u_{j^\dagger}}{b_{j^\dagger}} \geq \frac{\mu_{Q(j^\dagger)}}{b_{j^\dagger}} \geq \frac{\mu_{Q(i^\dagger)}}{b_{i^\dagger}} > \frac{u_i}{b_i}$, where the first inequality holds due to \textbf{Lemma}~\ref{le:bdui}, and we have the second one since $\frac{\mu_{Q(1^\dagger)}}{b_{1^\dagger}} \geq \frac{\mu_{Q(2^\dagger)}}{b_{2^\dagger}} \geq \cdots \geq \frac{\mu_{Q(i^\dagger)}}{b_{i^\dagger}}$ in $\widetilde{\mathcal N}^\dagger$. The above inequality implies that there would be $i$ workers whose QBRs are higher than the one of the $i$-th worker in ordered $\widetilde{\mathcal N}$, which is a contradiction to the fact that worker $i$ is the one with the $i$-th highest QBR value. 
    \end{proof}

    Now we are ready to prove \textbf{Proposition}~\ref{prop:subregret2}. First, we have
    \begin{align*}
      &\mathbb E \left[ \frac{\mu_{Q((K+1)^\dagger)}}{b_{(K+1)^\dagger}} - \frac{\sum_{i\in \widetilde{\mathcal N}} \mu_i}{\sum_{i\in \widetilde{\mathcal N}} \tilde{p}_i}  \right]  \nonumber\\
      =& \mathbb E \left[ \frac{1}{\sum_{i\in \widetilde{\mathcal N}} \tilde{p}_i} \left( \frac{\mu_{Q((K+1)^*)}}{b_{(K+1)^*}}  \sum_{i\in \widetilde{\mathcal N}} \tilde{p}_i - {\sum_{i\in \widetilde{\mathcal N}} \mu_i} \right) \right]  \nonumber\\
      \leq& \frac{1}{K b_{min}} \mathbb E \left[ \frac{\mu_{Q((K+1)^*)} }{b_{(K+1)^*}} \sum_{i\in \widetilde{\mathcal N}} \tilde{p}_i - {\sum_{i\in \widetilde{\mathcal N}} \mu_i} \right]   \nonumber\\
      %
      %=& \frac{1}{\tau b_{min}} \left( \sum_{i\in \widetilde{\mathcal N}} \mathbb E\left[ \frac{\mu_{Q((\tau+1)^*)}}{b_{(\tau+1)^*}} \cdot \frac{u_i b_{\tau+1}}{u_{\tau+1}} \right] - \sum_{i\in \widetilde{\mathcal N}} \mu_i  \right)  \nonumber\\
      %
      \leq& \frac{1}{K b_{min}}  \sum_{i\in \widetilde{\mathcal N}} \mathbb E[u_i - \mu_i]  
      \leq \frac{2}{b_{min}}  \sqrt{\frac{d^M b_{max} \ln B}{B^\#}} + \frac{\delta}{b_{min}}
    \end{align*}
    holds with probability at least $1 - {2d^M}/{B^2}$, where the first inequality holds because $\tilde{p}_i \geq b_{min}$ for $\forall i\in \mathcal N$, the second one is ensured due to $\frac{\mu_{Q((K+1)^\dagger)}}{b_{(K+1)^\dagger}} < \frac{u_{(K+1)}}{b_{K+1}}$ as shown in \textbf{Lemma}~\ref{le:workercube}, and we have the last one hold according to \textbf{Lemma}~\ref{le:ucb}. We then substitute the above inequality into (\ref{eq:regret2-1}), such that
    \begin{align} 
      &\mathsf{Regret} \left( \mathsf{F}^*_{\mathsf{off}}(\overline{\mathbf{m}}, \mathbf{b}, B), \mathsf{F}_{\mathsf{off}}(\mathbf{s}, \mathbf{b}, B) \right)   \nonumber\\
      \leq&  \frac{B^\# \mu_{max}}{b_{min}} + \frac{B\delta}{b_{min}} + K \mu_{max}  \nonumber\\
      & + (B-B^\#) \left[ \frac{2}{b_{min}}  \sqrt{\frac{d^M b_{max} \ln B}{B^\#}} + \frac{\delta}{b_{min}}  \right]   \nonumber\\
      \leq&  \frac{2(B-B^\#)}{b_{min}} \sqrt{\frac{d^M b_{max} \ln B}{B^\#}}  + \frac{2B\delta}{b_{min}} + \frac{B^\# \mu_{max}}{b_{min}} +  K \mu_{max}  \nonumber\\
      \leq&  \frac{2B}{b_{min}} \sqrt{\frac{d^M b_{max} \ln B}{B^\#}}  + \frac{2B\delta}{b_{min}} + \frac{B^\# \mu_{max}}{b_{min}} +  K \mu_{max}
    \end{align}
    which completes the proof.

  \section{Proof of Theorem~\ref{thm:offcaci-truth}}  \label{appsec:offcaci-truth}
    We first prove the truthfulness of our off-line CACI mechanism. As demonstrated by \textbf{Algorithm}~\ref{alg:offline-caci}, in the exploration phase, we select among the given workers in a uniform manner, and pay each of the selected workers $b_{max}$. For each selected worker $i$, its utility is $b_{max} - c_i$, while the utility for each of the others is $0$. Therefore, the truthfulness of the exploration phase can be ensured. In the following, we concentrate on analyzing the one of the exploitation phase
      
    The corresponding proof is conducted by following the basic idea of the one of \textbf{Proposition}~\ref{prop:bl-truthfulness}. For any specific worker $i$, fix the bids of the others, and let $z_i \in \{0,1\}$ and $z'_i \in \{0,1\}$ be indicators specifying if worker $i$ is selected (i.e., $i \in \widetilde{\mathcal{N}}$) by bidding $b_i$ and $c_i$, respectively. When $z_i = z'_i$, the utility of worker $i$ actually does not depend on $b_i$. Specifically, i)when $z_i = z'_i = 1$, the utility of worker $i$ is $\min \left\{ \frac{u_i}{\rho_{K+1}}, b_{max} \right\} - c_i$, regardless of whether $b_i = c_i$ or not; ii) when $z_i = z'_i = 0$, the utility worker $i$ is $0$, whatever it bids. In another word, there is no need for worker $i$ to bid untruthfully, if doing so cannot change the selection decision. Furthermore, when $z'_i \neq z_i$, we consider the following two cases: i) worker $i$ submits $b_i > c_i$ such that $z_i=0$ and $z'_i=1$, and ii) worker $i$ submits $b_i < c_i$ such that $z_i=1$ and $z'_i=0$. In the first case i), the utility of worker $i$ is $0$ when it bids $b_i$, while the utility is $\tilde{p}^*_i - c_i >0$ when worker $i$ bids $c_i$. Hence, worker $i$ can obtain higher utility by bidding truthfully. In the second case ii), we have $\frac{u_i}{b_i} \geq \frac{u_{K+1}}{b_{K+1}}$ and $\frac{u_i}{c_i} \leq \frac{u_{K+1}}{b_{K+1}}$ and hence $b_i \leq \frac{u_i}{\rho_{K+1}}$ and $c_i \geq \frac{u_i}{\rho_{K+1}}$. Therefore, in this case with $b_i < c_i$, the utility of worker $i$ is $\tilde{p} - c_i \leq \frac{u_i}{\rho_{K+1}} \leq 0$, which implies $i$ cannot get higher utility by bidding $b_i < c_i$. 

%\section*{Proof of Theorem~3}  
%
    We then prove the individual rationality of our off-line CACI mechanism. In each time slot $t$, each worker $i \notin \widetilde{\mathcal N}^{[t]}$ have $x^{[t]}_i = 0$ and $p^{[t]}_i = 0$, and thus get zero utility, according to \textbf{Definition}~\ref{def:ration}. We then show that each $i \in \widetilde{\mathcal N}^{[t]}$ have non-negative utility. When $\tilde{p}^{[t]}_i = u_i / \rho_{K+1}$, we have $\tilde{p}^{[t]}_i / b_i = \rho_i / \rho_{K+1} \geq 1$ and thus $\tilde{p}^{[t]}_i \geq b_i$. Since the truthfulness of our mechanism holds such that $b_i = c_i$ as shown in \textbf{Theorem}~\ref{thm:offcaci-truth}, the utility of worker $i \in \widetilde{\mathcal{N}}^{[t]}_i$ with $x^{[t]}_i = 1$ is non-negative. When $\tilde{p}^{[t]}_i = b_{max}$, we have $\tilde{p}^{[t]}_i - c_{i} = b_{max} - b_{i} \geq 0$, the utility of worker $i \in \widetilde{\mathcal{N}}^{[t]}_i$ is also non-negative.

  %\vspace{-4ex}
  \section{Proof of Proposition~\ref{prop:on-subregret1}}  \label{secapp:on-subregret1}
    Assume $C^{[t]}(Q)$ is a counter for $\forall Q \in \mathcal Q$ in the exploration-and-exploitation phase. Let $\widetilde{\mathcal{Q}}^{[t]} \subseteq \mathcal{Q}$ be the set of the hypercubes selected by our on-line CACI mechanism in time slot $t$, i.e., $\widetilde{\mathcal{Q}}^{[t]} = \{Q(i)\}_{i \in \widetilde{\mathcal N}^{[t]}}$. Similarly, let $\widetilde{\mathcal{Q}}^{\dagger[t]} = \{Q(i)\}_{i \in \widetilde{\mathcal N}^{\dagger[t]}}$. In each time slot $t$ of the exploration-and-exploitation phase, the counter $C^{[t]}(Q)$ is updated according to the following rule: if there exists some worker $i \in \widetilde{\mathcal N}^{[t]}$ such that $Q(i) \in \widetilde{\mathcal Q}^{[t]}$ but $Q(i) \notin \widetilde{\mathcal Q}^{\dagger [t]}$, we have $C^{[t]}(Q) =  C^{[t-1]}(Q) + 1$ for $Q = \arg\min_{Q'\in \mathcal Q} C^{[t]}(Q')$ and $C^{[t]}(Q) =  C^{[t-1]}(Q)$ for others; if $\widetilde{\mathcal Q}^{[t]} = \widetilde{\mathcal Q}^{\dagger [t]}$, we have $C^{[t]}(Q)=  C^{[t-1]}(Q)$ for $\forall Q \in \mathcal Q$. According to the above update rule, we have $\lambda^{[t]}(Q) \geq C^{[t]}(Q)$. We reveal the bound on $\mathbb{E} \left[ C^{[\widetilde{T}]} (Q) \right]$ for $\forall Q \in \mathcal Q$ in the following \textbf{Lemma}~\ref{le:bdcounter}, based on which, We also show the upper bound of $\mathbb{E} \left[ \widetilde{T}^* - \widetilde{T} \right]$ in \textbf{Lemma}~\ref{le:expdifftime}.
    \begin{lemma} \label{le:bdcounter}
      For $\forall Q \in \mathcal Q$, the expected value of $C^{[\widetilde{T}]} (Q)$ is bounded by
      \begin{align} \label{eq:bdcounter}
        \mathbb{E} \left[ C^{\left[ \widetilde{T} \right]}_Q \right] \leq \frac{4K^2(K+1)}{b^2_{min}\Delta^2_{min}} \ln{\frac{B}{K b_{min}}} + \frac{K \pi^2}{3} + 1
      \end{align}
    \end{lemma}
    \begin{proof}
    Assume ${I}^{[t]}: \mathcal Q \rightarrow \{0,1\}$ denotes an indicator function such that ${I}^{[t]}(Q)=1$ if $C^{[t]}(Q)$ is increased in time slot $t$; otherwise, ${I}^{[t]}(Q)=0$. Then, for $\forall Q \in \mathcal{Q}$ we have
    \begin{align*}
      & C^{\left[ \widetilde{T} \right]}(Q) \leq \ell + \sum^{\widetilde{T}}_{t=2} \mathbb{I}( {I}^{[t]}(Q)=1,  C^{[t-1]}(Q) \geq \ell )  \\
      \leq & \ell + \sum^{\widetilde{T}}_{t=2}  \mathbb{I} \left( \sum_{i \in \widetilde{\mathcal N}^{[t]}} \frac{u^{[t]}_i}{b_i} \geq \sum_{i \in \widetilde{\mathcal N}^{\dagger [t]}} \frac{\mu_{Q(i)}}{b_i},  C^{[t-1]} (Q) \geq \ell \right)  \\
      \leq & \ell + \sum^{\widetilde T}_{t=2} \mathbb{I} \Bigg( \max_{\widetilde{\mathcal N}^{[t]} \subseteq \mathcal{N}^{[t]} }  \sum_{i \in \widetilde{\mathcal N}^{[t]}} \frac{u^{[t]}_i}{b_i}   \geq \min_{\widetilde{\mathcal N}^{\dagger [t]} \subseteq \mathcal{N}^{[t]}} \sum_{i \in \widetilde{\mathcal N}^{\dagger [t]}} \frac{\mu_{Q(i)}}{b_i} \Bigg)
    \end{align*}
    Assuming $\widetilde{\mathcal N}^{[t]} = \left\{ \tilde{i}_1, \tilde{i}_2, \cdots, \tilde{i}_K \right\}$ with $\ell \leq \lambda^{[t-1]} ( Q(\tilde{i}_j) ) \leq t-1$ for $\forall j=1,2,\cdots,K$ and $\widetilde{\mathcal N}^{\dagger} = \left\{ \tilde{i}^{\dagger}_1, \tilde{i}^{\dagger}_2, \cdots, \tilde{i}^{\dagger}_K \right\}$ with $1 \leq \lambda^{[t-1]} ( Q(\tilde{i}^{\dagger}_j) ) \leq t-1$ for $\forall j=1,2,\cdots,K$, we continue the above inequality as follows
    \begin{align*}
      & C^{[\widetilde{T}]}(Q) \\
      & \leq \ell + \sum^{\widetilde{T}}_{t=2} \sum^{t}_{\lambda^{[t-1]}\left(Q(\tilde{i}_1)\right)=\ell} \cdots \sum^{t}_{\lambda^{[t-1]}\left(Q(\tilde{i}_K)\right)=\ell} \\
      & \sum^{t}_{\lambda^{[t-1]}\left(Q(\tilde{i}^{\dagger}_1)\right)=1} \cdots \hspace{-2ex} \sum^{t}_{\lambda^{[t-1]}\left(Q(\tilde{i}^{\dagger}_K)\right)=1} \mathbb{I} \left( \sum^K_{j=1} \frac{u^{[t]}_{\tilde{i}_j}}{b_{\tilde{i}_j}} \geq \sum^K_{j=1}\frac{\mu_{Q(\tilde{i}^{\dagger}_j)}}{b_{\tilde{i}^{\dagger}_j}} \right)
    \end{align*}

    We then demonstrate the bound of $\sum^K_{j=1} \frac{u^{[t]}_{\tilde{i}_j}}{b_{\tilde{i}_j}} \geq \sum^K_{j=1} \frac{u^{[t]}_{\tilde{i}^\dagger_j}}{b_{\tilde{i}^\dagger_j}}$. Specifically, if the following event holds
    \begin{align}
      &\sum^K_{j=1} \frac{1}{b_j} \left( \bar{r}^{[t-1]}(Q(\tilde{i}_j)) + \sqrt{\frac{(K+1)\ln t}{\lambda^{[t-1]}(Q(\tilde{i}_j))}}\right)  \nonumber\\
      \geq& \sum^K_{j=1} \frac{1}{b_j} \left( \bar{r}^{[t-1]}(Q(\tilde{i}^\dagger_j)) + \sqrt{\frac{(K+1)\ln t}{\lambda^{[t-1]}(Q(\tilde{i}^\dagger_j))}}\right)
    \end{align}
    at least one of the following three events is true
    \begin{align*}
      \mathsf{E}_1:&\sum^K_{j=1} \frac{\bar{r}^{[t-1]}(Q(\tilde{i}_j))}{b_{\tilde{i}_j}} \geq \sum^K_{j=1} \frac{1}{b_{\tilde{i}_j}}\left( \mu_{Q(\tilde{i}_j)} + \sqrt{\frac{(K+1)\ln t}{\lambda^{[t-1]}(Q(\tilde{i}_j))}}\right)  \\
      \mathsf{E}_2:&\sum^K_{j=1} \frac{\bar{r}^{[t-1]}(Q(\tilde{i}^\dagger_j))}{b_{\tilde{i}^\dagger_j}} \leq \sum^K_{j=1} \frac{1}{b_{\tilde{i}^\dagger_j}}\left( \mu_{Q(\tilde{i}^\dagger_j)} - \sqrt{\frac{(K+1)\ln t}{\lambda^{[t-1]} ( Q (\tilde{i}^\dagger_j ) )}}\right)  \\
      \mathsf{E}_3:&\sum^K_{j=1} \frac{\mu_{Q_{\tilde{i}^\dagger_j}}}{b_{\tilde{i}^\dagger_j}} < \sum^K_{j=1} \frac{1}{b_{\tilde{i}_j}}\left( \mu_{Q(\tilde{i}_j)} + 2\sqrt{\frac{(K+1)\ln t}{\lambda^{[t-1]}(Q(\tilde{i}_j))}}\right)
    \end{align*}
    According to the Chernoff-Hoeffding bound~\cite{DubhashiP-book09}, we have
    \begin{align*}
      &\mathbb{P}(\mathsf{E}_1) \\%
      \leq& \sum^K_{j=1} \mathbb{P} \left( {\bar{r}^{[t-1]}(Q(\tilde{i}_j))} \geq  \mu_{Q(\tilde{i}_j)} + \sqrt{\frac{(K+1)\ln t}{\lambda^{[t-1]}(Q(\tilde{i}_j))}}  \right) \\
      \leq& \exp\left( -2(K+1) \ln t\right) = t^{-2(K+1)}
    \end{align*}
    Similarly, we also have
    \begin{align*}
      \mathbb{P}(\mathsf{E}_2) \leq \exp\left( -2(K+1) \ln t\right) = t^{-2(K+1)}
    \end{align*}
    When $\ell \geq \frac{4K^2 (K+1) \ln \widetilde{T}}{b_{min} \Delta_{min}}$, we have
    \begin{align}
      &\sum^K_{j=1} \frac{\mu_{Q_{\tilde{i}^\dagger_j}}}{b_{\tilde{i}^\dagger_j}} - \sum^K_{j=1} \frac{1}{b_{\tilde{i}_j}}\left( \mu_{Q_{\tilde{i}_j}} + 2\sqrt{\frac{(K+1)\ln t}{\gamma^{[t-1]}(Q(\tilde{i}_j))}}\right)  \nonumber\\
      =& \sum^K_{j=1} \left( \frac{\mu_{Q_{\tilde{i}^\dagger_j}}}{b_{\tilde{i}^\dagger_j}} - \frac{\mu_{Q_{\tilde{i}_j}}}{b_{\tilde{i}_j}} \right) - \sum^K_{j=1} \frac{2}{b_{\tilde{i}_j}} \sqrt{\frac{(K+1)\ln t}{\gamma^{[t-1]}(Q(\tilde{i}_j))}}  \nonumber\\
      \geq& \Delta_{min} - \frac{2}{b_{min}} \sum^K_{j=1} \sqrt{\frac{(K+1)\ln t}{\ell}} \geq 0
    \end{align}
    which implies $\mathbb P(\mathsf{E}_3) = 0$. We finally combine the above probabilities of the three events and have
    \begin{align}
      & \mathbb{E} \left[ C^{[t]}_Q \right]  \nonumber\\
      \leq& \frac{4K^2(K+1)}{b^2_{min}\Delta^2_{min}}\ln{\widetilde{T}} + \sum^{\infty}_{t=1}(t-\ell+1)^K t^K 2 K t^{-2(K+1)}  \nonumber\\
      \leq&  \frac{4K^2(K+1)}{b^2_{min}\Delta^2_{min}}\ln{\widetilde{T}} + 2K\sum^{\infty}_{t=1} t^{-2} + 1  \nonumber\\
      \leq& \frac{4K^2(K+1)}{b^2_{min}\Delta^2_{min}}\ln{\widetilde{T}} + \frac{K \pi^2}{3}
    \end{align}
    We finally complete the proof by considering $\widetilde{T} \leq \frac{B}{K b_{min}}$.
  \end{proof}

  \begin{lemma}  \label{le:expdifftime}
    Recall that $\widetilde{T}$ and $\widetilde{T}^*$ denotes the number of time slots our mechanism $\mathsf{F_{on}} \left( (\mathbf{s}^{[t]}, \mathbf{b}^{[t]})_{t\in \mathcal T}, B \right)$ proceeds and the one of the baseline mechanism $\mathsf{F_{on}} \left( (\mathbf{m}^{[t]}, \mathbf{b}^{[t]})_{t \in \mathcal T}, B \right)$, respectively. We have
    \begin{align} \label{eq:expdifftime}
      &\mathbb{E} \left[ \widetilde{T}^* - \widetilde{T} \right] \leq \frac{b_{max} (\mu_{max}+\delta)B}{b^2_{min} \mu_{min} K}  \nonumber\\
      & + \frac{4b_{max}K (K+1) d^M}{b^3_{min} \Delta^2_{min}} \ln \frac{B}{K b_{min}} + \frac{b_{max} d^M}{K b_{min}} \left( \frac{K \pi^2}{3} + 2 \right) 
    \end{align}
  \end{lemma}
  \begin{proof}
    We let $\widetilde{T}^\dagger$ denote the number of time slots $\mathsf{F_{on}} \left( (\overline{\mathbf{m}}^{[t]}, \mathbf{b}^{[t]})_{t \in \mathcal{T}}, B \right)$ proceeds, and decompose $\widetilde{T}^* - \widetilde{T}$ as
    \begin{equation}
      \widetilde{T}^* - \widetilde{T} = \widetilde{T}^* - \widetilde{T}^\dagger + \widetilde{T}^\dagger - \widetilde{T}
    \end{equation}
    According to our mechanism, we have
    \begin{align*}
      \widetilde{T}^\dagger - \widetilde{T} \leq \frac{b_{max}}{K b_{min}} \left( \sum_{Q \in \mathcal Q} C^{[\widetilde{T}]}(Q) + d^M \right)  \nonumber\\
    \end{align*}
    and hence
    \begin{align} \label{eq:bd-T-dagger}
      &\mathbb{E} \left[ \widetilde{T}^\dagger - \widetilde{T} \right]  \nonumber\\
      \leq& \frac{b_{max}}{K b_{min}} \left( \sum_{Q \in \mathcal Q} \mathbb{E} \left[ C^{[\widetilde{T}]} (Q) \right] + d^M \right)  \nonumber\\
      \leq& \frac{4 b_{max} K (K+1) d^M}{b^3_{min} \Delta^2_{min}} \ln \frac{B}{K b_{min}} + \frac{b_{max}}{K b_{min}} \left( \frac{K \pi^2}{3} + 2 \right) d^M 
    \end{align}
    according to \textbf{Lemma}~\ref{le:bdcounter}. Since
    \begin{equation*}
      \sum_{i \in \mathcal{N}^{*[t]}} \frac{\mu_{i}}{b_{min}} \geq \sum_{i \in \mathcal{N}^{*[t]}} \frac{\mu_{i}}{b_i} \geq \sum_{i \in \mathcal{N}^{\dagger[t]}} \frac{\mu_{i}}{b_i}
    \end{equation*}
    we have
    \begin{equation}  \label{eq:star-dagger}
      \sum_{i \in \mathcal{N}^{*[t]}} {\mu_{i}} \geq b_{min} \sum_{i \in \mathcal{N}^{\dagger[t]}} \frac{\mu_{i}}{b_i} \geq \frac{b_{min}}{b_{max}} \sum_{i \in \mathcal{N}^{\dagger[t]}} \mu_i
    \end{equation}

    Assume $p^{*[t]}_i$ be the payment to worker $i \in \widetilde{\mathcal N}^{*[t]}$ in time slot $t$ by $\mathsf{F_{on}}( (\mathbf{m}^{[t]}, \mathbf{b}^{[t]})_{t \in \mathcal T}, B)$ and $p^{\dagger[t]}_i$ the one by $\mathsf{F_{on}}( ( \overline{\mathbf{m}}^{[t]}, \mathbf{b}^{[t]} )_{t \in \mathcal{T}}, B)$. We also suppose $b^{*[t]}_{K+1}$ and $\mu^{*[t]}_{K+1}$ (resp. $b^{\dagger[t]}_{K+1}$ and $\mu^{\dagger[t]}_{K+1}$) denote the bid and the sensing ability of the $(K+1)$-th worker in decreasingly ordered ${\mathcal{N}}^{[t]}$ in $\mathsf{F_{on}}( (\mathbf{m}^{[t]}, \mathbf{b}^{[t]})_{t \in \mathcal{T}}, B)$ (resp. $\mathsf{F_{on}}( (\overline{\mathbf{m}}^{[t]}, \mathbf{b}^{[t]})_{t \in \mathcal T}, B )$), respectively. We then have the bound of $\sum_{i \in \mathcal{N}^{\dagger[t]}} p^{\dagger[t]}_i - \sum_{i \in \mathcal{N}^{*[t]}} p^{*[t]}_i$ as follows
    \begin{align} \label{eq:pay-dagger-star}
      &\sum_{i \in \mathcal{N}^{\dagger[t]}} p^{\dagger[t]}_i - \sum_{i \in \mathcal{N}^{*[t]}} p^{*[t]}_i  \nonumber\\
      =& \frac{b^{\dagger[t]}_{K+1}}{\mu^{\dagger[t]}_{K+1}} \sum_{i \in \widetilde{\mathcal N}^{\dagger[t]}} \mu_{Q(i)} - \frac{b^{*[t]}_{K+1}}{\mu^{*[t]}_{K+1}} \sum_{i \in \widetilde{\mathcal N}^{*[t]}} \mu_{i}  \nonumber\\
      %
      %\leq& \frac{b^{\dagger[t]}_{K+1}}{\mu^{\dagger[t]}_{K+1}} \sum_{i \in \widetilde{N}^{\dagger[t]}} \mu_{Q(i)} - \frac{c_{min} b^{*[t]}_{K+1}}{\mu^{*[t]}_{K+1}} \sum_{i \in \widetilde{N}^{\dagger[t]}} \frac{\mu_{i}}{b_i}  \nonumber\\
      %
      \leq& \frac{b^{\dagger[t]}_{K+1}}{\mu^{\dagger[t]}_{K+1}} \sum_{i \in \widetilde{\mathcal N}^{\dagger[t]}} \mu_{Q(i)} - \frac{b_{min} b^{*[t]}_{K+1}}{b_{max} \mu^{*[t]}_{K+1}} \sum_{i \in \widetilde{\mathcal N}^{\dagger[t]}} {\mu_{i}}  \nonumber\\
      \leq& \frac{b^{\dagger[t]}_{K+1}}{\mu^{\dagger[t]}_{K+1}} \sum_{i \in \widetilde{\mathcal N}^{\dagger[t]}} (\mu_i + \delta) - \frac{b_{min} b^{*[t]}_{K+1}}{b_{max} \mu^{*[t]}_{K+1}} \sum_{i \in \widetilde{\mathcal N}^{\dagger[t]}} {\mu_{i}}  \nonumber\\
      %
      %=& \left( \frac{b^{\dagger[t]}_{K+1}}{\mu^{\dagger[t]}_{K+1}} - \frac{c_{min} b^{*[t]}_{K+1}}{c_{max} \mu^{*[t]}_{K+1}} \right) \sum_{i \in \widetilde{\mathcal N}^{\dagger[t]}} \mu_i  +  \frac{b^{\dagger[t]}_{K+1}}{\mu^{\dagger[t]}_{K+1}} K L (M^{\frac{1}{2}} d^{-1})^\alpha  \nonumber\\
      %
      %\leq& \left( \frac{b^{\dagger[t]}_{K+1}}{\mu^{\dagger[t]}_{K+1}} - \frac{b_{min} b^{*[t]}_{K+1}}{b_{max} \mu^{*[t]}_{K+1}} \right) K \mu_{max}  +  \frac{b^{\dagger[t]}_{K+1}}{\mu^{\dagger[t]}_{K+1}} K L (M^{\frac{1}{2}} d^{-1})^\alpha  \nonumber\\
      %
      \leq& \left( \frac{b_{max} \mu_{max}}{\mu_{min}} - \frac{b^2_{min}}{b_{max}} \right) K   +  \frac{b_{max} \delta}{\mu_{min}} K 
    \end{align}  
    where we have the first inequality by substituting (\ref{eq:star-dagger}), the second one according to \textbf{Lemma}~\ref{le:intracube}, and the last one as $\mu_{min} \leq \mu_i \leq \mu_{max}$ and $b_{min} \leq b_i \leq b_{max}$  for $\forall i \in \widetilde{\mathcal N}^{[t]}$. Therefore, $\mathbb{E}[\widetilde{T}^* - \widetilde{T}^\dagger]$ can be bounded by
    \begin{align} \label{eq:bd-T-start-dagger}
      & \mathbb{E} \left[ \widetilde{T}^* - \widetilde{T}^\dagger \right]  \nonumber\\
      %
      %\leq& \frac{B}{K c_{min}} \cdot \left( \frac{b^{\dagger[t]}_{K+1}}{\mu^{\dagger[t]}_{K+1}} - \frac{c_{min} b^{*[t]}_{K+1}}{c_{max} \mu^{*[t]}_{K+1}} \right) K \mu_{max}  +  \frac{b^{\dagger[t]}_{K+1}}{\mu^{\dagger[t]}_{K+1}} K L (M^{\frac{1}{2}} d^{-1})^\alpha  \nonumber\\
      %
      \leq& \frac{1}{K b_{min}} \mathbb{E} \left[ \sum^{\widetilde{T}^\dagger}_{t=1} \left( \sum_{i\in \widetilde{\mathcal N}^{\dagger[t]}} p^{\dagger[t]}_i - \sum_{i\in \widetilde{\mathcal N}^{*[t]}} p^{*[t]}_i \right) \right]  \nonumber\\
      \leq& \frac{1}{K b_{min}} \mathbb{E} \left[ \sum^{\frac{B}{K b_{min}}}_{t=1} \left( \sum_{i\in \widetilde{\mathcal N}^{\dagger[t]}} p^{\dagger[t]}_i - \sum_{i\in \widetilde{\mathcal N}^{*[t]}} p^{*[t]}_i \right) \right]  \nonumber\\
      \leq& \frac{\mu_{max} b^2_{max} - \mu_{min} b^2_{min}}{K \mu_{min} b_{max} b^2_{min}} B + \frac{b_{max} \delta}{K \mu_{min} b^2_{min}} B
    \end{align}
    We finally complete the proof by combining (\ref{eq:bd-T-dagger}) and (\ref{eq:bd-T-start-dagger})
  \end{proof}

  According to \textbf{Algorithm}~\ref{alg:online-caci}, we have
  \begin{align} \label{eq:q2bq00}
    \sum_{i \in \widetilde{\mathcal N}^{*[t]}}  \frac{\mu_{Q(i)}}{b_i} \leq \sum_{i \in \widetilde{\mathcal N}^{\dagger[t]}}  \frac{\mu_{Q(i)}}{b_i}
  \end{align}
  We also have
  \begin{align} \label{eq:q2bq01}
    \mu_{Q(i)} - \delta \leq \mu_i \leq \mu_{Q(i)} + \delta, ~\forall i \in \mathcal{N}^{[t]}
  \end{align}
  according to \textbf{Lemma}~\ref{le:intracube}. Therefore, by substituting (\ref{eq:q2bq01}) into (\ref{eq:q2bq00}), we have
  \begin{align*}
    & \sum_{i \in \mathcal{N}^{*[t]}} \mu_i -  \sum_{i \in \mathcal{N}^{\dagger[t]}} \mu_i  \\
    \leq& 2K \delta + \frac{c_{max} - c_{min}}{c_{max}} \sum_{i \in \widetilde{\mathcal N}^{*[t]}} \mu_i \\
    \leq& \left(2\delta + \frac{(c_{max} - c_{min})\mu_{max} }{c_{max}} \right) K 
  \end{align*}
  and thus
  \begin{align} \label{eq:on-main-1}
    \mathbb{E} \left[ \sum^{\widetilde{T}}_{t=1} \left( \sum_{i \in \widetilde{\mathcal N}^{*[t]}} \mu_i - \sum_{i \in \widetilde{\mathcal N}^{\dagger[t]}} \mu_i \right) \right] \leq \frac{2\delta + \mu_{max}}{b_{min}}B
  \end{align}
  by considering $\widetilde{T} \leq \frac{B}{Kb_{min}}$. Therefore, by considering (\ref{eq:on-main-1}) and (\ref{eq:expdifftime}), we have
  \begin{align} \label{eq:on-main-3}
    & \mathbb{E} \left[ \sum^{\widetilde{T}^*}_{t=1} \sum_{i \in \widetilde{\mathcal N}^{*[t]}} \mu_i - \sum^{\widetilde{T}}_{t=1} \sum_{i \in \widetilde{\mathcal N}^{\dagger[t]}} \mu_i \right]  \nonumber\\
    %
    %=& \sum^{\widetilde{T}}_{t=1} \left( \sum_{i \in \widetilde{\mathcal N}^{*[t]}} \mu_i - \sum_{i \in \widetilde{\mathcal N}^{\dagger[t]}} \mu_i \right)+ \sum^{\widetilde{T}^*}_{t=\widetilde{T}+1} \sum_{i \in \widetilde{\mathcal N}^{*[t]}} \mu_i  \\
    %
    \leq& \mathbb{E} \left[ \sum^{\widetilde{T}}_{t=1} \left( \sum_{i \in \widetilde{\mathcal N}^{*[t]}} \mu_i - \sum_{i \in \widetilde{\mathcal N}^{\dagger[t]}} \mu_i \right) \right]+ K \mu_{max} \mathbb{E} \left[ \widetilde{T}^* - \widetilde{T} \right]    \nonumber\\
    %
    % \leq& \left( \frac{b_{max} \mu_{max} (\mu_{max}+\delta)}{b^2_{min} \mu_{min}} + \frac{2 \delta + \mu_{max}}{b_{min}} \right) B   \nonumber\\
    % %
    % & + \frac{4 b_{max} \mu_{max} K^2 (K+1) d^M}{b^3_{min} \Delta^2_{min}} \ln \frac{B}{K c_{min}}     \nonumber\\
    % %
    % & + \frac{b_{max} \mu_{max}}{b_{min}} \left( \frac{K \pi^2}{3} + 2 \right) d^M \nonumber\\
    %
    \leq& \left( \frac{b_{max}\mu^2_{max}}{b^2_{min}\mu_{min}} + \frac{\mu_{max}}{b_{min}} \right) B  +  \left( \frac{b_{max}\mu_{max}}{b^2_{min} \mu_{min}} + \frac{2}{b_{min}} \right)\delta B \nonumber\\
    & + \frac{4 b_{max} \mu_{max} K^2 (K+1) d^M}{b^3_{min} \Delta^2_{min}} \ln \frac{B}{K b_{min}}     \nonumber\\
    & + \frac{b_{max} \mu_{max}}{b_{min}} \left( \frac{K \pi^2}{3} + 2 \right) d^M
  \end{align}

\section{Proof of Proposition~\ref{prop:on-subregret2}}  \label{secapp:on-subregret2}
  According to (\ref{eq:q2bq01}), we get
  \begin{align}
    &\sum^{\widetilde{T}}_{t=1} \sum_{i \in \widetilde{\mathcal N}^{\dagger[t]}} \mu_i - \sum^{\widetilde{T}}_{t=1} \sum_{i \in \widetilde{\mathcal N}^{[t]}} \mu_i  \nonumber\\
    %
    %\leq& \sum^{\widetilde{T}}_{t=1} \sum_{i \in \widetilde{\mathcal N}^{\dagger[t]}} (\mu_{Q(i)} + L(M^{\frac{1}{2}} d^{-1})^\alpha)  \\
    %
    %&- \sum^{\widetilde{T}}_{t=1} \sum_{i \in \widetilde{\mathcal N}^{[t]}} (\mu_{Q(i)} - L(M^{\frac{1}{2}} d^{-1})^\alpha)  \\
    %
    \leq& \sum^{\widetilde{T}}_{t=1} \left( \sum_{i \in \widetilde{\mathcal N}^{\dagger[t]}} \mu_{Q(i)} - \sum_{i \in \widetilde{\mathcal N}^{[t]}} \mu_{Q(i)} \right) + 2 \widetilde{T} K \delta  \nonumber\\
    \leq& \sum^{\widetilde{T}}_{t=1} \left( \sum_{i \in \widetilde{\mathcal N}^{\dagger[t]}} \mu_{Q(i)} - \sum_{i \in \widetilde{\mathcal N}^{[t]}} \mu_{Q(i)} \right) + \frac{2\delta}{b_{min}}B 
  \end{align}
  where the second inequality holds as $\widetilde{T} \leq \frac{B}{K b_{min}}$. Since
  \begin{align}
    &\mathbb{E} \left[ \sum^{\widetilde{T}}_{t=1} \sum_{i \in \widetilde{\mathcal N}^{\dagger[t]}} \mu_{Q(i)} - \sum^{\widetilde{T}}_{t=1} \sum_{i \in \widetilde{\mathcal N}^{[t]}} \mu_{Q(i)} \right]  \nonumber\\
    \leq& \nabla_{max} \sum_{Q \in \mathcal{Q}}  \mathbb{E} \left[ C^{[\widetilde{T}]}_Q \right]   \nonumber\\
    \leq& d^M \nabla_{max} \left( \frac{4K^2 (K+1)}{(b_{min} \Delta_{min})^2} \ln\frac{B}{K b_{min}} + \frac{K \pi^2}{3} + 1 \right)
  \end{align}
  where we have the second inequality according to \textbf{Lemma}~\ref{le:bdcounter}, we then have
  \begin{align}  \label{eq:on-main-4}
    & \mathbb{E} \left[ \sum^{\widetilde{T}}_{t=1} \sum_{i \in \widetilde{\mathcal N}^{\dagger[t]}} \mu_i - \sum^{\widetilde{T}}_{t=1} \sum_{i \in \widetilde{\mathcal N}^{[t]}} \mu_i \right]  \nonumber\\
    \leq& \left( \frac{4K^2 (K+1)}{(b_{min} \Delta_{min})^2} \ln\frac{B}{K b_{min}} + \frac{K \pi^2}{3} + 1 \right) \nabla_{max} d^M  + \frac{2\delta}{b_{min}}B
  \end{align}

\end{appendices}

\end{document}